\documentclass[aos]{imsart}

\RequirePackage[OT1]{fontenc}
\RequirePackage{amsthm,amsmath}
\RequirePackage[]{natbib}
\RequirePackage[colorlinks,citecolor=blue,urlcolor=blue]{hyperref}
\usepackage{amsfonts}
\usepackage{amssymb}
\usepackage{multirow}
\usepackage{graphicx}
\usepackage{mathtools}
\usepackage[ruled,boxed,commentsnumbered]{algorithm2e}
\usepackage{subcaption}
\usepackage{enumitem}
\usepackage{booktabs}
\usepackage{pbox}
\usepackage{xr-hyper}

\arxiv{arXiv:1602.02210}

\startlocaldefs
\numberwithin{equation}{section}
\theoremstyle{plain}

\endlocaldefs

\makeatletter

\newcommand*{\addFileDependency}[1]{
	\typeout{(#1)}
	\@addtofilelist{#1}
	\IfFileExists{#1}{}{\typeout{No file #1.}}
}
\makeatother

\newcommand{\convP}{\overset{p}{\longrightarrow}}
\newcommand{\convD}{\overset{d}{\longrightarrow}}

\newcommand{\E}{\mathbb E}
\newcommand{\V}{\mathbb{V}}
\newcommand{\tr}{\text{tr}}
\newcommand{\R}{\mathbb R}
\newcommand{\N}{\mathcal N}

\newcommand{\mP}{\mu_0}
\newcommand{\mQ}{\mu_1}
\newcommand{\diag}{\mathrm{diag}}
\newcommand{\dP}{\mathbb{P}}
\newcommand{\dQ}{\mathbb{Q}}

\newcommand{\I}{\mathbb{I}}
\newcommand{\LDA}{\mathrm{LDA}}
\newcommand{\defn}{\stackrel{\mathrm{def}}{=}}

\newtheorem{theorem}{Theorem}[section]
\newtheorem{lemma}{Lemma}[section]

\newtheorem{proposition}{Proposition}[section]

\newtheorem{remark}{Remark}[section]

\begin{document}

\begin{frontmatter}
\title{Classification accuracy as a proxy\\ for two-sample testing}
\runtitle{Classification accuracy test}

\begin{aug}
\author{\fnms{Ilmun} \snm{Kim}
\ead[label=e1]{ilmunk@stat.cmu.edu}},
\author{\fnms{Aaditya} \snm{Ramdas}
\ead[label=e2]{aramdas@stat.cmu.edu}},\\
\author{\fnms{Aarti} \snm{Singh}
\ead[label=e3]{aarti@cs.cmu.edu}},
\and
\author{\fnms{Larry} \snm{Wasserman}
\ead[label=e4]{larry@stat.cmu.edu}
\ead[label=u1,url]{}}

\runauthor{Kim, Ramdas, Singh and Wasserman}

\affiliation{Carnegie Mellon University}

\address{Department of Statistics \& Data Science\\
	Carnegie Mellon University\\
	Pittsburgh, Pennsylvania 15213\\
	\printead{e1} \\
	\phantom{E-mail:}
	\printead*{e2} \\ 
	\phantom{E-mail:}
	\printead*{e3} \\
	\phantom{E-mail:}
	\printead*{e4}
}
\end{aug}

\begin{abstract}
	When data analysts train a classifier and check if its accuracy is significantly different from chance, they are implicitly performing a two-sample test. We investigate the statistical properties of this flexible approach in the high-dimensional setting. 
	We prove two results that hold for all classifiers in any dimensions: if its true error remains $\epsilon$-better than chance for some $\epsilon>0$ as $d,n \to \infty$, then (a) the permutation-based test is consistent (has power approaching to one), (b) a computationally efficient test based on a Gaussian approximation of the null distribution is also consistent. To get a finer understanding of the rates of consistency, we study a specialized setting of distinguishing Gaussians with mean-difference $\delta$ and common (known or unknown) covariance $\Sigma$, when $d/n \to c \in (0,\infty)$. 
	We study variants of Fisher's linear discriminant analysis (LDA) such as ``naive Bayes'' in a nontrivial regime when $\epsilon \to 0$ (the Bayes classifier has true accuracy approaching 1/2), and contrast their power with corresponding variants of Hotelling's test. 
	Surprisingly, the expressions for their power match exactly in terms of $n,d,\delta,\Sigma$, and the LDA approach is only worse by a constant factor, achieving an asymptotic relative efficiency (ARE) of $1/\sqrt{\pi}$ for balanced samples.
	We also extend our results to high-dimensional elliptical distributions with finite kurtosis.
	Other results of independent interest include minimax lower bounds, and the optimality of Hotelling's test when $d=o(n)$. Simulation results validate our theory, and we present practical takeaway messages along with natural open problems.
\end{abstract}

\begin{keyword}[class=MSC]
\kwd[Primary ]{62H15}
\kwd[; secondary ]{62E20}
\end{keyword}

\begin{keyword}
\kwd{classification accuracy}
\kwd{two sample testing}
\kwd{high-dimensional asymptotics}
\kwd{Hotelling's $T^2$ test}
\kwd{linear discriminant analysis}
\kwd{permutation test}
\end{keyword}

\end{frontmatter}

\section{Introduction}

The recent popularity of machine learning has resulted in the extensive teaching and utilization of prediction methods in theoretical and applied communities.
When faced with a hypothesis testing problem in practice, data scientists sometimes opt for a prediction-based test-statistic.
We study one example of this common practice in this paper, concerning arguably the most classical testing and prediction problems  ---   \textit{two-sample testing} (are the two underlying distributions the same?) and  \textit{classification} (learning a  classifier that separates the two distributions, implicitly assuming they are not the same). Practitioners familiar with machine learning but not the hypothesis testing literature often find it intuitive to perform testing in the following way: first learn a classifier, and then see if its accuracy is significantly different from chance and if it is, then conclude that the distributions are different. 

The central question that this paper seeks to answer is ``\textit{what are the pros and cons of the classifier-based approach to two-sample testing?}''. As we shall detail in Section \ref{sec:back}, the notion of \textit{cost} or \textit{price} that is appropriate for the Neyman-Pearson or Fisherian hypothesis testing paradigm, is the power achievable at a fixed false positive level $\alpha$ (in other words, the lowest possible type-2 error achievable at some prespecified target type-1 error). Indeed, we approach this question using the frequentist perspective of minimax theory. More formally, we can restate our question as ``\textit{when is the classifier-based test consistent, and how does its power compare to the minimax power?}''.

\subsection{Practical motivation}

Before we delve into the details, it is worth mentioning that even though this paper is a theoretical endeavor, the question was initially practically motivated. Many scientific questions are naturally posed as two-sample tests 
--- examples abound in epidemiology and neuroscience. As a hypothetical example from the latter, say we are interested in determining whether a particular brain region responds differently under two situations (say listening to loud harsh sounds vs soft smooth sounds), or for a person with a medical condition (patient) and a person without the condition (control). Often, one collects and analyzes brain data for the same patient under the two contrasting stimuli (to study the effect of change in that stimulus), or for different normal and ill patients under the same stimulus (to study effect of a medical condition). Since the work of \cite{golland2003permutation} where the authors examined permutation tests for classification with application to neuroimaging analysis, it has been increasingly common in the field of neuroscience---see \cite{zhu2008fisher,etzel2009introduction,pereira2009machine,stelzer2013statistical}---to assess whether there is a significant difference between the two sets of data collected by learning a classifier to differentiate between them (because, for instance, they may be more familiar with classification than two-sample testing). Neuroscientists call this style of brain decoding as pattern discrimination and a positive answer can be seen as preliminary evidence that the mental process of interest might occur within the portion of the brain being studied; see \cite{olivetti2012induction} for a discussion of related issues. This classification approach to two-sample testing has been considered in other application areas 
including genetics~\citep{yu2007two}, speech analysis~\citep{chen2009large}, credit scoring~\citep{xiao2014transfer}, churn prediction~\citep{xiao2015feature} and video content analysis~\citep{liu2018classifier}.

\subsection{Overview of the main results}
Our first contribution is to identify weak conditions on the classifier that suffice for both finite-sample or asymptotic type-1 error control, as well as for asymptotic consistency.
\begin{itemize}
	\item \textbf{Asymptotic test (Proposition~\ref{Proposition: Asymptotic test}):} We identify mild conditions under which the sample-splitting error of a general classifier \eqref{eq:errS} is asymptotically Gaussian as $n,d \rightarrow \infty$. We introduce a test based on this Gaussian approximation and prove its asymptotic type-1 error control. We also prove that a sufficient condition for its consistency (for its power to approach one) is that its true accuracy converges to $1/2+\epsilon$ for any constant $\epsilon>0$ as $n,d \rightarrow \infty$ at any relative rate. \\[-.8em]
	\item \textbf{Permutation test (Theorem~\ref{Theorem: permutation test}):} In addition to the asymptotic approach, we consider two types of random permutation procedures that yield a valid level $\alpha$ test in finite-sample scenarios. 
		Under the same conditions made before, we find the minimum number of permutations that guarantees that the resulting permutation test is consistent.
\end{itemize}
For technical reasons, it is convenient to present these results last, after suitable notation, lemmas and assumptions are developed in earlier sections.

The above results leave two natural questions open: first, whether we can derive a rate of consistency in special cases, and second, whether testing can be consistent even when the classifier accuracy asymptotically approaches chance (is not bounded away from half). We answer both affirmatively; our second contribution is to rigorously analyze the asymptotic power of tests using classification accuracy for Gaussian and elliptical distributions in a high-dimensional setting when the error of the Bayes optimal classifier approaches half. 
	In this direction, we have three main results:
\begin{itemize}
	\item \textbf{Power of the accuracy of LDA for Gaussian distributions with known $\Sigma$ (Theorem~\ref{thm: power of g-LDA}):} The considered test statistic~\eqref{Eq:LDA-test} is the centered and rescaled classification error of LDA estimated via sample splitting, when $\Sigma$ is known. Under standard interpretable assumptions (Section \ref{subsec:assumptions}), this test statistic converges to a standard normal in the high-dimensional setting (Theorem~\ref{thm: asymptotic distribution}) under both null and local alternative. Using this fact, we describe its local asymptotic power in expression~(\ref{Eq: Power of Fisher's LDA}). Comparing the latter with the minimax power~(\ref{eq:lowSNR}), we highlight that the performance of the accuracy test is comparable to but worse than the minimax optimal test, achieving an asymptotic relative efficiency (ARE) of $1/\sqrt{\pi} \approx 0.564$ for balanced sample sizes. \\[-.8em]
	\item \textbf{Extensions to unknown $\Sigma$ using naive Bayes and other variants (Theorem~\ref{thm: naive Bayes}):} We generalize the previous findings to other linear classifiers for unknown $\Sigma$, like naive Bayes. We again find that classifier-based tests are underpowered, achieving the same aforementioned ARE of $1/\sqrt{\pi}$ compared to corresponding variants of Hotelling's test such as \cite{bs} and \cite{sd}. \\[-.8em] 
	\item \textbf{Extensions to elliptical distributions (Theorem~\ref{thm: power under elliptical distributions}):} We extend Theorem~\ref{thm: power of g-LDA} to the class of (heavy-tailed) elliptical distributions with finite kurtosis, and prove that the asymptotic power expression matches the Gaussian setting up to an explicit constant factor, which is $\sqrt2$ times the marginal density evaluated at 0. Restricting our attention to multivariate $t$-distributions, we also find an interesting phenomenon that the classifier-based test becomes relatively more efficient when the underlying distributions have heavier tails (lower degrees of freedom).
\end{itemize}
As two side contributions, we formally study the fundamental minimax power of high-dimensional two-sample mean testing for Gaussians. In this direction, we have two main results. 
\begin{itemize}
	\item \textbf{Explicit and exact expression for asymptotic minimax power (Proposition~\ref{Proposition: minimax lower bound}):} By building on prior work \citep{luschgy1982minimax}, we provide an explicit expression for the asymptotic minimax power of high-dimensional two-sample mean testing that is valid for any (shared) positive definite covariance matrix and unbalanced sample sizes when $d,n\to\infty$ at any relative rate. \\[-.8em]
	\item \textbf{Minimax optimality of Hotelling's $T^2$ test when $d = o(n)$ (Theorem~\ref{Theorem: minimax power of Hotelling's test with unknown Sigma}):} It is well known that Hotelling's test is minimax optimal when $d$ is fixed and $n\to\infty$. In the high-dimensional setting, when the dimension $d$ and the sample size $n$ both increase to infinity with $d/n \rightarrow c \in (0,1)$, \cite{bs} show that Hotelling's test may have low power. Since then, Hotelling's test has been largely undervalued in the setting where $d$ increases with $n$. In contrast to the aforementioned negative result, we prove that Hotelling's test remains asymptotically minimax optimal when $d\to\infty$ as long as $d/n \rightarrow 0$.
\end{itemize}

\subsection{Interpreting our results and practical takeaway messages}

There may be two somewhat contradictory ways that our results may be interpreted:
\begin{enumerate}
	\item Practitioners may (possibly unjustly) use our results to reassure themselves that their utilization of prediction methods for testing, even in the high dimensional setting, may not hurt their power too much. 
	\item For scientific disciplines in which data is not abundant, scientists may be wary of using prediction methods for hypothesis testing problems due to the loss in power. 
\end{enumerate}
After our manuscript appeared on arXiv in early 2016, a few different papers have cited our results to justify their choices in both of these above ways. We take the liberty to weigh in on this possible conundrum, using our intuition from this paper and from experiments in other followup papers \citep[e.g.][]{lopez2016revisiting,rosenblatt2016better,hediger2019use} to instead propose complementary, non-contradictory takeaway messages:
\begin{enumerate}
	\item If the data is relatively unstructured or not abundant, and if the alternative can be accurately specified in such a manner that is both practically meaningful and for which a provably powerful two-sample test statistic is available (or can be easily designed), then we recommend using such a well-tailored statistic. \\[-.8em]
	\item Suppose the data is highly structured or abundant (say, images of two species of beetles), but the potential differences between the two distributions cannot be easily specified. In this case, constructing a refined test that has high power against an accurately prespecified alternative may be too hard, and thereby we recommend using a flexible two-sample test statistic like classification accuracy (say using a convolutional neural network classifier or random forests). 
\end{enumerate} 
Of course, it seems very challenging to theoretically study these setups in their full generality to provide a thorough formal backing to such practical suggestions. However, we are hopeful that our work will spur others to extend our concrete results to new settings.

\subsection{Related work}

The idea of using binary classifiers for two-sample testing was conceptualized by \cite{friedman2004multivariate}. However, Friedman's proposal was fundamentally different from the one proposed here:~he suggested using training a classifier on all points, and using that classifier to assign a score to each point. Then, he compared the scores in each class using a univariate two-sample test like Mann-Whitney or Kolmogorov-Smirnov. In other words, Friedman proposed using classifiers to reduce a multivariate two-sample test into a univariate one. A different classifier-based approach to the two-sample problem was proposed by \cite{blanchard2010semi}. Although their test is built upon classification algorithms, it estimates the a priori probability of a contamination model, instead of classification accuracy.

In contrast, this paper considers held-out accuracy as the test statistic. The held-out accuracy of any classifier in any dimension can be used as the test statistic, and type-1 error can always be controlled non-asymptotically at the desired level using permutations (see Section~\ref{sec: permutation}). Hence, the main question of genuine mathemtical interest is what we can prove about the power of such a test. 
To overcome the computational burden of permutations, if we instead use a Gaussian approximation to the null distribution, then it is unclear whether it remains valid in the high-dimensional setting and again its power is unclear.

To the best of our knowledge, our 2016 ArXiv manuscript was the first mathematical attempt to study the power of this general approach in a specialized setting. There has been a growing interest in this idea in both the statistics and the machine learning communities \citep{rosenblatt2016better,lopez2016revisiting,borji2019pros,hediger2019use,gagnon2019classification}, most of which directly build on our work, but further provide valuable practical insight into the problem using various classifiers under different scenarios. However, most of these other works couple informal heuristic arguments with numerical experiments, motivating us to fully formalize and further generalize our earlier analysis.



In an orthogonal work, \cite{scott05np} proposed a Neyman-Pearson classification framework within which one would like to minimize the probability of classification error for one class, subject to a bound on the probability of classification error for the other class. Their problem is a variant of classification in which the classifier is judged by a different error metric, but it is quite different from our goal of two-sample testing.
Other connections between classification and two-sample testing have also been explored by \cite{ben2007analysis}, \cite{fukumizu2009kernel} and \cite{mmd12}, but none of them set out to solve our problem.

Another class of two-sample tests is based on geometric graphs; examples include the $k$-nearest neighbor (NN) graph~\citep{schilling1986multivariate,henze1988multivariate}, the minimum spanning tree~\citep{friedman1979multivariate} and the cross-matching~\citep{rosenbaum2005exact}. Recently \cite{bhattacharya2018two} presented general asymptotic properties of graph-based tests under the fixed dimensional setting. Comparing the performance of the $k$-NN graph test and the $k$-NN classifier test (based on its heldout classification accuracy, as studied in this paper) may be interesting to explore in future work. 

There is of course a very large body of work that just analyzes classifiers, or just analyzes two-sample tests~\citep[e.g.][and the references therein]{hu2016review,arias2018remember}, but without connecting the two. 



\paragraph{Paper Outline} The rest of this paper is organized as follows. In Section~\ref{sec:back}, we formally define both testing and classification problems. In Section~\ref{sec:lowertst}, we discuss a minimax lower bound for two-sample testing in high-dimensional settings and  in Section~\ref{sec: Minimax Optimality of Hotelling's Test with unknown Sigma}, we prove that Hotelling's $T^2$ test achieves this lower bound when $d/n \rightarrow 0$. Section~\ref{sec: asymptotic distribution} studies the limiting distribution of Fisher's LDA accuracy in the high-dimensional setting. Building on this limiting distribution, Section~\ref{sec: power of LDA} presents the asymptotic power of Fisher's LDA for two-sample mean testing under known $\Sigma$. Section~\ref{sec: unknown sigma} extends this asymptotic power expression to other linear classifiers with unknown $\Sigma$, like naive Bayes. Generalizations to elliptical distributions are in Section~\ref{sec: Extension to elliptical distributions}. In Section~\ref{sec: Results on general classifiers}, we examine the type-1 error control and consistency of the asymptotic test as well as the permutation test for \emph{any} classifier. 
In Section~\ref{sec: experiments}, we provide simulation results that confirm our theoretical analysis, before concluding in Section \ref{sec:conc}. The proofs of all the results along with the discussion on open problems are provided in the supplement.


\paragraph{Notation}
Let $\N_d(\mu,\Sigma)$ refer to the $d$-variate Gaussian distribution with mean $\mu \in \R^d$ and $d \times d $ positive definite covariance matrix $\Sigma$. With a slight abuse of notation, we sometimes use $\N_d(z; \mu,\Sigma)$ to denote the corresponding density evaluated at $z$. The symbol $\| \cdot \|$ refers to the $L_2$ norm. Let $\I [ \cdot ]$ denote the standard 0-1 indicator function. Let $\Phi(\cdot)$ denote the standard Gaussian CDF, and let $z_\alpha$ be its upper $1-\alpha$ quantile. For a square matrix $A$, let $\text{diag}(A)$ denote the diagonal matrix formed by zeroing out the off-diagonal entries of $A$, and let $\lambda_{\text{min}}(A)$ and $\lambda_{\text{max}}(A)$ be the minimum and the maximum eigenvalues of $A$. We write the identity matrix as $I$. For sequences of constants $a_n$ and $b_n$, we write $a_n = O(b_n)$ if there exists a universal constant $c$ such that $|a_n/b_n| \leq c$ for all $n$ larger than some $n_0$, and we write $a_n=o(b_n)$ if $a_n/b_n \to 0$. Similarly, for a sequence of random variables $X_n$ and constants $a_n$, we write $X_n = O_P(a_n)$ if $a_n^{-1}X_n$ is stochastically bounded and $X_n =o_P(a_n)$ if $a_n^{-1}X_n$ converges to zero in probability.

\section{Background}\label{sec:back}

In this section, we introduce two-sample testing, including the special case of two-sample mean testing using Hotelling-type statistics and Fisher's linear discriminant analysis (LDA). We only introduce the basic versions here, later introducing variants like naive Bayes. 
We will be working in the high-dimensional setting where the number of samples $n$ and  dimension $d$ can both increase to infinity simultaneously. 

\subsection{Two-sample testing} \label{sec: two-sample mean testing}

Suppose that $X_{1},\ldots,X_{n_0},Y_1,\ldots,Y_{n_1}$ are independent random vectors in $\mathbb{R}^d$ such that $\mathcal{X}_1^{n_0} \defn \{X_{1},\ldots,X_{n_0}\}$ are identically distributed with the distribution $\dP_0$ and $\mathcal{Y}_1^{n_1} \defn \{Y_1,\ldots,Y_{n_1}\}$ are identically distributed with the distribution $\dP_1$. Given these samples, the two-sample problem aims at testing whether 
\begin{align} \label{Eq: distribution testing}
H_0: \dP_0 = \dP_1 ~\text{~ vs. ~}~ H_1: \dP_0 \neq \dP_1. 
\end{align}
While some of our results are on general classifiers and distributions (Section~\ref{sec: Results on general classifiers}), we often focus on the specific case where $\dP_0$ and $\dP_1$ are $d$-variate Gaussian distributions with densities $p_0(x) \defn \N_d(x; \mP, \Sigma )$ and $p_1(y) \defn \N_d(y; \mQ, \Sigma)$ respectively. We discuss the extension to heavy-tailed elliptical distributions in Section~\ref{sec: Extension to elliptical distributions}. When the Gaussians have equal covariance, the previous problem boils down to testing whether two distributions have the same mean vector or not.
This two-sample mean testing is a fundamental decision-theoretic problem, having a long history in statistics; for example, the past century has seen a wide adoption of the $T^2$-statistic by \cite{hotelling} to decide if two-samples have different population means (see \cite{hu2016review} for a review). Given the sample mean vectors $\widehat{\mu}_0 \defn \sum_{i=1}^{n_0} X_i/{n_0}$ and $\widehat{\mu}_1 \defn \sum_{i=1}^{n_1} Y_i/{n_1}$ and the pooled sample covariance matrix
\begin{align*}
\widehat{\Sigma} \defn \frac{1}{n_0+n_1-2} \Bigg[ \sum_{i=1}^{n_0} (X_i - \widehat{\mu}_0) (X_i - \widehat{\mu}_0)^\top + \sum_{i=1}^{n_1} (Y_i - \widehat{\mu}_1) (Y_i - \widehat{\mu}_1)^\top \Bigg],
\end{align*}
Hotelling's $T^2$-statistic is given by
\begin{align*}
T_{H} = (\widehat{\mu}_0 - \widehat{\mu}_1)^\top \widehat{\Sigma}^{-1} (\widehat{\mu}_0 - \widehat{\mu}_1).
\end{align*}
Hotelling's $T^2$ test based on $T_{H}$ was introduced for Gaussians, but it has been generalized to non-Gaussian settings as well \citep[e.g.,][]{kariya81}.

\subsection{Held-out classification accuracy}
Consider the same distributional setting described in the previous section. Given the samples $\mathcal{X}_1^{n_0}$ and $\mathcal{Y}_1^{n_1}$, classification is the problem of predicting to which class a new observation $Z$ belongs, i.e.~we want to predict whether $Z$ came from $\dP_0$ or $\dP_1$. Let the samples from $\dP_0$ and $\dP_1$ be given labels 0 and 1, respectively. A classifier $C$ is a function that maps a datapoint $Z$ to $\{0,1\}$.
Define the \emph{conditional} error of a classifier $C$ trained on the labeled data as:
\begin{eqnarray}
\hspace{-0.2in} \mathcal{E} &\defn& (\mathcal{E}_0 + \mathcal{E}_1)/2, \label{eq:errcond}\\
\hspace{-0.2in} \text{where} \quad  \mathcal{E}_0 &\defn&  \Pr_{Z \sim \dP_0}(C(Z) = 1 ~|~ \mathcal{X}_1^{n_0}, \mathcal{Y}_1^{n_1}), \nonumber \\
\hspace{-0.2in} \mathcal{E}_1 &\defn& \Pr_{Z \sim \dP_1}(C(Z) = 0 ~|~ \mathcal{X}_1^{n_0}, \mathcal{Y}_1^{n_1}). \nonumber
\end{eqnarray}
Clearly, $\mathcal{E}$ is a random variable that depends on the input data. Next, define the \emph{unconditional} error of $C$ as
\begin{eqnarray}
\hspace{-0.2in} E &\defn& (E_0 + E_1)/2,  \label{eq:err}\\
\hspace{-0.2in}  \text{where} \quad E_0 &\defn& \E_{n_0,n_1} \left[ \Pr_{Z \sim \dP_0}(C(Z) = 1 ~|~ \mathcal{X}_1^{n_0}, \mathcal{Y}_1^{n_1}) \right], \nonumber \\
\hspace{-0.2in} E_1 &\defn& \E_{n_0,n_1} \left[ \Pr_{Z \sim \dP_1}(C(Z) = 0 ~|~ \mathcal{X}_1^{n_0}, \mathcal{Y}_1^{n_1}) \right], \nonumber
\end{eqnarray}
where $\E_{n_0,n_1}$ denotes the expectation with respect to the $n_0$ and $n_1$ labeled datapoints. Note that $E$, $E_0$, $E_1$ do not depend on the input data and are only functions of $d,\delta,\Sigma,n \defn n_0 + n_1$. Importantly, if $\dP=\dQ$, chance performance is always $E=1/2$, no matter the ratio of sample sizes from each class (hence predicting the dominant label also achieves accuracy half).

Even though $E$ is unknown, one can estimate $E$ in a few different ways. One simple way is via sample splitting where the samples are split into training and test sets. Let us denote the number of samples of each class in the training (or test) set by $n_{0,\text{tr}}$ and $n_{1,\text{tr}}$ (or $n_{0,\text{te}}$ and $n_{1,\text{te}}$). In other words, there are $n_{\text{tr}} \defn n_{0,\text{tr}} + n_{1,\text{tr}}$ samples in the training set and $n_{\text{te}} \defn n_{0,\text{te}} + n_{1,\text{te}}$ samples in the test set. We then learn a classifier $\widehat{C}$ using $n_{\text{tr}}$ samples, and estimate its sample-splitting error using the remaining $n_{\text{te}}$ samples as:
\begin{eqnarray}
\widehat E^S &\defn& (\widehat E^S_0 + \widehat E^S_1)/2, \label{eq:errS}\\
\text{where} \quad  \widehat E^S_0 &\defn& \frac1{n_{0,\text{te}}}\sum_{i=1}^{n_{0,\text{te}}} \I \Big[ \widehat{C}(X_{n_{0,\text{tr}} + i}) = 1 \Big], \nonumber \\
\widehat E^S_1 &\defn& \frac1{n_{1,\text{te}}}\sum_{i=1}^{n_{1,\text{te}}} \I \Big[ \widehat{C}(Y_{n_{1,\text{tr}} + i}) = 0 \Big]. \nonumber
\end{eqnarray}
It is clear that the classifier will have a true accuracy significantly above half only if $\dP~\neq~\dQ$.  Hence one can use $\widehat E^S$ as a test statistic for two-sample testing, by checking whether $\widehat E^S$ is significantly less than half. 
The power of this approach is examined in Section~\ref{sec: Results on general classifiers}, but we begin with the special case of mean-testing using linear discriminant analysis.

\subsection{Fisher's linear discriminant classifier}

In the Gaussian setting, the optimal classifier is given by Bayes rule:
\begin{eqnarray*}
	\I \left[\log \frac{p_1(Z)}{p_0(Z)} > 0 \right] 
	= \I \left[ (\mu_1 - \mu_0)^\top \Sigma^{-1} \left(Z - \frac{(\mP + \mQ)}{2} \right) > 0 \right].
\end{eqnarray*}
We denote $\delta \defn \mu_1- \mu_0$ and $\mu_{\text{pool}} \defn (\mu_0 + \mu_1)/2$ so that we can succinctly write the Bayes rule as
\begin{align} \label{Eq: Bayes classifier}
C_{\text{Bayes}}(Z) \defn \I \Big[ \delta^\top \Sigma^{-1} (Z - \mu_{\text{pool}} ) > 0 \Big].
\end{align}
Then, by plugging in the estimators $\widehat{\delta} \defn \widehat{\mu}_1 - \widehat{\mu}_0$, $\widehat{\mu}_{\text{pool}} \defn (\widehat{\mu}_0 + \widehat{\mu}_1)/2$, and some appropriate choice if $\widehat \Sigma$, the linear discriminant analysis (LDA) rule is given by
$$
\LDA(Z) \defn \I \Big[ \widehat{\delta}^\top \widehat{\Sigma}^{-1} (Z - \widehat{\mu}_{\text{pool}} ) > 0 \Big].
$$
This classifier was derived by \cite{fisher1936use,fisher1940precision} from a generalized eigenvalue problem (hence also called Fisher's LDA) and was later developed further by \cite{wald1944statistical} and \cite{anderson1951classification}. 
We will show that the held-out accuracy of Fisher's LDA in the high-dimensional Gaussian setting is asymptotically Gaussian, and derive its power when used for two-sample testing (for various choices of $\widehat \Sigma$). 
We later extend these results to heavy-tailed ellpitical distributions.
However, we begin by understanding the fundamental minimax lower bounds for two-sample mean testing. 

\section{Lower bounds for two-sample mean testing}\label{sec:lowertst}

We first introduce some notation. Let $\mathcal{P}$ be a set that consists of all pairs of $d$-dimensional multivariate normal density functions whose covariance matrices coincide, and is positive definite. 
Let $\mathcal{P}_0$ be the subset of $\mathcal{P}$ such that each pair also has the same mean. For a given $\alpha \in (0,1)$, let us write  a level $\alpha$ test based on $\mathcal{X}_1^{n_0}$ and $\mathcal{Y}_1^{n_1}$ by $\varphi_\alpha$ and the collection of all level $\alpha$ tests by 
\begin{align*}
\mathcal{T}_{\alpha} \defn \big\{ \varphi_\alpha: \mathcal{X}_1^{n_0} \cup \mathcal{Y}_1^{n_1} \mapsto \{0,1\}: \sup_{p_0,p_1 \in \mathcal{P}_0 }\mathbb{E}_{p_0,p_1}[ \varphi_\alpha] \leq \alpha  \big\}.
\end{align*}
Additionally, we define a class of two multivariate normal density functions $p_0$ and $p_1$ whose distance is measured in terms of Mahalanobis distance parameterized by $\rho > 0$ as:
\begin{align*}
\mathcal{P}_1(\rho) \defn \{(p_0,p_1) \in \mathcal{P} : (\mu_0 - \mu_1)^\top \Sigma^{-1} (\mu_0 - \mu_1) \geq \rho^2 \}.
\end{align*}
The use of Mahalanobis distance is conventional and has been considered in \cite{giri1963minimax}, \cite{giri1964local} and \cite{salaevskii71} to study the minimax character of Hotelling's one-sample test. The ``oracle'' Hotelling's two sample test is defined as
\begin{align*}
\varphi_{H}^\ast = \I \bigg[ \frac{n_0n_1}{n_0 + n_1}(\widehat{\mu}_0 - \widehat{\mu}_1)^\top \Sigma^{-1} (\widehat{\mu}_0 - \widehat{\mu}_1) \geq c_{\alpha,d} \bigg],
\end{align*}
where $c_{\alpha,d}$ is the $1-\alpha$ quantile of the chi-squared distribution with $d$ degrees of freedom, and ``oracle'' signifies that $\Sigma$ is known.
\cite{luschgy1982minimax} extends the previous one-sample results and shows that $\varphi_{H}^\ast$
is minimax optimal over $\mathcal{P}_1(\rho)$, or more explicitly,
\begin{align} \label{Eq: Lower Bound}
\sup_{\varphi_\alpha \in \mathcal{T}_\alpha} \inf_{p_0,p_1 \in \mathcal{P}_1(\rho)} \mathbb{E}_{p_0,p_1}[ \varphi_\alpha ] = \inf_{p_0,p_1 \in \mathcal{P}_1(\rho)} \mathbb{E}_{p_0,p_1}[ \varphi_{H}^\ast],
\end{align}
for any finite $n$ and $d$. However, this result does not clearly show how the underlying parameters (e.g., $n$, $d$, $\rho$) interact to determine the power. To shed light on this, we study the asymptotic expression for the minimax power. Denote the sample size ratio by $\lambda_1 = \lambda_{1,n}  \defn n_1/n$. Recalling that $\Phi$ is the standard normal CDF and $z_\alpha$ its $1-\alpha$ quantile, we prove the following:

\begin{proposition} \label{Proposition: minimax lower bound}
	Consider a high-dimensional regime where $n,d \rightarrow \infty$ (at any rate). Then the minimax power for Gaussian two-sample mean testing is
	\begin{equation}
	\begin{aligned}  
	& \sup_{\varphi_\alpha \in \mathcal{T}_\alpha} \inf_{p_0,p_1 \in \mathcal{P}_1(\rho)} \mathbb{E}_{p_0,p_1}[ \varphi_\alpha ]  \\[.5em]
	=  ~ & \Phi \left( -\frac{\sqrt{2d}}{\sqrt{2d + n\lambda_1(1-\lambda_1) \rho^2}} z_\alpha + \frac{n\lambda_1(1-\lambda_1) \rho^2}{\sqrt{2d + 4n\lambda_1(1-\lambda_1) \rho^2}} \right) + o(1). \label{Eq: Power Expression}
	\end{aligned}
	\end{equation}
\end{proposition}
The proof is based on the central limit theorem and can be found in Appendix~\ref{sec: proof of proposition: minimax lower bound}. Notably, the expression (\ref{Eq: Power Expression}) is asymptotically precise including all constant terms and is valid without any restrictions on $d/n$ and $\lambda_1$. The way to interpret the bound in (\ref{Eq: Power Expression}) is as follows. The first term inside the parentheses is not of interest for our purposes, its magnitude being bounded by the constant $z_\alpha$. The second term is what determines the rate at which the power approaches one. When $\rho=0$, the power reduces to $\Phi(-z_\alpha)  = \alpha$ and if $d$ and $n$ are thought of as fixed, larger $\rho$ leads to larger power. The key in high dimensions, however, is how the power depends jointly on the signal to noise ratio (SNR) $\rho$, the dimension $d$ and the sample size $n$. To see this clearer, in the low SNR regime where $\rho^2= o(d/n)$ and $\lambda_1 \rightarrow \lambda \in (0,1)$, the minimax lower bound simplifies to
\begin{align}  \label{eq:lowSNR}
\Phi \left( -z_\alpha + \frac{n \lambda (1-\lambda) \rho^2}{\sqrt{2d}} \right) + o(1). 
\end{align}
It can be already seen that at constant SNR, $n$ only needs to scale faster than $\sqrt{d}$ for test power to asymptotically approach unity --- this $\sqrt{d}/n$ scaling is unlike the $d/n$ scaling typically seen in prediction problems \citep[for prediction error or classifier recovery, see][]{raudys2004results}. 
Next, we prove that this lower bound is tight even when $\Sigma$ is unknown, as long as $d=o(n)$.

\section{Minimax optimality of Hotelling's test when $d = o(n)$} \label{sec: Minimax Optimality of Hotelling's Test with unknown Sigma}
When $\Sigma$ is unknown, $\varphi_{H}^\ast$ is not implementable and thus it remains unclear whether the previous asymptotic lower bound is tight. In other words, we do not know whether there exists a test that has the same asymptotic minimax power as $ \varphi_{H}^\ast$ in all high-dimensional regimes with unknown $\Sigma$. Below, we partially close this gap by showing that Hotelling's test with unknown $\Sigma$ can achieve the same asymptotic minimax power as $\varphi_{H}^\ast$ when $d/n \rightarrow 0$. 
By letting $q_{\alpha,n,d}$ be the $1-\alpha$ quantile of the $F$ distribution with parameters $d$ and $n-1-d$, Hotelling's two-sample test with unknown $\Sigma$ is given by
\begin{align*}
\varphi_{H} = \I \bigg[ \frac{n_0n_1(n-d-1)}{n(n-2)d}(\widehat{\mu}_0 - \widehat{\mu}_1)^\top \widehat{\Sigma}^{-1} (\widehat{\mu}_0 - \widehat{\mu}_1) \geq q_{\alpha,n,d} \bigg].
\end{align*}
For Gaussians, it is well-known that $\varphi_{H} $ satisfies $\sup_{p_0,p_1 \in \mathcal{P}_0 } \mathbb{E}_{p_0,p_1}[ \varphi_{H} ]  \leq \alpha$ \citep[e.g.,][]{anderson58}. The next theorem studies the power of $\varphi_{H}$.

\begin{theorem} \label{Theorem: minimax power of Hotelling's test with unknown Sigma}
	Consider an asymptotic regime where $d/n \rightarrow 0$. Then the uniform power of $\varphi_{H}$ is asymptotically the same as that of $\varphi_{H}^\ast$ for Gaussian two-sample mean testing. In other words, as $n,d \rightarrow \infty$ with $d/n \rightarrow 0$, we have that $\inf_{p_0,p_1 \in \mathcal{P}_1(\rho)} \mathbb{E}_{p_0,p_1}[ \varphi_{H}]$ is equal to
	\begin{align*}
	\Phi \left( -\frac{\sqrt{2d}}{\sqrt{2d + n\lambda_1(1-\lambda_1) \rho^2}} z_\alpha + \frac{n\lambda_1(1-\lambda_1) \rho^2}{\sqrt{2d + 4n\lambda_1(1-\lambda_1) \rho^2}} \right) + o(1).
	\end{align*}
\end{theorem}
The proof can be found in Appendix~\ref{sec: proof of theorem: minimax power of Hotelling test with unknown Sigma}. 
 When $d>n$, $T_H$ is not even well-defined, but \cite{bs} demonstrate that even when $d/n \rightarrow c \in (0, 1)$ the power of $\varphi_{H} $ is poor. Due to its limitations, Hotelling's test has been largely neglected when $d$ increases with $n$. Unlike the previous negative results, Theorem~\ref{Theorem: minimax power of Hotelling's test with unknown Sigma} shows that it is minimax optimal when $d$ is allowed to grow with $n$, but $d/n \rightarrow 0$. We also provide empirical support for our asymptotic results in Figure~\ref{Figure: Power of Hotelling Test} of Section~\ref{sec: Asymptotic power of Hotelling's Test}.


\begin{remark} \label{Remark: Hotelling test with unknown Sigma}
	Combining the previous theorem with \cite{bs} and our simulation results in Section~\ref{sec: Asymptotic power of Hotelling's Test}, we may describe the phase transition behavior of Hotelling's test with unknown $\Sigma$ as
	\begin{itemize}
		\item Optimal regime (same power as $\varphi_{H}^\ast$): $d/n \rightarrow 0,$ \\[-.8em]
		\item Suboptimal regime (lower power than $\varphi_{H}^\ast$): $d/n \rightarrow c \in (0,1),$ \\[-.8em]
		\item Not applicable: $d/n \rightarrow c \geq 1.$
	\end{itemize}
\end{remark}

Even though Hotelling's test is suboptimal when $d=O(n)$, it is still an open problem to determine whether the lower bound is achievable by some other test, or whether a stronger lower bound can be proved.

\section{Asymptotic normality of the accuracy of generalized LDA} \label{sec: asymptotic distribution}
Here, we investigate the high-dimensional limiting distribution of the sample-splitting error in (\ref{eq:errS}). Building on the results developed in this section, we will present the power of the classification test in Section~\ref{sec: power of LDA}. Our main interest is in the setting where the dimension is comparable to or potentially much larger than the sample size. In this high-dimensional scenario, \cite{bickel2004some} prove that Fisher's LDA performs poorly in classification problems. When $d>n$, Fisher's LDA classifier is not even well-defined since $\widehat{\Sigma}$ is not invertible. Thus, \cite{bickel2004some} consider the naive Bayes (NB) classification rule by replacing $\widehat{\Sigma}^{-1}$ with the inverse of $\text{diag}(\widehat{\Sigma})$ and show that it outperforms Fisher's LDA in the high-dimensional setting. In the context of two-sample testing, we encounter the same issue on $\widehat{\Sigma}$ as mentioned earlier. To simplify our analysis, we start by assuming that $\Sigma$ is \emph{known} and analyze the asymptotic behavior of the corresponding Fisher's LDA statistic. Later in Section~\ref{sec: unknown sigma}, we extend the results to \emph{unknown} $\Sigma$ by considering the NB classifier and others.

\subsection{Assumptions}\label{subsec:assumptions}
Recalling that we work in the high-dimensional Gaussian setting with common covariance, let us detail some assumptions that facilitate our analysis. We assume that as $n = n_0 + n_1 \rightarrow \infty$, we have
\begin{itemize} \setlength{\itemindent}{1em}
	\item[\textbf{(A1)}] \emph{High-dimensional asymptotics}:  $\exists c \in (0, \infty)$ such that $d/n \rightarrow c$. \\[-.8em]
	\item[\textbf{(A2)}] \emph{Local alternative}: $\delta^\top \Sigma^{-1} \delta = O(n^{-1/2})$. \\[-.8em]
	\item[\textbf{(A3)}] \emph{Sample size ratio}: there exists $\lambda \in (0,1)$ such that $n_0/n \rightarrow \lambda$. \\[-.8em]
	\item[\textbf{(A4)}] \emph{Sample splitting ratio}: there exists $\kappa \in (0,1)$ such that $n_{\text{tr}}/n \rightarrow \kappa$. 
\end{itemize}
The asymptotic regime in \textbf{(A1)} is called \emph{Raudys-Kolmogorov double asymptotics} \citep[e.g.][]{zollanvari2011analytic} and assumes that $d$ increases linearly with $n$. In \textbf{(A2)}, we assume that  $\delta^\top \Sigma^{-1} \delta$ is close to zero such that a minimax test has nontrivial power. Note that under \textbf{(A1)}, the low SNR regime $\delta^\top \Sigma^{-1} \delta=o(d/n)$ is implied by \textbf{(A2)}. It is also interesting to note that the classification error of the Bayes optimal classifier~(\ref{Eq: Bayes classifier}) is computed as
	\begin{align*}
	\frac{1}{2} \Pr_{Z \sim \dP_0} \big\{ C_{\text{Bayes}}(Z)=1 \big\} +  \frac{1}{2} \Pr_{Z \sim \dP_1} \big\{ C_{\text{Bayes}}(Z)=0 \big\} = & 1 -  \Phi \Bigg( \frac{ \sqrt{\delta^\top \Sigma^{-1}\delta}}{2} \Bigg),
	\end{align*}
	which means that the classification error of the Bayes classifier, and hence \emph{any} classifier, approaches chance under \textbf{(A2)}. Assumption~\textbf{(A3)} rules out highly imbalanced cases and is common in the two-sample literature \citep[e.g.][]{bs,cq,skk13}. \textbf{(A4)} assumes that the user-chosen sample-splitting ratio is within $(0,1)$. We show in Theorem~\ref{thm: power of g-LDA} that the asymptotic power of the  test based on held-out classification accuracy is maximized when $\kappa = 1/2$ for the balanced case of $\lambda = 1/2$. In other cases, Theorem~\ref{thm: power of g-LDA} may serve as a guideline for choosing $\kappa$ that maximizes the asymptotic power. 
For any $d \times d$ symmetric positive definite matrix $A$, we define the generalized LDA classifier by
\begin{align} \label{Eq: generalized LDA}
\LDA_{A,n_0,n_1}(Z) \defn \I \Big[ \widehat{\delta}^\top A (Z - \widehat{\mu}_{\text{pool}} ) > 0 \Big].
\end{align}
Its sample-splitting error can be calculated using expression (\ref{eq:errS}):
\[
\widehat{E}_A^S \equiv \textnormal{classification error of } \LDA_{A,n_{0,\text{tr}},n_{1,\text{tr}}}(Z),
\]
emphasizing the dependency on the user-chosen matrix $A$. In terms of $\Sigma$ and $A$, we assume that:
\begin{itemize} \setlength{\itemindent}{1em}
	\item[\textbf{(A5)}] \emph{$\Sigma$ has bounded eigenvalues:} there exist constants $c_1,c_2$ such that 
	$0 < c_1 \leq \lambda_{\text{min}}(\Sigma) \leq \lambda_{\text{max}}(\Sigma) \leq c_2 < \infty$. \\[-.8em]
	\item[\textbf{\textbf{(A6)}}] \emph{$A$ has bounded eigenvalues:} there exist constants $c_1^\prime, c_2^\prime$ such that $0 < c_1^\prime \leq \lambda_{\text{min}}(A) \leq \lambda_{\text{max}}(A) \leq c_2^\prime < \infty$.
\end{itemize}
The same eigenvalue condition for $\Sigma$ was used by \cite{bickel2004some}. Assumption \textbf{(A6)} is satisfied when $A$ is diagonal with uniformly bounded entries, and when $A = \Sigma^{-1}$ under \textbf{(A5)}.

\subsection{Asymptotic normality for non-random $A$}
Given the previous assumptions, we study the asymptotic distribution of the sample-splitting error of the generalized LDA classifier when $A$ is non-random. Since Fisher's LDA with known $\Sigma$ is a special case of generalized LDA, it is straightforward to derive the limiting distribution of $\widehat{E}^S_{\Sigma^{-1}}$ from the general result. 

We first observe that the sample-splitting error of the generalized LDA classifier can be viewed as the average of independent observations when conditioning on the training set. Therefore it is natural to expect that the sample-splitting error is asymptotically normally distributed. To make this statement formal, we define $\mathcal{E}_{i,A}$ and $E_{i,A}$ similarly as $\mathcal{E}_{i}$ and $E_{i}$ for $i=1,2$ from definitions \eqref{eq:errcond} and \eqref{eq:err}, but by replacing the LDA classifier with the generalized LDA classifier with a given $A$. Then let us write the standardized test statistic as
\begin{align} \label{Eq: W_A}
W_{A} \defn \frac{\widehat{E}_A^S - \mathcal{E}_{0,A}/2 - \mathcal{E}_{1,A}/2}{\sqrt{\mathcal{E}_{0,A}(1-\mathcal{E}_{0,A})/(4n_{0,\text{{te}}})  + \mathcal{E}_{1,A}(1-\mathcal{E}_{1,A})/(4n_{1,\text{{te}}})}}.
\end{align}
In the next proposition, we present both \emph{conditional} and \emph{unconditional} limiting distributions of $W_{A} $ in the high dimensional setting. 

\begin{proposition} \label{proposition: conditional CLT}
	Suppose that the assumptions \textbf{(A1)}--\textbf{(A6)} hold. Then $W_A$ converges to a standard Gaussian conditional on the training set:
	\begin{align*}
	\sup_{t \in \mathbb{R}} | \Pr(W_A \leq t | \mathcal{X}_1^{n_{0,\text{\emph{tr}}}}, \mathcal{Y}_1^{n_{1,\text{\emph{tr}}}}) - \Phi(t) | = O_P(n^{-1/2}).
	\end{align*}
	Moreover, under the same assumptions, $W_A$ converges to the standard normal distribution unconditional on the training set:
	\begin{align*}
	\sup_{t \in \mathbb{R}} | \Pr(W_A \leq t ) - \Phi(t) | = o(1).
	\end{align*}
\end{proposition}

The proof is given in Appendix~\ref{sec: proof of proposition: conditional CLT}. Although the limiting distribution of $W_A$ is known from the previous lemma, it is quite challenging to determine the power of a test based classification accuracy by analyzing $W_A$. The reason is that $\mathcal{E}_{0,A}$ and $\mathcal{E}_{1,A}$ are random since they depend on the training set. To address this issue, we shall present a tractable approximation of $W_A$ that replaces $\mathcal{E}_{0,A}$ and $\mathcal{E}_{1,A}$ with non-random quantities.
To ease notation, let us denote $V_{0,A} \defn \widehat{\delta}^\top A (\mu_0 - \widehat{\mu}_{\text{pool}})$, $V_{1,A} \defn  \widehat{\delta}^\top A (\widehat{\mu}_{\text{pool}} -\mu_1)$ and $U_A \defn \widehat{\delta}^\top A \Sigma A \widehat{\delta}$. We would like to stress that $\widehat{\delta}$ and $\widehat{\mu}_{\text{pool}}$ are computed based only on the training set. Using this fact, $\mathcal{E}_{0,A}$ and $\mathcal{E}_{1,A}$ can be written as
\begin{align} \label{eq: definition of E}
\mathcal{E}_{0,A} = \Phi \bigg( \frac{V_{0,A}}{\sqrt{U_A}} \bigg)  \quad \text{and} \quad \mathcal{E}_{1,A}= \Phi \bigg( \frac{V_{1,A}}{\sqrt{U_A}} \bigg).
\end{align}
Further write the expectations of $V_{0,A}$, $V_{1,A}$ and $U_A$ by $\E[V_{0,A}] = \Psi_{A,n,d}  + \Xi_{A,n,d}$, $\E[V_{1,A}] = \Psi_{A,n,d}  - \Xi_{A,n,d}$ and $\E[U_A] = \Lambda_{A,n,d}$ where
\begin{equation}
\begin{aligned} \label{eq: psi/lambda}
& \Psi_{A,n,d} \defn -\frac{1}{2}\delta^\top A \delta, \\[.5em]
& \Lambda_{A,n,d} \defn \delta^\top A\Sigma A \delta + \left( \frac{1}{n_{0,\text{tr}}} + \frac{1}{n_{1,\text{tr}}} \right) \text{tr} \big\{  (A\Sigma)^2 \big\}, \\[.5em]
\text{ and } \quad & \Xi_{A,n,d} \defn  \frac{1}{2} \left( \frac{1}{n_{0,\text{tr}}} - \frac{1}{n_{1,\text{tr}}} \right) \tr(A\Sigma).
\end{aligned}
\end{equation}
Here the first two terms $\Psi_{A,n,d}$ and $\Lambda_{A,n,d}$ can be viewed as signal and noise terms, respectively, which ultimately determine the asymptotic power of the accuracy test. The third term $\Xi_{A,n,d}$ is an extra variance that comes from unbalanced sample sizes. Finally, we define a scaling factor
\begin{align} \label{Eq: definition of gamma}
\gamma_{A,n,d} \defn 2 \sqrt{\frac{n_{0,\text{te}}n_{1,\text{te}}}{n_{0,\text{te}} + n_{1,\text{te}}}}\frac{1}{\sqrt{\Phi (\Xi_{A,n,d}/\sqrt{\Lambda_{A,n,d}}) \{1 - \Phi (\Xi_{A,n,d}/\sqrt{\Lambda_{A,n,d}})\}}}.
\end{align}
With this notation in hand and letting $\phi(\cdot)$ be the standard normal density function, we now introduce an approximation of $W_A$ defined as 
\begin{align*}
W_A^\dagger \defn \gamma_{A,n,d} \cdot \Bigg\{ \widehat{E}_A^S - \frac{1}{2} -\phi \left( \frac{\Xi_{A,n,d}}{\sqrt{\Lambda_{A,n,d}}} \right) \frac{\Psi_{A,n,d}}{\sqrt{\Lambda_{A,n,d}}} \Bigg\}.
\end{align*}
It is clear that $W_A^\dagger$ is centered and scaled by explicit and non-random quantities. Next, we show that the difference between $W_A$ and $W_A^\dagger$ is asymptotically negligible and therefore $W_A^\dagger$ is also asymptotically standard normal. 
\begin{theorem} \label{thm: asymptotic distribution}
	Suppose that the assumptions \textbf{(A1)}--\textbf{(A6)} hold. Then we have that $W_A = W_A^\dagger + o_P(1)$ and thus the distribution of $W_A^\dagger$ converges to a standard normal:
	\begin{align*}
	\sup_{t \in \mathbb{R}} | \Pr(W_A^\dagger \leq t ) - \Phi(t) | = o(1).
	\end{align*}
\end{theorem}
The proof of Theorem~\ref{thm: asymptotic distribution} can be found in Appendix~\ref{sec: proof of thm: asymptotic distribution and thm: naive Bayes}. The asymptotic normality, established in the above theorem, holds under the null as well as under the local alternative \textbf{(A2)}. This enables us to explore the asymptotic power of the generalized LDA test with known $\Sigma$ in the next section, and we deal with unknown $\Sigma$ in the following section. 


\section{Asymptotic power of generalized LDA with non-random $A$} \label{sec: power of LDA}
Here, we study the asymptotic power of the generalized LDA test for known $\Sigma$. Since a smaller value of $\widehat{E}_A^S-1/2$ (or equivalently a larger value of the average per-class accuracy $1-\widehat{E}_A^S$) is in favor of $H_1: \mu_0 \neq \mu_1$, we define the test function by
\begin{align} \label{Eq:LDA-test}
\varphi_A \defn \mathbb{I} \Bigg[ \gamma_{A,n,d} \bigg( \widehat{E}_A^S - \frac{1}{2}\bigg) < -z_{\alpha} \Bigg].
\end{align} 
It is then clear from Theorem~\ref{thm: asymptotic distribution} that $\varphi_A$ has an asymptotic type-1 error controlled by $\alpha$. Now under the local alternative hypothesis, $\varphi_A$ has power given by
\begin{align}
\E[\varphi_A ] & = \Pr \Bigg( W_A^\dagger < -z_{\alpha} - \gamma_{A,n,d} \cdot \phi \left( \frac{\Xi_{A,n,d}}{\sqrt{\Lambda_{A,n,d}}} \right)  \frac{\Psi_{A,n,d}}{\sqrt{\Lambda_{A,n,d} }}  \Bigg), \nonumber \\[.5em]
& = \Phi \Bigg(-z_{\alpha} - \gamma_{A,n,d} \cdot \phi \left( \frac{\Xi_{A,n,d}}{\sqrt{\Lambda_{A,n,d}}} \right)  \frac{\Psi_{A,n,d}}{\sqrt{\Lambda_{A,n,d} }}  \Bigg) + o(1), \label{eq:power-midstep}
\end{align} 
where the second equality uses Theorem~\ref{thm: asymptotic distribution}. Let us write
\begin{align} \label{Eq: definition of beta}
\beta_{A,\lambda,\kappa} \defn \frac{\lambda-1/2}{\sqrt{\lambda(1-\lambda)\kappa}} \frac{n^{-1}\tr(A\Sigma)}{\sqrt{n^{-1} \tr\{(A\Sigma)^2\} }}.
\end{align}
Using assumptions \textbf{(A1)}--\textbf{(A6)}, 
the main term in the power function \eqref{eq:power-midstep} simplifies as
\begin{align*}
& - \gamma_{A,n,d} \cdot \phi \left( \frac{\Xi_{A,n,d}}{\sqrt{\Lambda_{A,n,d}}} \right)  \frac{\Psi_{A,n,d}}{\sqrt{\Lambda_{A,n,d} }} \\[.5em]
~=~ & \frac{\sqrt{2\kappa(1-\kappa)} \phi(\beta_{A,\lambda,\kappa})}{\sqrt{\Phi(\beta_{A,\lambda,\kappa})\{1 -\Phi(\beta_{A,\lambda,\kappa}) \}}} \cdot \frac{ n \lambda(1-\lambda) \delta^\top A \delta}{\sqrt{2\tr\{(A\Sigma)^2 \}}} + o(1).
\end{align*}
Resubstituting the above into expression \eqref{eq:power-midstep}, we finally infer that
\begin{align} \label{Eq: Power approximation}
\E[\varphi_A ]  = \Phi \Bigg( -z_{\alpha} + \frac{\sqrt{2\kappa(1-\kappa)} \phi(\beta_{A,\lambda,\kappa})}{\sqrt{\Phi(\beta_{A,\lambda,\kappa})\{1 -\Phi(\beta_{A,\lambda,\kappa}) \}}} \cdot \frac{ n \lambda(1-\lambda) \delta^\top A \delta}{\sqrt{2\tr\{(A\Sigma)^2 \}}}  \Bigg) + o(1).
\end{align}
Since $\sup_{x \in \mathbb{R}} \phi(x) / \sqrt{\Phi(x)\{1-\Phi(x)\}} = \sqrt{2/\pi}$ and its maximum is achieved at $x = 0$, the asymptotic power~(\ref{Eq: Power approximation}) is maximized when $\lambda = 1/2$ and $\kappa = 1/2$, further supported by simulations in Appendix~\ref{sec: Sample-Splitting Ratio}. However it is unknown whether the same result continues to hold for a random $A$ (e.g.~$A = \widehat{\Sigma}^{-1}$). 
In this balanced setting, the asymptotic power is further simplified as
\begin{align} \label{Eq: asymptotic power under Gaussian}
\Phi \Bigg( -z_{\alpha} + \frac{ n  \delta^\top A \delta}{\sqrt{32\pi\tr\{(A\Sigma)^2 \}}}  \Bigg) + o(1).
\end{align}
For ease of reference, we summarize our discussion as a theorem.

\begin{theorem} \label{thm: power of g-LDA}
	Suppose that the assumptions \textbf{(A1)}--\textbf{(A6)} hold. Then the generalized LDA test \eqref{Eq:LDA-test} asymptotically controls type-1 error at level $\alpha$ and its power for Gaussian two-sample mean testing is given by
	\begin{align} \label{Eq: Power of (general) Fisher's LDA}
	\E[\varphi_A ]  =\Phi \Bigg( -z_{\alpha} + \frac{\sqrt{2\kappa(1-\kappa)} \phi(\beta_{A,\lambda,\kappa})}{\sqrt{\Phi(\beta_{A,\lambda,\kappa})\{1 -\Phi(\beta_{A,\lambda,\kappa}) \}}} \cdot \frac{ n \lambda(1-\lambda) \delta^\top A \delta}{\sqrt{2 \emph{\tr}\{(A\Sigma)^2 \}}}  \Bigg)  + o(1).
	\end{align}
	Furthermore, keeping other parameters fixed, the asymptotic power is maximized when $\lambda = \kappa = 1/2$ (corresponding to a balanced train/test split).
\end{theorem}
The proof of the above theorem follows immediately from the preceding discussion and so is omitted. As a direct consequence of Theorem~\ref{thm: power of g-LDA}, when $\lambda = 1/2$ and $\kappa = 1/2$, the power of the ``oracle'' Fisher's LDA test that uses $A=\Sigma^{-1}$ (again, ``oracle'' is used because it uses $\Sigma^{-1}$) becomes
\begin{align} \label{Eq: Power of Fisher's LDA}
\E[\varphi^*_{\Sigma^{-1}}] = \Phi \bigg( -z_{\alpha} +  \frac{n \delta^\top \Sigma^{-1} \delta}{\sqrt{32\pi d}} \bigg) + o(1).
\end{align}
Comparing the above power with the minimax lower bound expression~(\ref{eq:lowSNR}) with $\lambda = 1/2$, we may conclude that the classification accuracy test can achieve essentially minimax optimal power, up to the small constant factor $1/\sqrt{\pi} \approx 0.564$. In other words, we pay a constant factor by performing a two-sample test via classification. However, this conclusion should be treated with caution as emphasized below:
\begin{itemize}
	\item First, Theorem~\ref{thm: power of g-LDA} is a pointwise result. That means, the result holds for any sequence of distributions satisfying the assumptions, but not uniformly over a class of distributions. Hence, conceptually, this is weaker than the uniform power achieved by $\varphi_{H}^\ast$ in Theorem~\ref{Theorem: minimax power of Hotelling's test with unknown Sigma}. However, this drawback actually applies to almost every published result on high-dimensional two-sample testing that we are aware of (or certainly all those that we cite), and it is a much broader open problem to prove that the power guarantees for these tests hold uniformly over the relevant classes. \\[-.5em] 
	\item Second, although a constant factor is not of major concern in determining the minimax rate, it may have a significant effect on power in practice. To see this, let $n_{\text{Fisher}}$ and $n_{\text{Hotelling}}$ be the sample sizes needed for $\varphi^*_{\Sigma^{-1}}$ and $\varphi_{H}^\ast$ to obtain the same power against the local alternative considered in Theorem~\ref{thm: power of g-LDA}. Then the asymptotic relative efficiency (ARE) of $\varphi^*_{\Sigma^{-1}}$ with respect to $\varphi_{H}^\ast$ is defined as the limit of the ratio $n_{\text{Hotelling}}/n_{\text{Fisher}}$~\citep[e.g.~Chapter 14 of][]{van2000asymptotic}. Based on the asymptotic power expressions~(\ref{eq:lowSNR}) and (\ref{Eq: Power of (general) Fisher's LDA}), a simple closed-form expression of the ARE is available as
		\begin{align} 
		\text{ARE}(\varphi_{\Sigma^{-1}}^\ast;\varphi_{H}^\ast) & = \frac{\sqrt{2\kappa(1-\kappa)} \phi(\beta^\ast)}{\sqrt{\Phi(\beta^\ast)\{1 -\Phi(\beta^\ast)\}}} \leq \frac{1}{\sqrt{\pi}} \approx 0.564,  \label{Eq: ARE}
		\end{align} 
		where $\beta^\ast  = \lim_{n,d \rightarrow \infty} \beta_{\Sigma^{-1},\lambda,\kappa}$ if it exists. This ARE expression implies that $\varphi^*_{\Sigma^{-1}}$ requires (at least) $\sqrt{\pi} \approx 1.77$ more samples to attain approximately the same power as $\varphi_{H}^\ast$. In this context, Hotelling's test should be preferred over the classifier-based test to obtain higher power against the Gaussian mean shift alternative.
\end{itemize}
In the following sections, we extend the results on the oracle Fisher's LDA classifier to it variants with unknown $\Sigma$ and also to elliptical distributions.



\begin{remark} \normalfont
	As mentioned in Section~\ref{subsec:assumptions}, the accuracy of the Bayes optimal classifier approaches half under the considered asymptotic regime, meaning that no classifier can have accuracy better than a random guess in the limit. In contrast, under the same asymptotic regime, two-sample testing based on generalized LDA can have non-trivial power (strictly greater than $\alpha$) as shown in Theorem~\ref{thm: power of g-LDA}. These two results not only demonstrate that testing is easier than classification, but also that the local alternative \textbf{(A2)} is conceptually interesting --- it corresponds to a regime where the LDA classifier performs as poorly as a random guess for classification, but is essentially optimal for testing.
\end{remark}

\section{Naive Bayes: power of generalized LDA with unknown $\Sigma$} \label{sec: unknown sigma}

For low-dimensional Gaussians with unknown $\Sigma$, there are strong reasons to prefer Hotelling's test; it is well-known that it is \emph{uniformly most powerful} among all tests that are invariant with respect to nonsingular linear transformations \citep[e.g.,][]{anderson58}. We also refer to \cite{simaika41,giri1963minimax,giri1964local,salaevskii71,kariya81,luschgy1982minimax} for other optimality properties of Hotelling's test in finite $d$ and $n$ settings. Moreover our result in Theorem~\ref{Theorem: minimax power of Hotelling's test with unknown Sigma} says that $\varphi_H$ is asymptotically minimax optimal among all level $\alpha$ tests as long as $d/n \rightarrow 0$. Unfortunately, when $d$ is linearly comparable to or larger than $n$, these optimal properties of Hotelling's test becomes highly non-trivial. In particular, $\varphi_H$ has asymptotic power tending to the (trivial) value of $\alpha$ in the high dimensional setting, when $d,n \to \infty$ with $d/n \to 1-\epsilon$ for small $\epsilon>0$ \citep[][for details]{bs}. The problem becomes even worse when the dimension is larger than the sample size as $T_H$ is not well-defined.

The aforementioned issue on $T_H$ has motivated the study of alternative two-sample mean test statistics in the high-dimensional setting. For instance, \cite{bs} show that dropping $\widehat \Sigma$ from the Hotelling test statistic (i.e.~replacing $\widehat{\Sigma}$ with the identity matrix) entirely leads to a test that does have asymptotic power tending to one in the high-dimensional setting where Hotelling's test fails. The test statistic proposed by \cite{bs} can be essentially written as
\begin{align*}
T_{BS} \defn (\widehat{\mu}_0 - \widehat{\mu}_1)^\top (\widehat{\mu}_0 - \widehat{\mu}_1).
\end{align*}
Following that, \cite{sd} propose (in a similar spirit) the test statistic
\begin{align} \label{Eq: SD statistic}
T_{SD} \defn (\widehat{\mu}_0 - \widehat{\mu}_1)^\top \diag(\widehat \Sigma)^{-1} (\widehat{\mu}_0 - \widehat{\mu}_1),
\end{align}
by replacing $\widehat \Sigma$ with $\diag(\widehat \Sigma)$ in Hotelling's statistic. They show that $T_{SD}$ also leads to high-dimensional consistency.  

As mentioned earlier, the idea of using $\diag(\widehat{\Sigma})$ in place of $\widehat{\Sigma}$ has also been justified in the high-dimensional classification problem \citep{bickel2004some}. In particular, the naive Bayes classifier (corresponding to $T_{SD}$) outperforms Fisher's LDA classifier (corresponding to $T_H$) in terms of the worst-case classification error in the high-dimensional setting. We note that this relatively understated connection between two-sample testing and classification has important implications for extending our previous results to other linear classifiers. Specifically, as we shall see, the power of the classifier-based tests is only worse by a constant factor than the variants of Hotelling's test when both the classifier and the two-sample test use the same substitute for $\Sigma^{-1}$.

To start, let us consider two classifiers with unknown $\Sigma$. The first one is the naive Bayes classifier and the other is the generalized LDA classifier with the identity matrix, i.e.~$A=I$. We then compare the power of the corresponding classification accuracy tests with the two-sample mean tests based on $T_{SD}$ and $T_{BS}$. Throughout this section, we assume that $n_0=n_1$, $n_{0,\text{tr}}=n_{1,\text{tr}}$ and $n_{\text{tr}} = n_{\text{te}}$ for simplicity.

From Theorem~\ref{thm: power of g-LDA}, the asymptotic power of the test based on $\widehat{E}_I^S$ is already available as
\begin{align} \label{Eq: power of I}
\E[\varphi_I ]  = \Phi \Bigg( -z_{\alpha} + \frac{n\delta^\top \delta}{\sqrt{32\pi \text{{tr}}(\Sigma^2)}}  \Bigg) + o(1).
\end{align}
Under more general conditions than the assumptions \textbf{(A1)}--\textbf{(A6)}, \cite{bs} show that the asymptotic power of the test based on $T_{BS}$, denoted by $\varphi_{BS}$, is 
\begin{align} \label{Eq: power of BS}
\E[\varphi_{BS}] =  \Phi \Bigg( -z_{\alpha} + \frac{n\delta^\top \delta}{\sqrt{32 \text{{tr}}(\Sigma^2)}}  \Bigg) + o(1).
\end{align}
Now by comparing two power expressions in (\ref{Eq: power of I}) and (\ref{Eq: power of BS}), we arrive at the same conclusion as before that the classification accuracy test is less powerful than the corresponding two-sample test $\varphi_{BS}$ by the constant factor $1/\sqrt{\pi} \approx 0.564$. 

Next we focus on the naive Bayes classifier and compute the asymptotic power of the resulting test. Although the analysis proceeds similarly to the previous one, we now need to deal with the randomness from the inverse diagonal matrix, which requires extra non-trivial work. By putting $\widehat{D}^{-1} \defn \text{diag}(\widehat{\Sigma})^{-1}$ and $D^{-1} = \text{diag}(\Sigma)^{-1}$, the asymptotic power of the naive Bayes classifier is provided as follows. 
\begin{theorem} \label{thm: naive Bayes}
	Consider the case where $n_0=n_1$, $n_{0,\text{\emph{tr}}}=n_{1,\text{\emph{tr}}}$ and $n_{\text{\emph{tr}}} = n_{\text{\emph{te}}}$. Then under the assumptions \textbf{(A1)}, \textbf{(A2)} and \textbf{(A5)}, the power of the naive Bayes classifier test for Gaussian two-sample mean testing is 
	\begin{align} \label{eq: power of naive Bayes classifier}
	\E[\varphi_{\widehat{D}^{-1}} ]  = \Phi \Bigg( -z_{\alpha} + \frac{n \delta^\top D^{-1} \delta}{\sqrt{32\pi \text{\emph{tr}}\{ (D^{-1}\Sigma)^{2}\}}}  \Bigg) + o(1).
	\end{align}
\end{theorem}
The proof of Theorem~\ref{thm: naive Bayes} can be found in Appendix~\ref{sec: proof of thm: asymptotic distribution and thm: naive Bayes}. \cite{sd} study the asymptotic power of the test $\varphi_{SD}$ based on $T_{SD}$ \eqref{Eq: SD statistic}.
One can also check that their conditions are fulfilled under the assumptions \textbf{(A1)}--\textbf{(A5)}. Using $\lambda=1/2$, the power of $\varphi_{SD}$ is given by
\begin{align*}
\E[\varphi_{SD} ]  = \Phi \Bigg( -z_{\alpha} + \frac{n \delta^\top D^{-1} \delta}{\sqrt{32 \text{{tr}}\{ (D^{-1}\Sigma)^{2}\}}}  \Bigg) + o(1).
\end{align*}
Comparing this with the asymptotic power of $\varphi_{\widehat{D}^{-1}}$ in (\ref{eq: power of naive Bayes classifier}), we see that the power of the accuracy test based on the naive Bayes classifier is worse than the corresponding two-sample test $\varphi_{SD}$, once again achieving an ARE of exactly $1/\sqrt{\pi}$.

\section{Extension to elliptical distributions} \label{sec: Extension to elliptical distributions}
In this section we extend our main result (Theorem~\ref{thm: power of g-LDA}) to the class of elliptical distributions and show that the asymptotic power expression remains the same up to a constant factor. Let $\mu$ be a $d$-dimensional vector, $S$ be a $d \times d$ positive semi-definite matrix, $\xi(\cdot )$ be a nonnegative function. A random vector $Z$ in $\mathbb{R}^d$ is said to have an elliptical distribution with location parameter $\mu$, scale matrix $S$ and generator $\xi(\cdot)$ if its characteristic function satisfies
	\begin{align*}
	\E \big[e^{it^\top Z} \big] = e^{it^\top \mu} \xi \big( t^\top S t \big) \quad \text{for all $t \in \mathbb{R}^d$.}
	\end{align*}
	When the second moment exists, it can be verified that $\mu$ corresponds to the mean vector of $Z$ and $S$ is proportional to the covariance matrix of $Z$, denoted by $\Sigma$. More specifically, by letting $\xi'(0)$ be the first derivative of $\xi$ evaluated at zero, $S$ is explicitly linked to $\Sigma$ as $-2 \xi'(0) S = \Sigma$. Notable examples of elliptical distributions include the multivariate normal, the multivariate student $t$, the multivariate Laplace and the multivariate logistic distribution. We refer to \cite{gomez2003survey,frahm2004generalized,fang2018symmetric} for further properties and examples of elliptical distributions. To have an explicit power expression, we make two extra assumptions on $Z$ described as follows: 
	\begin{itemize} \setlength{\itemindent}{1em}
		\item[\textbf{(A7)}] \emph{Condition on kurtosis parameter}: let $\zeta_{\text{kurt}}$ be the kurtosis parameter of $Z$ defined as
		\begin{align*}
		\zeta_{\text{kurt}} \defn \frac{\E\big[ \big\{(Z-\mu)^\top \Sigma^{-1} (Z - \mu) \big\}^2\big]}{d(d+2)} - 1.
		\end{align*} 
		We assume that there exists a positive constant $M$ such that $\zeta_{\text{kurt}} < M$ for all $n,d$. \\[-0.8em]
		\item[\textbf{(A8)}] \emph{Condition on density function}: assume that the standardized first coordinate of $Z$, that is $e_1^\top (Z -\mu) /  (e_1^\top \Sigma e_1)^{1/2}$ where $e_1 = (1,0,\ldots,0)^\top$, has the density function $f_{\xi}(\cdot)$ with respect to the Lebesgue measure. We further assume that $f_\xi$ is bounded and continuously differentiable. 
	\end{itemize}
	We believe that the condition on $\zeta_{\text{kurt}}$ in \textbf{(A7)} is mild and satisfied for many elliptical distributions \citep[e.g.,][]{zografos2008mardia}. For example, the kurtosis parameter of the multivariate $t$-distribution with $\nu$ degrees of freedom is $2/(\nu -4)$ for $\nu > 4$, which in turn implies that $\zeta_{\text{kurt}}$ is zero for the Gaussian case. To interpret \textbf{(A8)}, we note that each component of an elliptical random  vector has the same distribution after standardization. Assumption \textbf{(A8)} then states that this common distribution has the density function $f_{\xi}$ with some extra regularity conditions. Clearly $f_{\xi}$ corresponds to the standard normal density function for the Gaussian case that is bounded and continuously differentiable. But \textbf{(A8)} fails to hold for the Laplace distribution whose density function is not differentiable at zero. With these extra assumptions, we are now ready to present the main result of this section, which generalizes Theorem~\ref{thm: power of g-LDA} to elliptical distributions. 
	\begin{theorem} \label{thm: power under elliptical distributions}
		Suppose that $\mathbb{P}_0$ and $\mathbb{P}_1$ are elliptical distributions with parameters $(\mu_0, S, \xi)$ and $(\mu_1, S, \xi)$, respectively. Consider the case where $n_0=n_1$, $n_{0,\text{\emph{tr}}}=n_{1,\text{\emph{tr}}}$ and $n_{\text{\emph{tr}}} = n_{\text{\emph{te}}}$, i.e.~$\lambda = \kappa = 1/2$, for simplicity. Then under the assumptions \textbf{(A1)}, \textbf{(A2)} and \textbf{(A5)}--\textbf{(A8)}, the generalized LDA test~(\ref{Eq:LDA-test}) asymptotically controls type-1 error at level $\alpha$ and has the asymptotic power for testing the hypothesis~(\ref{Eq: distribution testing}) as
		\begin{align} \label{Eq: asymptotic power for elliptical distributions}
		\E[\varphi_A ] = \Phi \Bigg( -z_{\alpha} + \frac{f_{\xi}(0) \cdot n  \delta^\top A \delta}{\sqrt{16 \emph{\tr}\{(A\Sigma)^2 \}}}  \Bigg) + o(1).
		\end{align}
	\end{theorem}
	The above result shows that the asymptotic power expression in Theorem~\ref{thm: power of g-LDA} does not change in terms of $n,d,\Sigma,A,\delta,$ for elliptical distributions. To further illustrate the result, let us consider the specific case where $\mathbb{P}_0$ and $\mathbb{P}_1$ are multivariate $t$-distributions with $\nu$ degrees of freedom and the same scale matrix. We additionally assume that $\nu > 4$ under which the assumption~\textbf{(A7)} is satisfied. In such a case, $f_{\xi}(0) =  f_{\xi}(0;\nu)$ equals
	\begin{align*}
	f_{\xi}(0;\nu) = \frac{\Gamma\left( \frac{\nu+1}{2} \right)}{\sqrt{\pi(\nu -2)} \Gamma\left( \frac{\nu}{2} \right)} \rightarrow \frac{1}{\sqrt{2\pi}} \approx 0.399 \quad \text{as $\nu \rightarrow \infty$.}
	\end{align*}
	Hence, by taking $\nu \rightarrow \infty$, the asymptotic power (\ref{Eq: asymptotic power for elliptical distributions}) recovers the previous power expression (\ref{Eq: asymptotic power under Gaussian}) for the Gaussian case. Indeed $f_{\xi}(0;\nu)$ is a decreasing sequence of $\nu$ such that $f_{\xi}(0;\nu) < f_{\xi}(0;4) \approx 0.530$ for all $\nu > 4$. This fact demonstrates that the generalized LDA test becomes relatively more efficient when the underlying $t$-distributions have heavier tails, which is also validated by simulations (see Figure~\ref{Figure: theory and practice}).


\section{Results on general classifiers} \label{sec: Results on general classifiers}
So far we have focused on the accuracy tests based on linear classifiers and derived their explicit asymptotic power against local alternatives under Gaussian or elliptical distribution assumptions. In this section, we turn to more general settings and examine two key properties, namely the type-1 error control and consistency, of the accuracy test based on a general classifier. The main result of this section shows that a classification accuracy test achieves asymptotic power equal to one, provided that the corresponding classifier has an accuracy higher than chance. This result naturally motivates questions about rate, for which more assumptions are needed, and also motivates studying a more challenging setting where the true accuracy approaches half, like the one we consider for the generalized LDA test.


Recall that for a generic classifier $\widehat{C}$ based on the training set, the per-class and total errors $\widehat E^S_{0}(\widehat{C})$, $\widehat E^S_{1}(\widehat{C})$ and $\widehat E^S(\widehat{C})$ are calculated using expression~(\ref{eq:errS}). 
	To facilitate analysis, we assume the following asymptotic properties of $\widehat E^S_{0}(\widehat{C})$ and $\widehat E^S_{1}(\widehat{C})$:
	\begin{itemize} \setlength{\itemindent}{1em}
		\item[\textbf{(A9)}] \emph{Asymptotic classification errors}: assume that $\widehat E^S_{0}(\widehat{C}) = E_{0}(C) + o_P(1)$ and $\widehat E^S_{1}(\widehat{C}) =  E_{1}(C) + o_P(1)$ where $E_{1}(C)$ and $E_{2}(C)$ are constants in $(0,1)$. Moreover, there exists a constant $\epsilon >0$ such that $E_{0}(C)/2 + E_{1}(C)/2 = 1/2 - \epsilon$ under the alternative hypothesis. 
	\end{itemize}
	To determine the significance threshold for deciding if the error is different from chance, we consider two methods:~(1) the Gaussian approximation that underlies our theory in the preceding sections and (2) the permutation procedure with finite sample guarantees that has been common in practice. 

\subsection{Asymptotic test} \label{sec: asymptotic test}
As discussed before, the sample-splitting error can be viewed as the sum of independent random variables given the training set. Therefore it is natural to expect that this empirical error follows closely a normal distribution even for a general classifier when the sample size is large. Building on this intuition, we define the asymptotic test as
\begin{align*}
\I \left[ \frac{2\widehat E^S(\widehat{C}) - 1}{\sqrt{\widehat E^S_{0}(\widehat{C}) \big\{1-\widehat E^S_{0}(\widehat{C}) \big\} \big/n_{0,\text{te}}  + \widehat E^S_{1}(\widehat{C}) \big\{1-\widehat E^S_{1}(\widehat{C}) \big\} \big/ n_{1,\text{te}} } } <  -z_\alpha \right]
\end{align*}
and denote it by $\varphi_{\widehat{C},\text{Asymp}}$. We note that the quantity inside of the indicator function is a studentized sample-splitting error under the null hypothesis. In the next proposition we prove that the normal approximation is indeed accurate and thus $\varphi_{\widehat{C},\text{Asymp}} $ is a valid test at least asymptotically. Moreover, when the sequence of classification errors tends to a constant that is strictly less than chance level, we show that the power of the asymptotic test tends to one as $n \rightarrow \infty$ potentially with $d \rightarrow \infty$. 
\begin{proposition} \label{Proposition: Asymptotic test}
	Suppose that the assumptions \textbf{(A3)}, \textbf{(A4)} and \textbf{(A9)} hold as $n \rightarrow \infty$ potentially with $d\rightarrow \infty$ at any relative rate. Then under the null hypothesis $H_0: \mathbb{P}_0 = \mathbb{P}_1$, we have $\lim_{n \rightarrow \infty} \E_{H_0} \big[\varphi_{\widehat{C},\text{\emph{Asymp}}} \big] \leq \alpha$. On the other hand, under the alternative hypothesis $H_1: \mathbb{P}_0 \neq \mathbb{P}_1$, the asymptotic test is consistent as $\lim_{n \rightarrow \infty} \E_{H_1} \big[\varphi_{\widehat{C},\text{\emph{Asymp}}}\big] = 1$.
\end{proposition}
Despite its simplicity, the asymptotic approach has no finite sample guarantee. 
Next, we prove consistency of permutation-based approaches.


\subsection{Permutation tests} \label{sec: permutation}


In practice, one often employs permutation tests that can offer exact control of the type-1 error rate. There are two possible ways of applying permutation testing within the classification via sample splitting framework. The methods below differ in the italicized text.

\vskip .5em

\noindent \textbf{Method 1 (Half-permutation):} 
\begin{itemize}
	\item Split data into two halves, $X^1, Y^1$ and $X^2, Y^2$. Train the classifier on $X^1,Y^1$, call it $f^*$. Evaluate accuracy of $f^*$ on $X^2, Y^2$, call it $a^*$.  \\[-.8em]
	\item Repeat $P$ times: \textit{Pool the samples $X^2, Y^2$ into one bag, randomly permute the samples, and then split it into two parts, $X^p, Y^p$.  Here each part of $X^p,Y^p$ has the same sample size as the corresponding part of $X^2,Y^2$. Evaluate the accuracy of $f^*$ on this permuted data, call this $a^p$.}  \\[-.8em]
	\item Sort $a^*,a^1,...,a^P$ and denote their order statistics by $a^{(1)} \leq \ldots \leq a^{(P+1)}$; Let $k \defn \lceil(1-\alpha)(1+P) \rceil$. If $a^\ast > a^{(k)}$, then reject the null.
\end{itemize}



\noindent \textbf{Method 2 (Full-permutation):} 
\begin{itemize}
	\item Split data into two halves, $X^1, Y^1$ and $X^2, Y^2$. Train the classifier on $X^1,Y^1$, call it $f^*$. Evaluate accuracy of $f^*$ on $X^2, Y^2$, call it $a^*$. \\[-.8em]
	\item Repeat $P$ times: \textit{Pool all samples $X^1,Y^1,X^2,Y^2$ into one bag, randomly permute the samples, and then split it into 4 parts $X^p,Y^p,X'^p,Y'^p$. Here each part of $X^p,Y^p,X'^p,Y'^p$ has the same sample size as the corresponding part of $X^1,Y^1,X^2,Y^2$. Train a new classifier $f^p$ on the first half, evaluate it on the second half, to get accuracy $a^p$.} \\[-.8em]
	\item Sort $a^*,a^1,...,a^P$ and denote their order statistics by $a^{(1)} \leq \ldots \leq a^{(P+1)}$. Let $k \defn \lceil(1-\alpha)(1+P) \rceil$. If $a^\ast > a^{(k)}$, then reject the null.
\end{itemize}



It is worth noting that both methods yield a valid level $\alpha$ test under $H_0: \mathbb{P}_0 = \mathbb{P}_1$ as a direct consequence of, for example, Theorem 1 in \cite{hemerik2018false}. In terms of power, method 2 may potentially be more powerful than method 1 as it uses the data more efficiently to determine a threshold. In particular, permuted accuracies via method 1 can take fewer values than those via method 2, which may result in a more conservative threshold depending on the nominal level. However, method 1 has a computational advantage over method 2  since it only requires to re-fit a classifier on the second half of the dataset. Nevertheless the following theorem shows that both methods provide a consistent test under the same assumptions made in Proposition~\ref{Proposition: Asymptotic test}. Let us denote the permutation test by $\varphi_{\widehat{C},\text{Perm}}$ via either method 1 or method 2 based on classifier $\widehat{C}$. 
\begin{theorem} \label{Theorem: permutation test}
	Consider the same assumptions made in Proposition~\ref{Proposition: Asymptotic test}. Then under the null hypothesis $H_0: \mathbb{P}_0 = \mathbb{P}_1$, we have $ \E_{H_0} \big[ \varphi_{\widehat{C},\text{\emph{Perm}}}\big] \leq \alpha$ for each $n$ and $d$. Under the alternative hypothesis $H_1: \mathbb{P}_0 \neq \mathbb{P}_1$, the (half or full) permutation test is consistent as $\lim_{n \rightarrow \infty} \E_{H_1} \big[\varphi_{\widehat{C},\text{\emph{Perm}}}\big] = 1$ given that the number of random permutations $P$ is greater than $(1-\alpha)/\alpha$.  
\end{theorem}
One interesting aspect of the above theorem is that consistency is guaranteed as long as the number of random permutations $P$ is greater than $(1-\alpha)/\alpha$ (e.g., $P \geq 20$ for $\alpha=0.05$), which is independent of the sample size. We would also like to point out that the permutation test relies on a data-dependent threshold and thus it is more difficult to analyze than the asymptotic test. In Appendix~\ref{sec: Proof of Theorem: permutation test}, we bound this data-dependent threshold with a more tractable quantity using Markov's inequality with the first two moments of the permuted test statistic. Leveraging this preliminary result, we prove that the permutation critical value cannot exceed the true accuracy in the limit, and  this is the critical fact that completes the proof.

\section{Experiments} \label{sec: experiments}

In this section, we present several numerical results that support our theoretical analysis. Throughout our simulations (except in Section~\ref{sec: Asymptotic power of Hotelling's Test}), we set the sample sizes and the dimension to be $n_0 = n_1 = d = 200$ and compare two multivariate Gaussian or multivariate $t$-distributions with the same identity covariance matrix, with means
\begin{align*}
\mu_0 = (0,\ldots,0)^\top \quad \text{and} \quad \mu_1 = \frac{\delta}{d^{1/4}} \cdot (1,\ldots,1)^\top
\end{align*}
for $\delta \in \{0,0.05,\ldots,0.35,0.40\}$. The simulations were repeated 500 times to estimate the power of each test at significance level $\alpha = 0.05$.

\subsection{Empirical power vs. theoretical power} \label{Sec: Empirical Power vs. Theoretical Power}
In the following experiment, we compare the empirical power of classification accuracy tests with the corresponding theoretical power. For the Gaussian case, we consider the accuracy tests $\varphi_{\Sigma^{-1}}$ and $\varphi_{\widehat{D}^{-1}}$ based on the Fisher's LDA classifier and the naive Bayes classifier, respectively. As specified in the definitions of $\varphi_{\Sigma^{-1}}$ and $\varphi_{\widehat{D}^{-1}}$, the critical values of both tests are based on a normal approximation. Here we split the samples into training and test sets with equal sample sizes so that the power is asymptotically maximized. In this case, the asymptotic power expression for each test is presented in (\ref{Eq: Power of Fisher's LDA}) and (\ref{eq: power of naive Bayes classifier}), respectively. For the case of multivariate $t$-distributions, we focus on the accuracy test $\varphi_{\Sigma^{-1}}$ and see whether the asymptotic power expression (\ref{Eq: asymptotic power for elliptical distributions}) approximates its empirical power over different values of degrees of freedom $\nu$.

\begin{figure}[t!]
	\begin{center}		
		\begin{minipage}[b]{0.495\textwidth}
			\includegraphics[width=\textwidth]{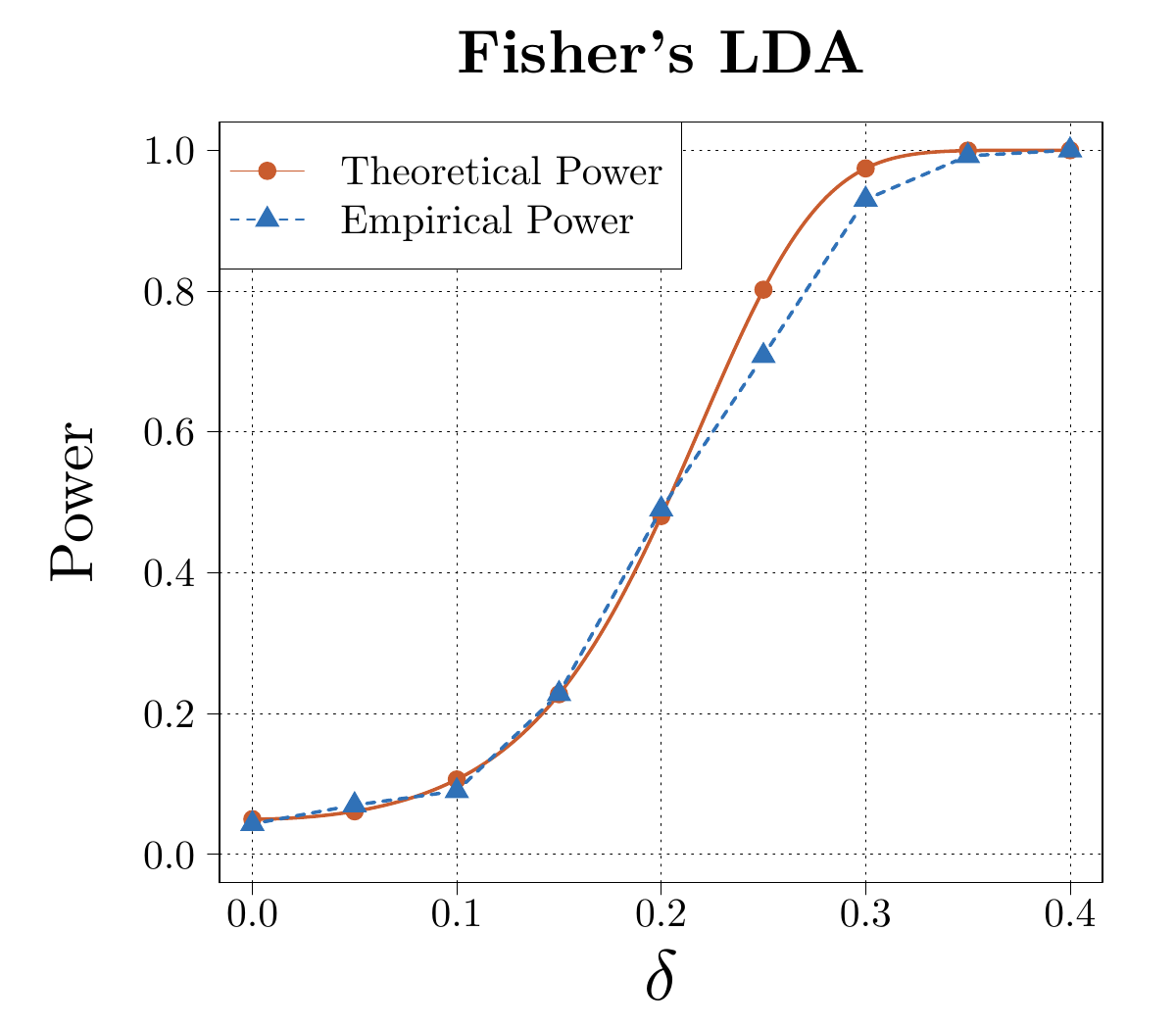}
		\end{minipage} 
		\begin{minipage}[b]{0.495\textwidth}
			\includegraphics[width=\textwidth]{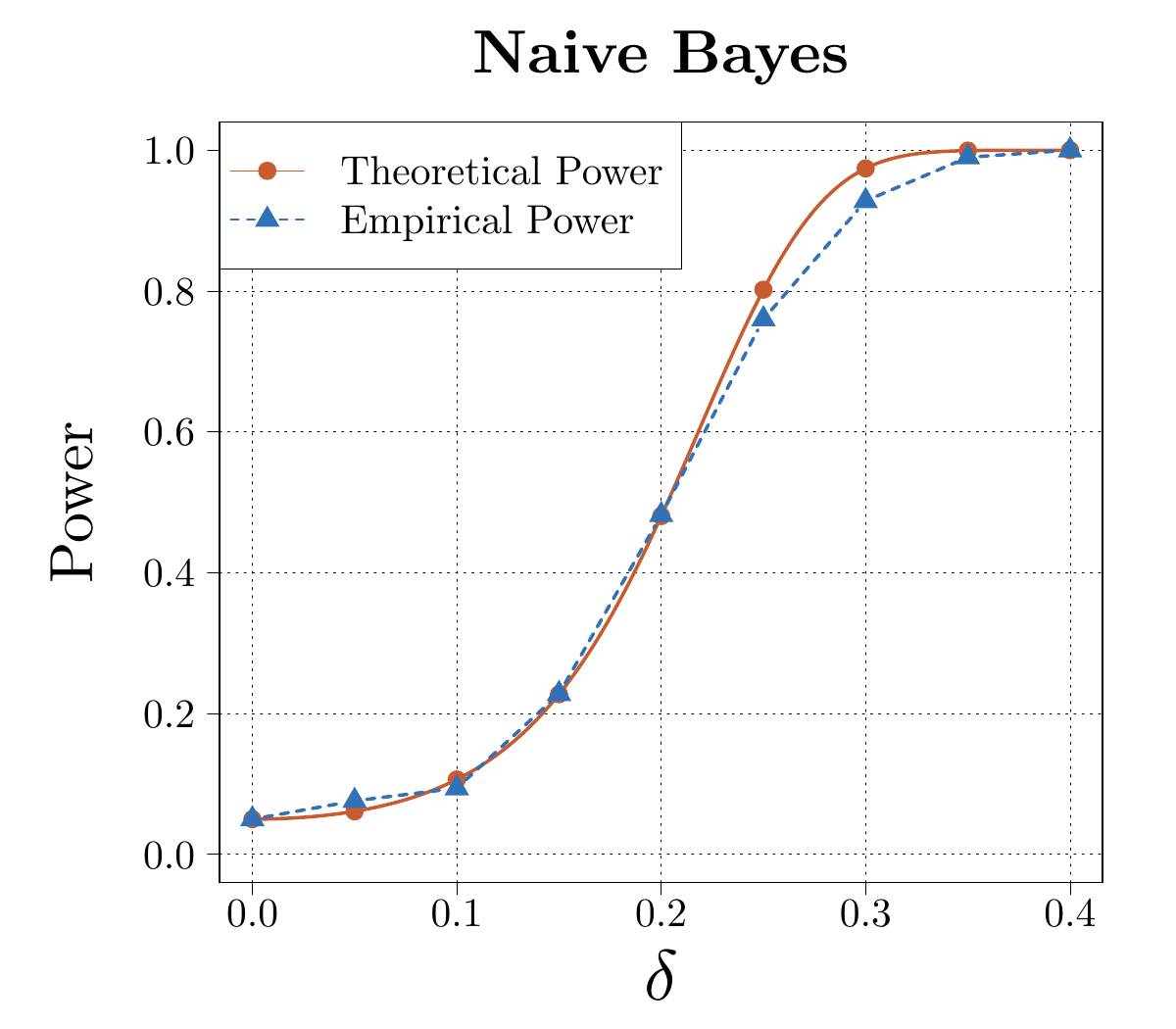}
		\end{minipage}
		\caption{\small Comparisons of the empirical power to our theoretically derived expression for (asymptotic) power under the Gaussian setting. The curves are almost identical especially when the size of $\delta$ is not too big, which suggests that our theory under local alternatives accurately predicts power. See Section~\ref{Sec: Empirical Power vs. Theoretical Power} for details.} \label{Figure: Study1}
	\end{center}
\end{figure}

\begin{figure}[t!]
	\begin{center}		
		\begin{minipage}[b]{0.495\textwidth}
			\includegraphics[width=\textwidth]{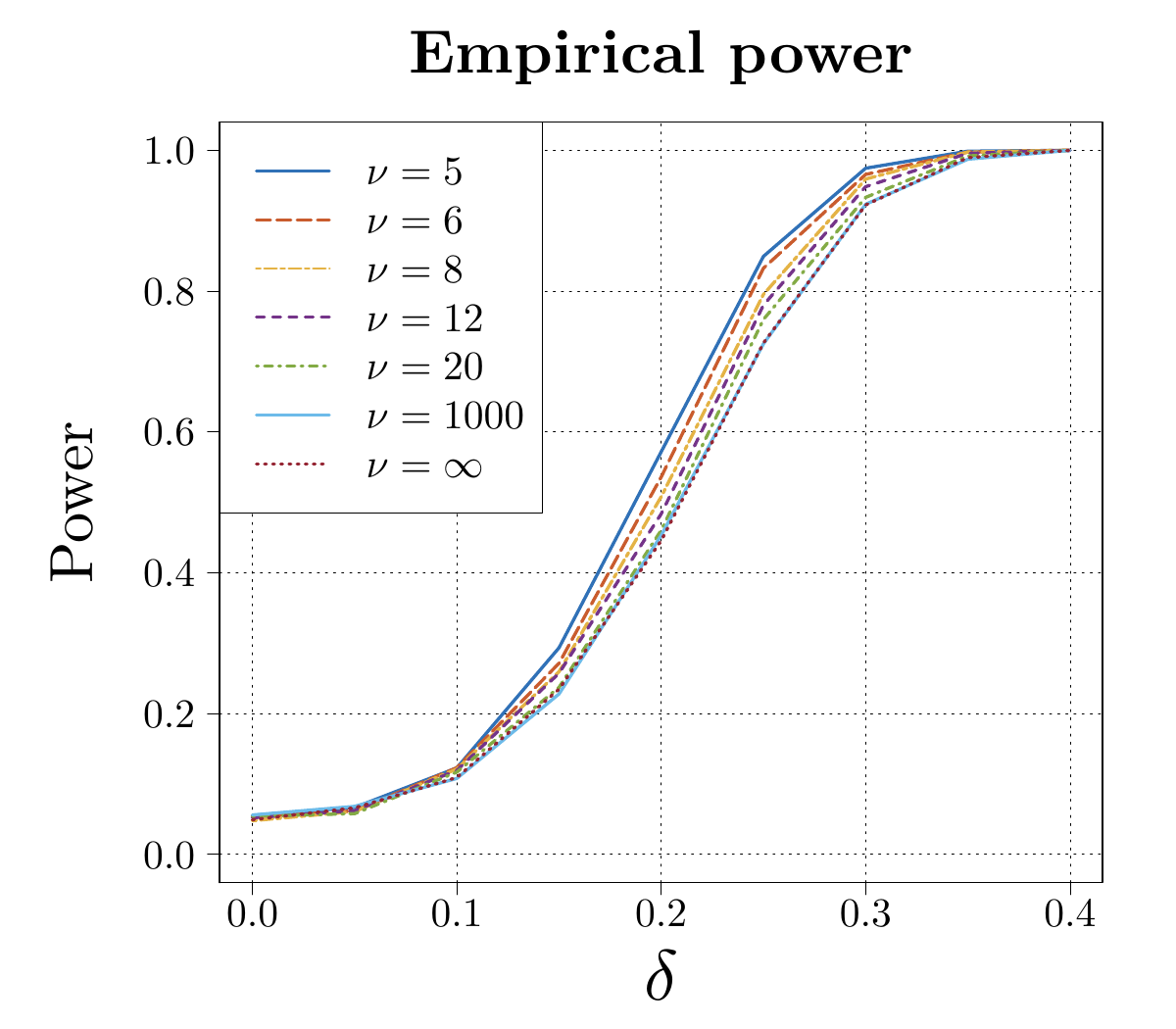}
		\end{minipage} 
		\begin{minipage}[b]{0.495\textwidth}
			\includegraphics[width=\textwidth]{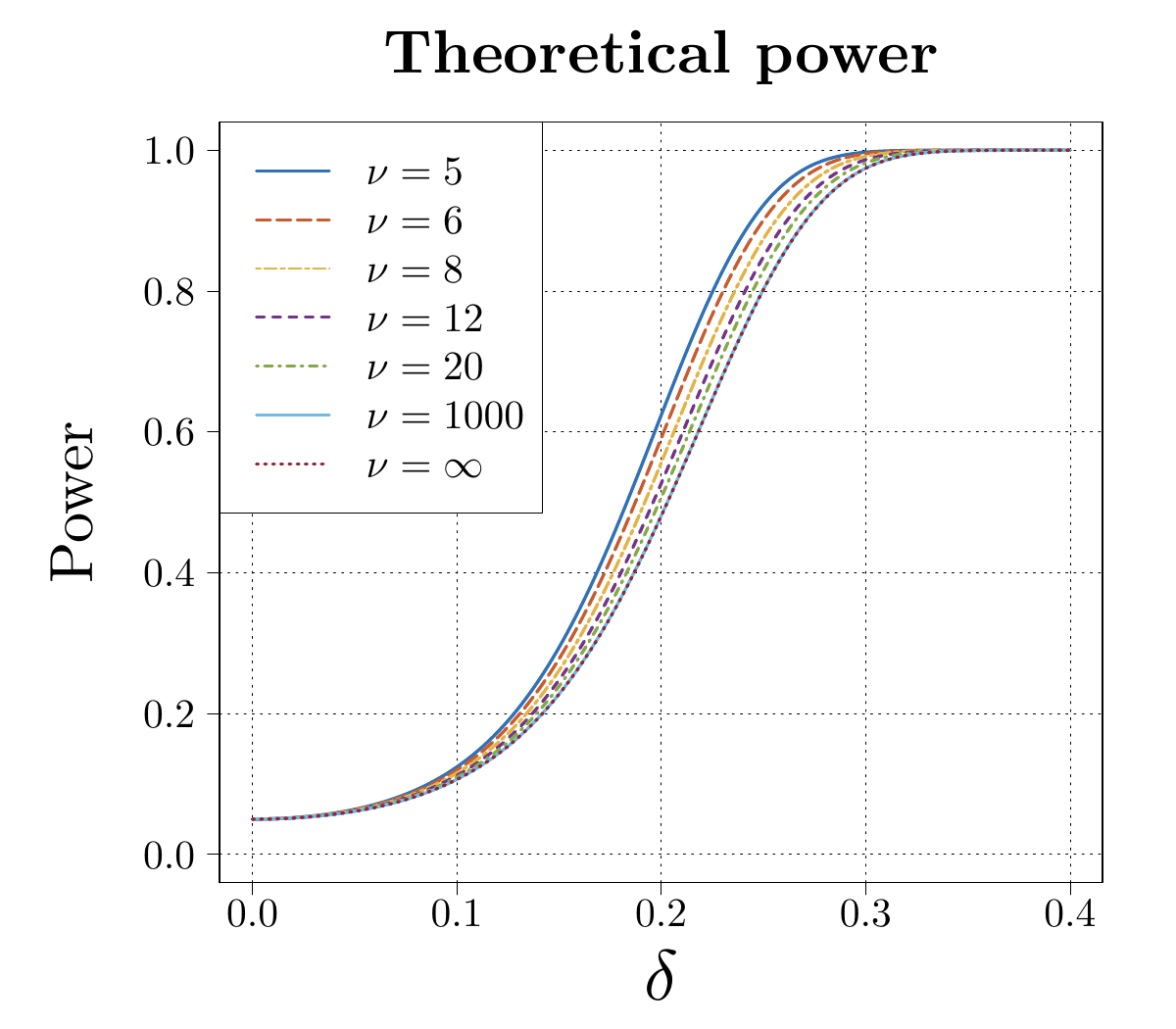}
		\end{minipage}
		\caption{\small The empirical power and theoretical (asymptotic) power of the accuracy test based on Fisher's LDA classifier for comparing multivariate $t$-distributions with $\nu$ degrees of freedom. The curves are tightly matched across $\nu$. Moreover, predicted by Theorem~\ref{thm: power under elliptical distributions}, the power \emph{decreases} with $\nu$. See Section~\ref{Sec: Empirical Power vs. Theoretical Power} for details.} \label{Figure: theory and practice}
	\end{center}
\end{figure}

The results are given in Figure~\ref{Figure: Study1} and Figure~\ref{Figure: theory and practice}. We see that the empirical power almost coincides with the theoretical counterpart especially when $\delta$ is not too big (i.e.~low SNR regime), which confirms our theoretical analysis. We also see that the accuracy test has higher power when the underlying $t$-distributions have smaller degrees of freedom, an interesting and initially surprising fact that is again predicted by our theory.

\subsection{Sample-splitting vs. resubstitution} \label{Sec: Sample-Splitting vs. Resubstitution}
\begin{figure}[t!]
	\begin{center}		
		\begin{minipage}[b]{0.495\textwidth}
			\includegraphics[width=\textwidth]{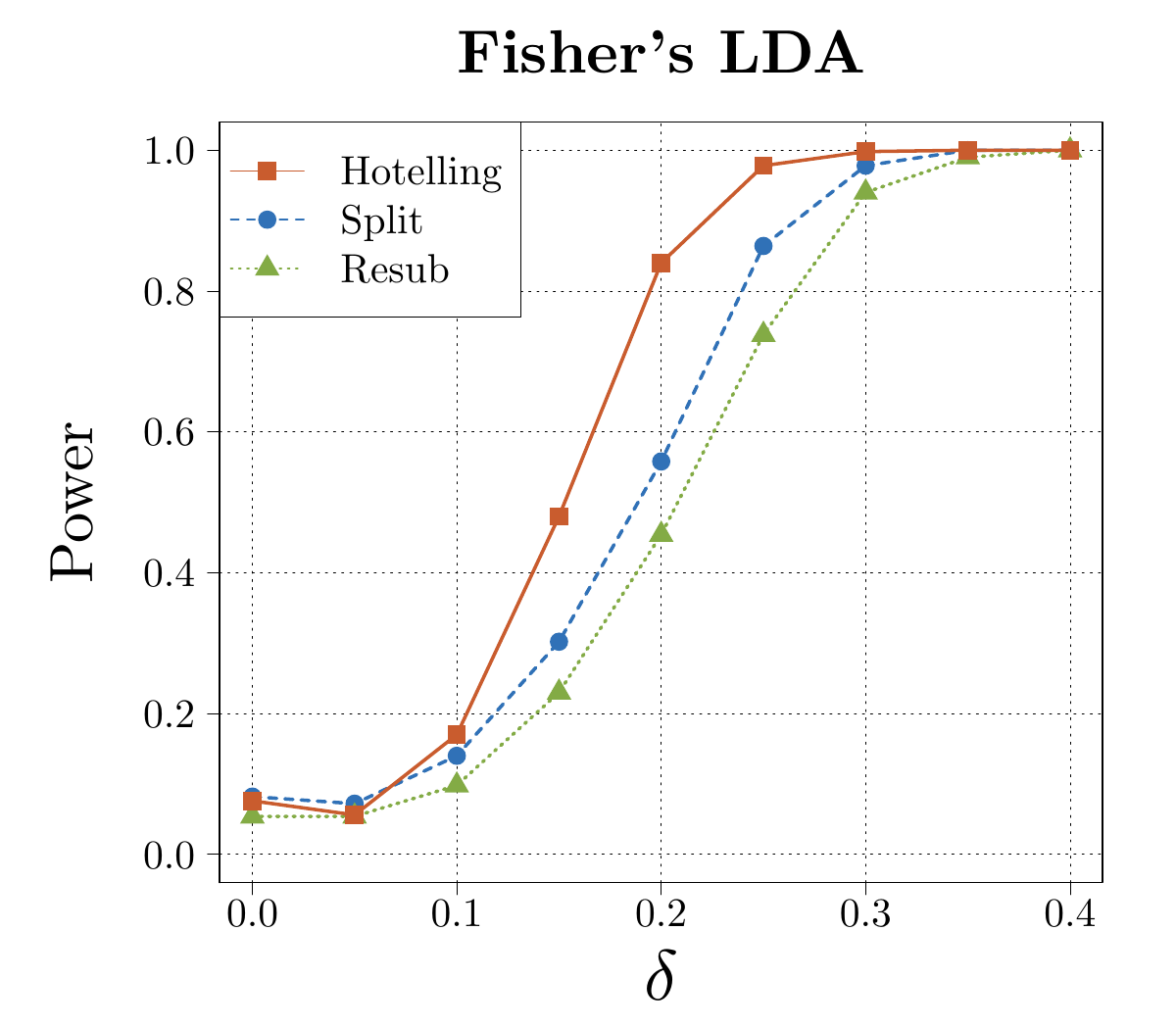}
		\end{minipage} 
		\begin{minipage}[b]{0.495\textwidth}
			\includegraphics[width=\textwidth]{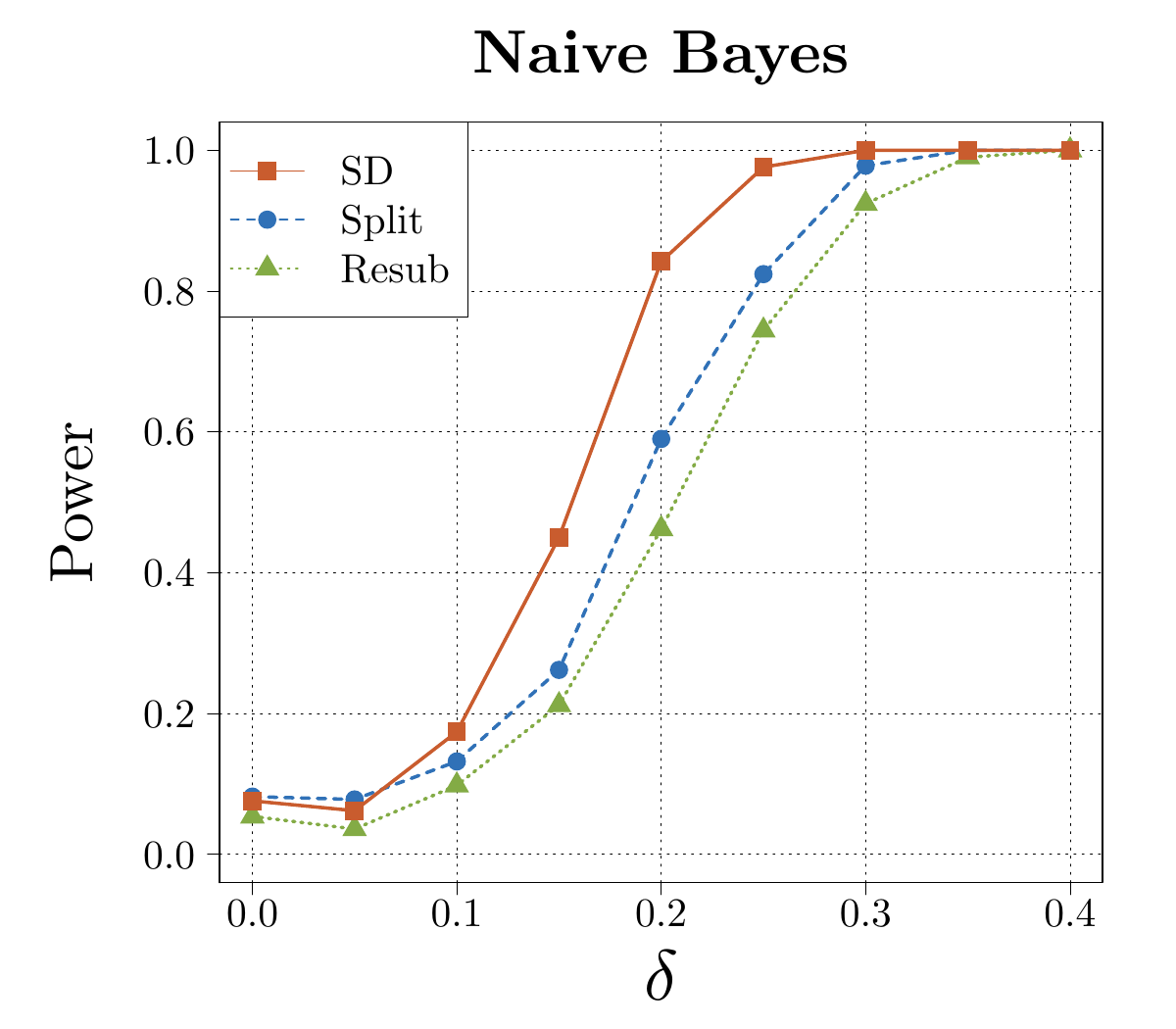}
		\end{minipage}
		\caption{\small Comparisons between sample-splitting (Split) and resubstitution (Resub) tests using LDA and naive Bayes. As reference points, we also consider Hotelling's test and  the test based on $T_{SD}$. Under the given scenarios, the sample-splitting tests have higher power than the resubstitution tests but lower power than Hotelling's and SD tests, the latter being predicted by our theory. See Section~\ref{Sec: Sample-Splitting vs. Resubstitution} for details.} \label{Figure: Study2}
	\end{center}
\end{figure}

In the following experiment, we compare the performance of sample-splitting tests with resubstitution accuracy tests under the Gaussian setting. As their name suggests, the resubstitution accuracy tests use resubstitution accuracy estimates as their test statistic. The precise definition of a resubstitution estimate is given in Appendix~\ref{sec:disc}. We also consider Hotelling's test and its variant proposed by \cite{sd} as reference points. The setup is almost the same as the previous experiment except for the choice of critical values. In particular, since the (asymptotic) null distribution of a resubstitution statistic is unknown, the critical values of all tests are determined using permutations for a fair comparison. Specifically,  to calibrate critical values, we use the full permutation method from Section~\ref{sec: permutation} with 200 random permutations.

In the first part, Fisher's LDA is considered as a base line classifier. Then the accuracy is estimated via (i) sample-splitting with $n_{\text{tr}}=n_{\text{te}}$ and (ii) resubstitution. As a reference point, we consider Hotelling's test as it shares the same weight matrix with Fisher's LDA. For both Hotelling's and Fisher's LDA tests, we assume that $\Sigma$ is known. In the second part, the naive Bayes classifier is considered as a base line classifier with unknown $\Sigma$. We then perform tests based on sample-splitting and resubstitution accuracy statistics defined similarly as before. In this part, we consider $T_{SD}$ given in (\ref{Eq: SD statistic}) as a reference point since it relies on the inverse of diagonal sample covariance matrix as in the naive Bayes classifier. 

From the results presented in Figure~\ref{Figure: Study2}, it stands out that Hotelling's test and its high-dimensional variant are more powerful than the corresponding tests via classification accuracy as we expected. The results also show that the powers of the sample-splitting tests are slightly higher than those of the resubstitution tests in both Fisher's LDA and naive Bayes classifier examples. 
However additional simulation studies, not presented here, suggest that resubstitution tests tend to be more powerful than sample-splitting tests in low-dimensional settings (or when the sample sizes are relatively small) and thus, at least empirically, neither of them is strictly better than the other under all scenarios. Similar empirical results were observed by \cite{rosenblatt2016better} where they conducted extensive simulation studies to compare the performance of the accuracy tests via resubstitution and 4-fold cross-validation and different versions of Hotelling's test. From their simulation results, one reaches the same conclusion that the accuracy tests tend to have lower power than Hotelling's test against Gaussian mean shift alternatives.

\subsection{Asymptotic power of Hotelling's Test} \label{sec: Asymptotic power of Hotelling's Test}

\begin{figure}[t!]
	\centering
	\begin{subfigure}[b]{0.31\textwidth}
		\includegraphics[width=\textwidth]{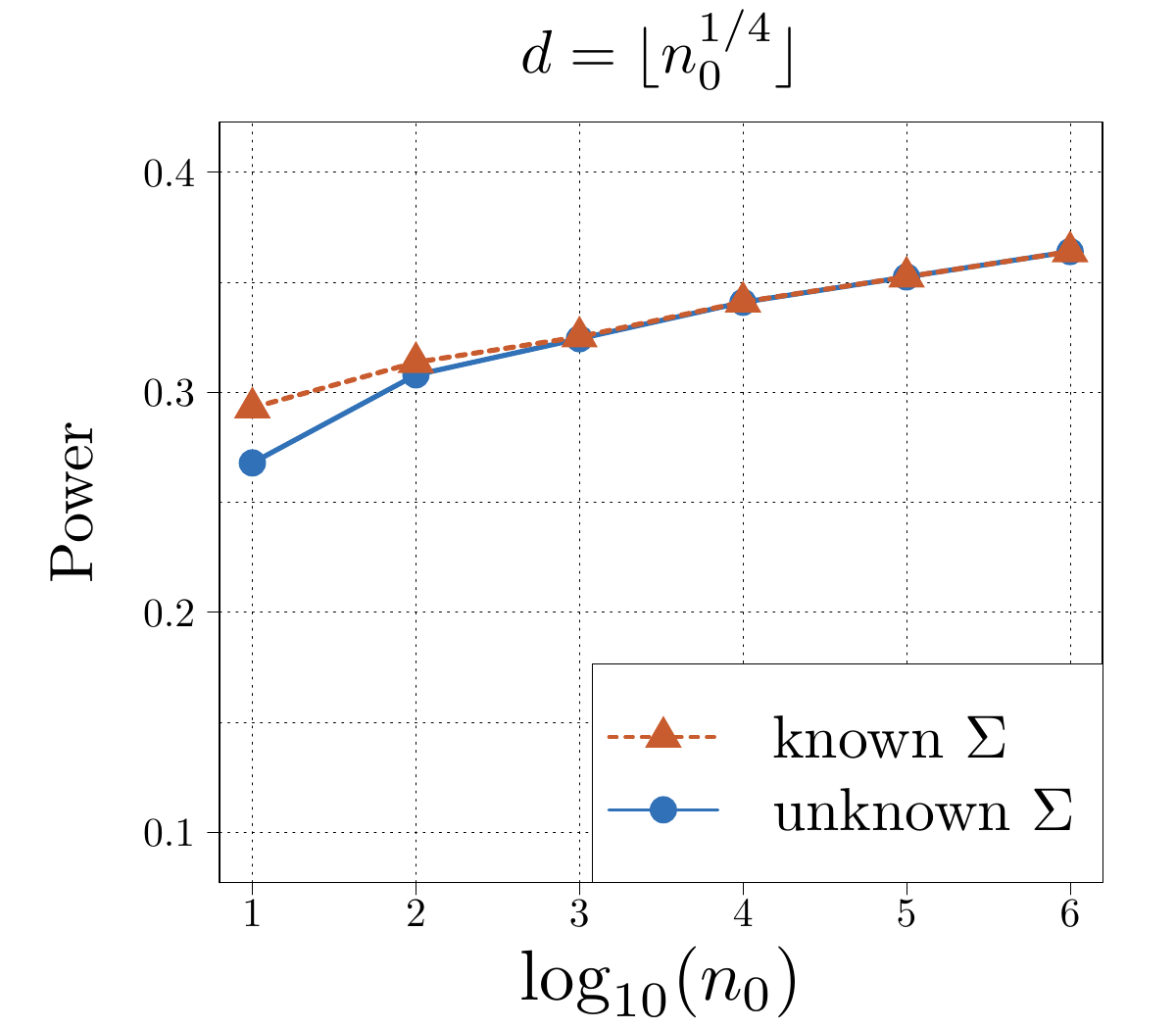}
	\end{subfigure}
	~ 
	\begin{subfigure}[b]{0.31\textwidth}
		\includegraphics[width=\textwidth]{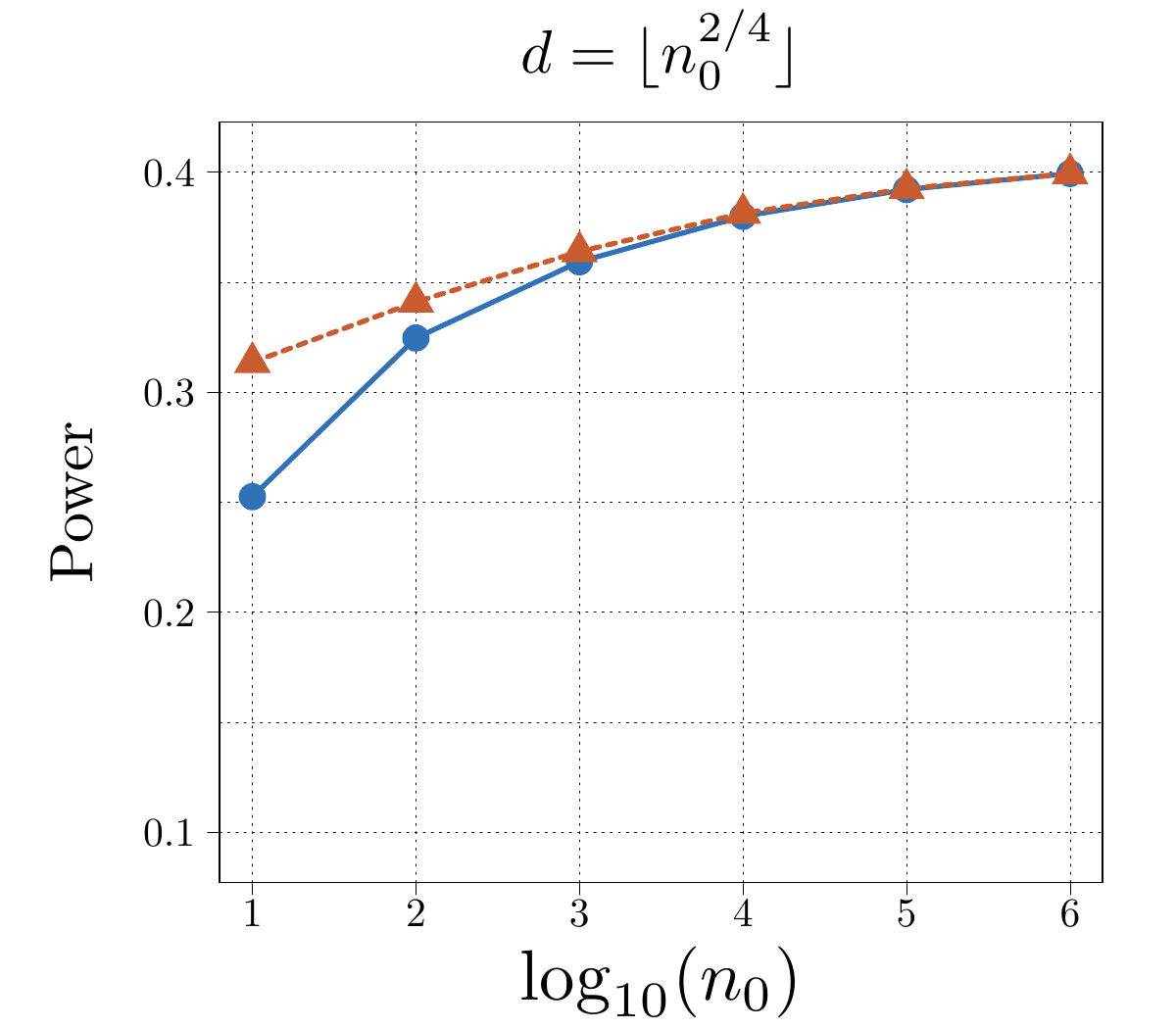}
	\end{subfigure}
	~ 
	\begin{subfigure}[b]{0.31\textwidth}
		\includegraphics[width=\textwidth]{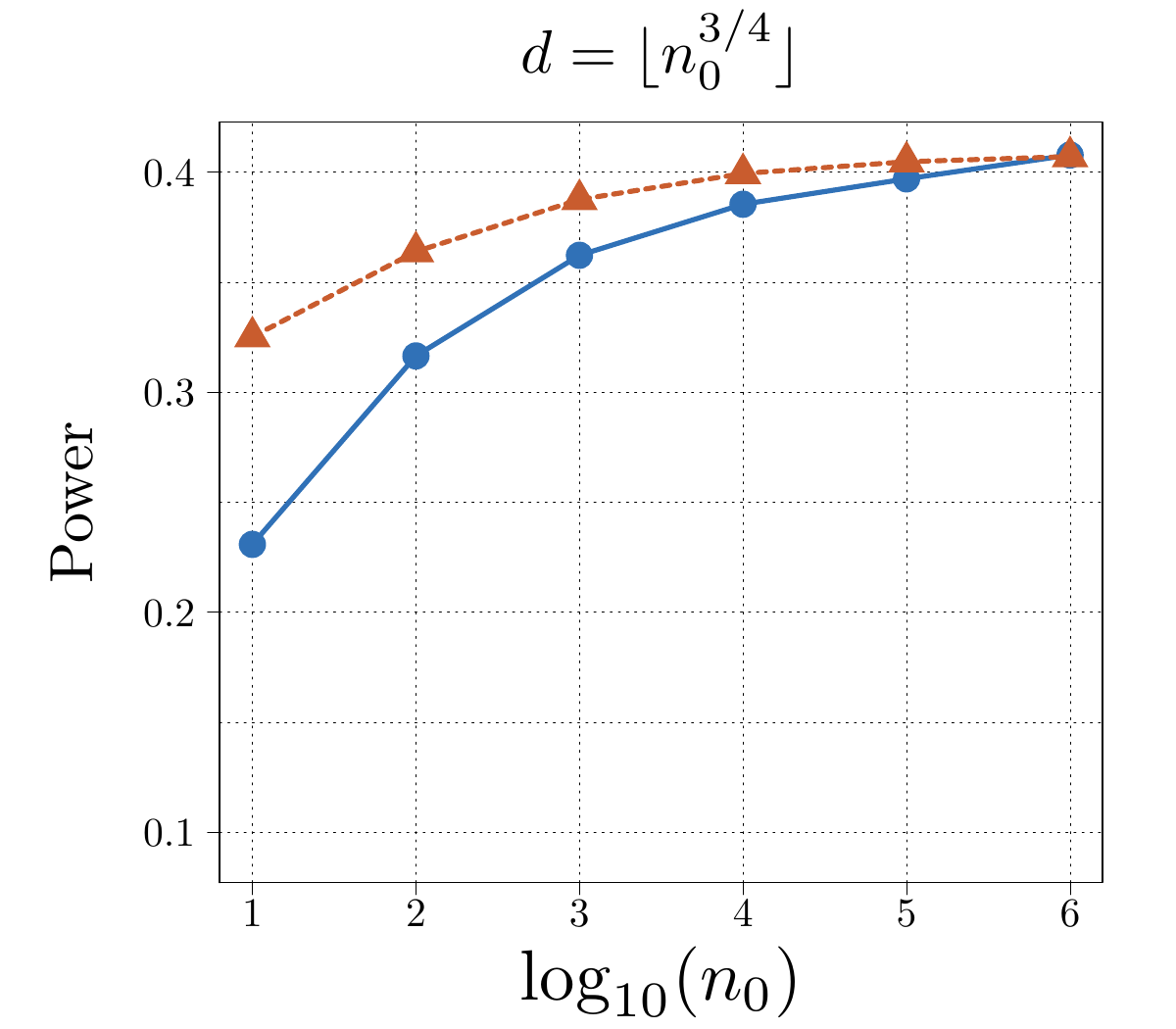}
	\end{subfigure}
	\vskip 1em
	\begin{subfigure}[b]{0.31\textwidth}
		\includegraphics[width=\textwidth]{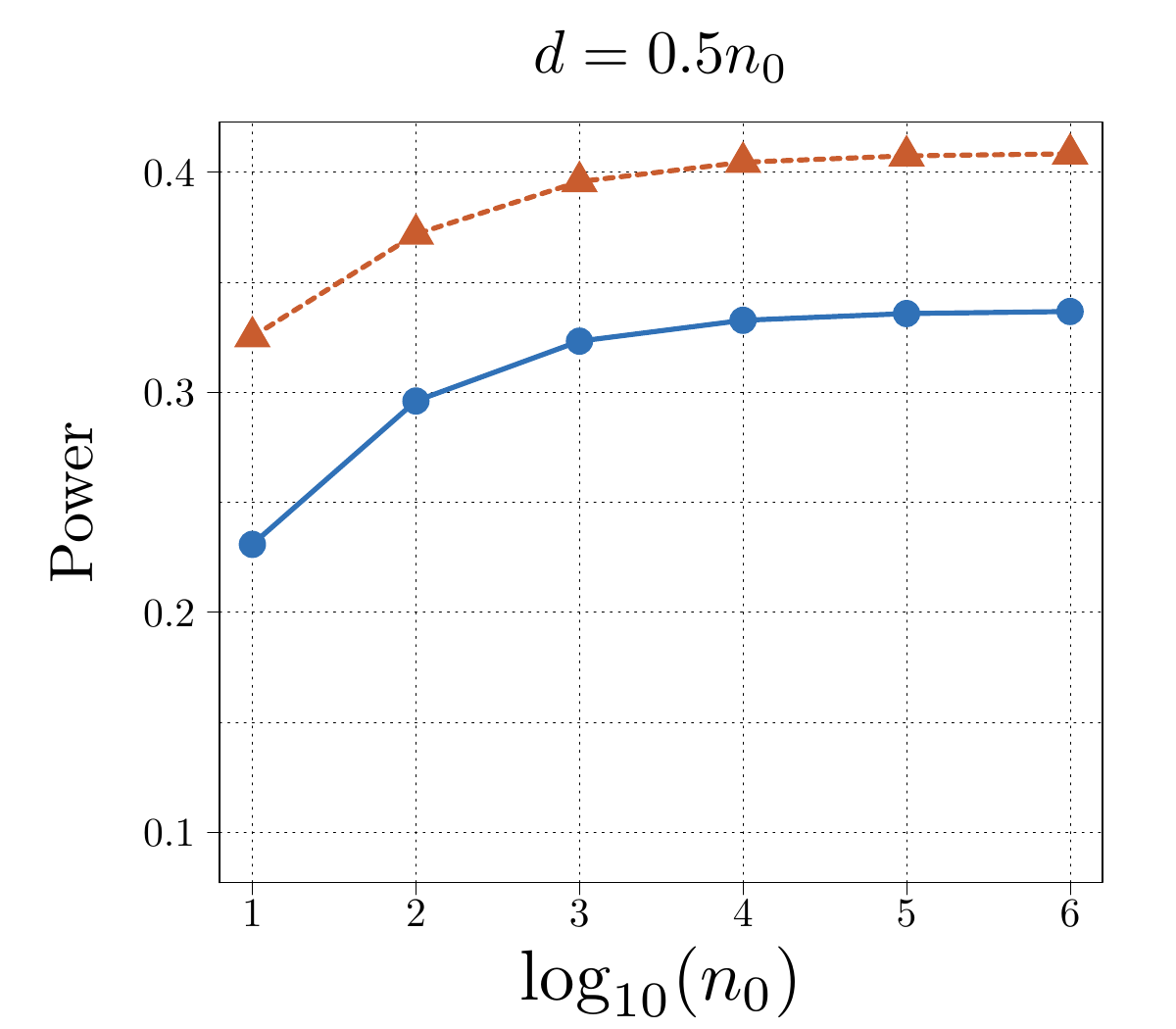}
	\end{subfigure}
	~ 
	\begin{subfigure}[b]{0.31\textwidth}
		\includegraphics[width=\textwidth]{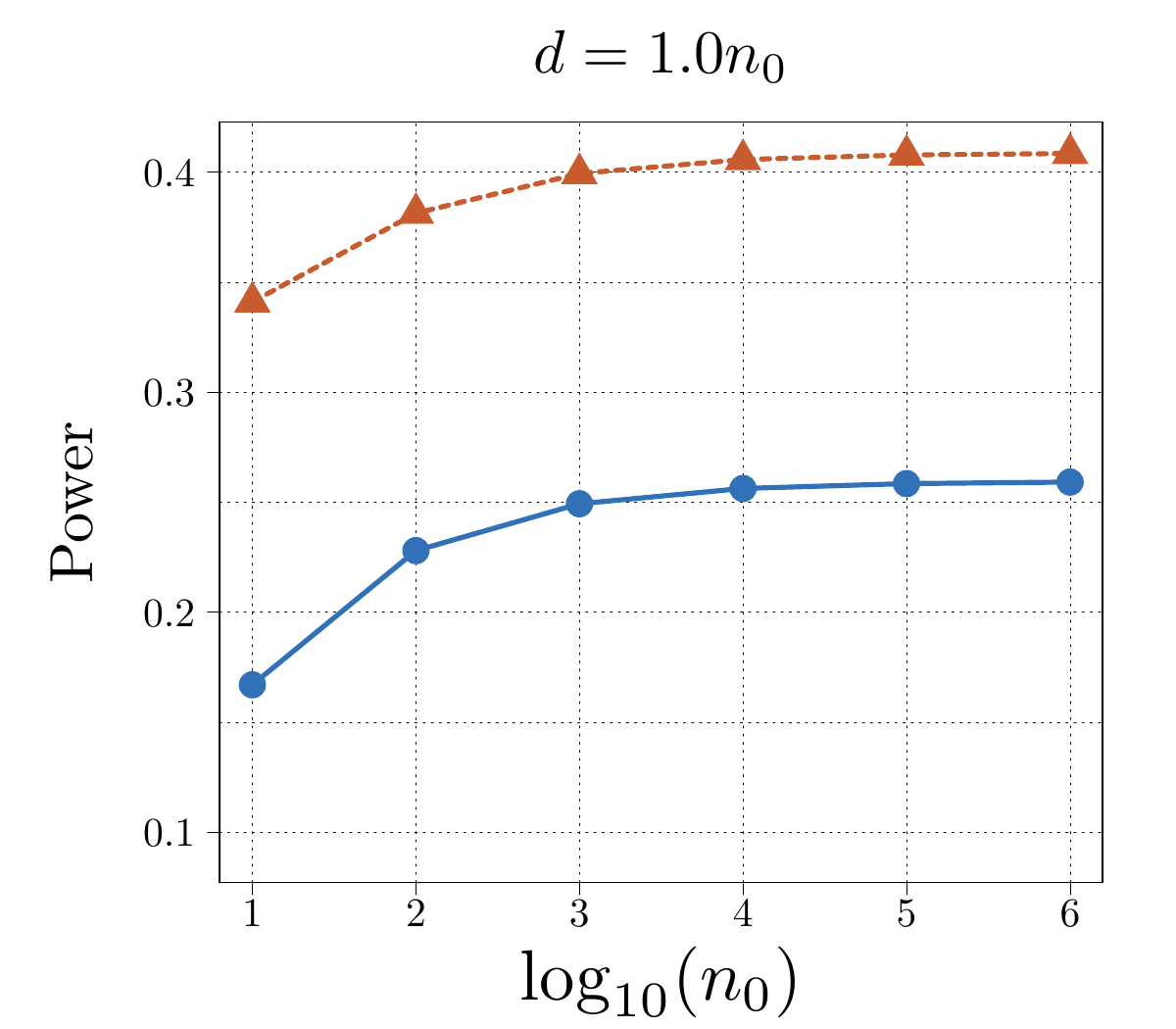}
	\end{subfigure}
	~ 
	\begin{subfigure}[b]{0.31\textwidth}
		\includegraphics[width=\textwidth]{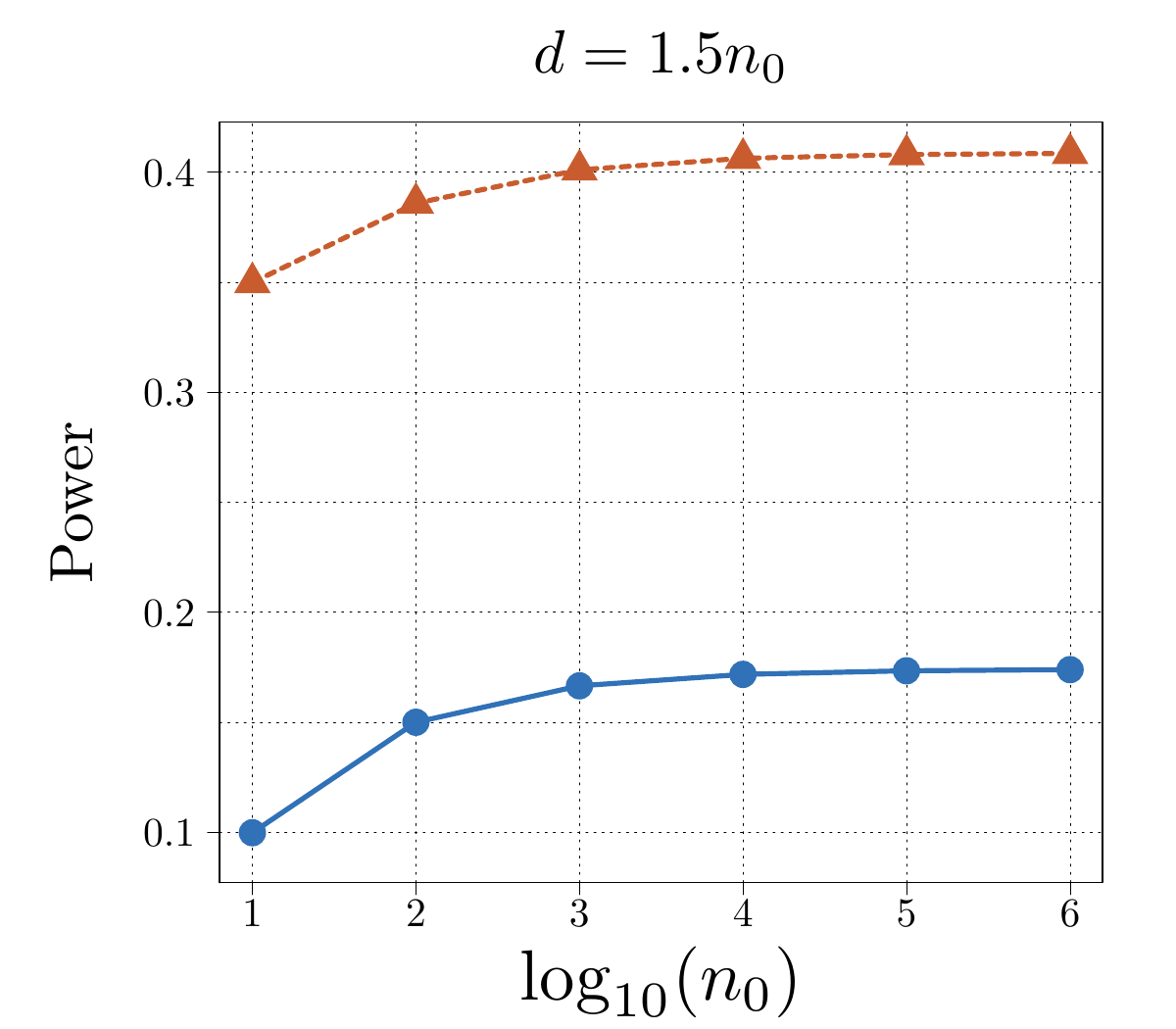}
	\end{subfigure}
	\caption{\small Comparisons of the power of 1) Hotelling's test $\varphi_{H}$ with unknown $\Sigma$ and 2) Hotelling's test $\varphi_{H}^\ast$ with known $\Sigma$ at $\alpha = 0.05$ in different asymptotic regimes. These results coincide with our theoretical results in Section~\ref{sec: Minimax Optimality of Hotelling's Test with unknown Sigma}, showing that $\varphi_{H}$ has asymptotically the same power as $\varphi_{H}^\ast$ when $d/n \rightarrow 0$ (first row) and it is less powerful when $d/n \rightarrow c  \in (0,1)$ (second row). See Section~\ref{sec: Asymptotic power of Hotelling's Test} for details.}\label{Figure: Power of Hotelling Test}
\end{figure}

In this subsection, we provide numerical support for the asymptotic optimality of Hotelling's test under Gaussian settings with unknown $\Sigma$ (Theorem~\ref{Theorem: minimax power of Hotelling's test with unknown Sigma}). Here we compare two multivariate Gaussian distributions with the mean vectors 
\begin{align*}
\mu_0 = \frac{1}{d^{1/4}n_0^{1/2}} \cdot (1,\ldots,1)^\top \quad \text{and} \quad \mu_1 = -\frac{1}{d^{1/4}n_0^{1/2}} \cdot (1,\ldots,1)^\top
\end{align*}
and the identity covariance matrix. In this case, by setting $n_0=n_1$, the asymptotic minimax power tends to be constant as in (\ref{eq:lowSNR}). Now we consider six different asymptotic regimes: i) $d = \lfloor n_0^{1/4} \rfloor$, ii) $d = \lfloor n_0^{2/4} \rfloor$, iii) $d = \lfloor n_0^{3/4} \rfloor$, iv) $d = 0.5 n_0$, v) $d = 1.0 n_0$ and vi) $d = 1.5 n_0$. According to Theorem~\ref{Theorem: minimax power of Hotelling's test with unknown Sigma}, Hotelling's test with unknown $\Sigma$ (denoted by $\varphi_H$) obtains asymptotically the same power as the minimax optimal test (denoted by $\varphi_H^\ast$) in the first three regimes. Whereas, in the last three regimes where $d$ and $n$ are linearly comparable, $\varphi_H$ becomes less powerful than $\varphi_H^\ast$ proved by \cite{bs}. To illustrate this numerically, we increase the sample size by $n_0 \in \{ 10^1, 10^2,\ldots,10^6 \}$ and compute the power of $\varphi_{H}^\ast$ and $\varphi_{H}$ for each $n_0$. To calculate the power, we use the fact that $\mathbb{E}[1-\varphi_{H}^\ast]$ and $\mathbb{E}[1-\varphi_{H}]$ are noncentral $\chi^2$ and $F$ distribution functions evaluated at their critical values, which are $c_{\alpha,d}$ and $q_{\alpha,n,d}$ respectively.

As can be seen in the first row of Figure~\ref{Figure: Power of Hotelling Test}, the power of $\varphi_{H}$ becomes approximately the same as that of $\varphi_{H}^\ast$ in the first three regimes as $n$ increases. 
On the other hand, in the last three regimes where $d/n\rightarrow c \in (0,1)$, we observe significantly different results. Specifically, from the second row of Figure~\ref{Figure: Power of Hotelling Test}, it is seen that the power of $\varphi_{H}$ is much lower than that of $\varphi_{H}^\ast$ and the gap does not decrease even in large $n$. This, thereby, supports our argument that $\varphi_{H}$ is asymptotically comparable to the minimax optimal test in the case of $d/n \rightarrow 0$, but it is underpowered otherwise. 


\section{Conclusion}\label{sec:conc}

This paper provided analyses on the use of classification accuracy as a test statistic for two-sample testing. We started by presenting a fundamental minimax lower bound for high-dimensional two-sample mean testing and showed that Hotelling's test with unknown $\Sigma$ can be optimal in high-dimensional settings as long as $d/n \rightarrow 0$. When $d=O(n)$, we found that two-sample tests via the classification accuracy of various versions of Fisher's LDA (including naive Bayes) have the same power as high-dimensional versions of Hotelling's test in terms of all problem parameters $(n,d,\delta,\Sigma)$, but having worse (but explicit) constants. 

Beyond linear classifiers, we also proved that both the asymptotic test and the permutation test based on a general classifier are consistent if the limiting value of the true accuracy is higher than chance. This consistency result naturally motivated a more challenging setting in which the Bayes error approaches half while the corresponding accuracy-based test can still have non-trivial power, which is the regime studied in most of this paper. Under such a challenging regime, it would be interesting to see whether explicit expressions of power can be derived for non-linear classifiers. Characterizing the high-dimensional power (beyond consistency as we have shown) of permutation-based tests is also an important open problem.





\subsection*{Acknowledgements}
We thank the AE and anonymous referees for their valuable comments that significantly improved the paper. We also thank Arthur Gretton, Leila Wehbe, Amit Datta, Sivaraman Balakrishnan and Eugene Katsevich for their helpful discussions and feedback.


\begin{supplement}
	\stitle{Supplement to ``Classification accuracy as a proxy for two-sample testing''}
		\slink[url]{URL}
	\sdescription{This supplemental file includes the technical proofs omitted in the main text and a discussion on open problems.}
\end{supplement}

\bibliographystyle{apalike}
\bibliography{reference}

\clearpage 

\allowdisplaybreaks

\begin{center}
	\textbf{SUPPLEMENTARY MATERIALS}
\end{center}

\appendix

\section{Outline of supplementary material} 
This supplementary material is organized as follows. In Section~\ref{sec:disc}, we discuss some open problems, raised by our main results. Section~\ref{sec:preliminaries} contains some lemmas that will prove useful in many of the proofs. In Section~\ref{sec: proof of proposition: minimax lower bound}, we provide the proof of Proposition~\ref{Proposition: minimax lower bound}, which shows the asymptotic expression for the minimax power. Section~\ref{sec: proof of theorem: minimax power of Hotelling test with unknown Sigma} presents the proof of Theorem~\ref{Theorem: minimax power of Hotelling's test with unknown Sigma}, which demonstrates the optimality of Hotelling's $T^2$ test when $d/n \rightarrow 0$. Section~\ref{sec: proof of proposition: conditional CLT} focuses on Proposition~\ref{proposition: conditional CLT} and proves the asymptotic normality of $W_A$. In Section~\ref{sec: proof of thm: asymptotic distribution and thm: naive Bayes}, Theorem~\ref{thm: asymptotic distribution} and Theorem~\ref{thm: naive Bayes} are proved, verifying the asymptotic normality of $W_A^\dagger$ and the asymptotic power of the naive Bayes classifier test. Section~\ref{sec: proof of lemma: E approximations} proves Lemma~\ref{lemma: E approximations}. By building on some moment expressions for (scaled) inverse chi-square random variables in Section~\ref{sec: Some moments of (scaled) inverse chi-square random variables}, we provide the proof of Lemma~\ref{lemma: Dinv approximations} in Section~\ref{sec: Lemma Dinv}. Section~\ref{sec: power under elliptical distributions} provides the proof of Theorem~\ref{thm: power under elliptical distributions}, which is an extension of our main result to elliptical distributions. In Section~\ref{sec: proof of Proposition: Asymptotic test} and Section~\ref{sec: Proof of Theorem: permutation test}, we prove the type-1 error control and consistency result of the asymptotic test and the permutation test, respectively. Lastly, some simulation results on sample-splitting ratio are presented in Section~\ref{sec: Sample-Splitting Ratio}.

\section{Open problems} \label{sec:disc}

Here we discuss how our results may be extended to a larger context while we leave a detailed analysis to future work. Four open problems that we first highlight are as follows:
\begin{itemize}
	\item The most obvious open problem is to extend our power guarantees, and most other published ones in the high-dimensional two-sample testing literature, to be uniform over an entire class of alternative distributions rather than just holding pointwise. Viewing our proofs in this material, uniform control of the relevant error terms seems extremely challenging. \\[-.5em]
	\item Determining whether our minimax lower bound can be achieved by any test when $d=O(n)$, or if tighter lower bounds can be proved, is an important open problem. Of course, we have settled this problem for $d=o(n)$ in Section~\ref{sec: Minimax Optimality of Hotelling's Test with unknown Sigma} even from the perspective of uniformity.\\[-.5em]
	\item Given that the focus of this study is mainly on Fisher's LDA classifier and its variants, there is a possibility that some other linear discrimination rules (e.g.~via empirical risk minimization) may achieve optimal power.   \\[-.5em]
	\item Beyond the consistency result, proving that one can achieve the same non-trivial power as the asymptotic tests using permutations seems like an interesting open problem.
\end{itemize}
From the perspective of the title of the current paper, we provide other four natural directions for future explorations.

\paragraph{Leave-one-out accuracy}
Another natural estimator for accuracy, as an alternative to sample-splitting, is a leave-one-out estimator $\widehat E^L$, defined as  $\widehat E^L \defn (\widehat E^L_0 + \widehat E^L_1)/2$, with
\begin{eqnarray}
\label{eq:errL} 
\widehat E^L_0 &\defn& \frac1{n_{0}}\sum_{i=1}^{n_{0}} \I \Big[ \LDA_{n_0 \backslash i, n_1 }(X_{i}) = 1 \Big], \\
\widehat E^L_1 &\defn& \frac1{n_1}\sum_{i=1}^{n_1} \I \Big[ \LDA_{n_{0},n_{1} \backslash  i}(Y_{i}) = 0 \Big], \nonumber
\end{eqnarray}
where $\LDA_{n_0 \backslash i,n_1}$ (or $\LDA_{n_0,n_1\backslash i}$) denotes the LDA classifier using all points except $X_i$ (or $Y_i$). 

\paragraph{Ensemble  accuracy}
The sample-splitting estimator $\widehat{E}^S$ in (\ref{eq:errS}) is based on an arbitrary split in training and test sets. Hence the resulting test is potentially unstable depending on the result of sample splitting. This issue can be simply overcome by considering all possible splits. Let $\sigma\defn \{\sigma(1),\ldots,\sigma(n_{0,\text{tr}}) \}$ be a subset of $\{1,\ldots,n_0\}$ drawn without replacement. Similarly, let $\sigma^\prime\defn \{\sigma^\prime(1),\ldots,\sigma^\prime(n_{1,\text{tr}}) \}$ be a subset of $\{1,\ldots,n_1\}$ drawn without replacement. By setting $\{X_{\sigma(1)},\ldots, X_{\sigma(n_{0,\text{tr}})} \} \cup \{Y_{\sigma^\prime(1)},\ldots, Y_{\sigma^\prime(n_{1,\text{tr}})} \}$ as the training set and the remaining as the test set, one can calculate $\widehat{E}^S\defn \widehat{E}^S(\sigma,\sigma^\prime)$. The ensemble estimator is then defined by
\begin{align*}
\widehat{E}^{Ens} = \frac{1}{\binom{n_0}{n_{0,\text{tr}}} \binom{n_1}{n_{1,\text{tr}}}} \sum_{\sigma}\sum_{\sigma^\prime} \widehat{E}^S(\sigma,\sigma^\prime),
\end{align*}
where the first sum is taken over all possible subsets of size $n_{0,\text{tr}}$ from $\{1,\ldots,n_0\}$ and the second sum is taken over all possible subsets of size $n_{1,\text{tr}}$ from $\{1,\ldots,n_1\}$.  Although it looks similar to the $U$-statistic considered in \cite{hediger2019use}, the ensemble estimator differs from theirs by allowing $\widehat{E}^S(\sigma,\sigma^\prime)$ to be a function of the entire dataset rather than a subset. Hence the proposed one uses the dataset more efficiently.

\paragraph{Resubstitution accuracy}
Since leave-one-out estimators and ensemble estimators are computationally intensive, one might be tempted to use the training data itself to test the  classifier. This resubstitution error would be defined as $\widehat E^R \defn (\widehat E^R_0 + \widehat E^R_1)/2$, with
\begin{eqnarray}\label{eq:resub}
\label{eq:errR} 
\widehat E^R_0 &\defn& \frac1{n_{0}}\sum_{i=1}^{n_{0}} \I \Big[ \LDA_{n_0, n_1 }(X_{i}) = 1 \Big],\\
\widehat E^R_1 &\defn& \frac1{n_1}\sum_{i=1}^{n_1} \I \Big[ \LDA_{n_{0},n_{1}}(Y_{i}) = 0 \Big], \nonumber
\end{eqnarray}
where we first train on all the data and then test on all the data. Of course such an estimate would be overoptimistic, and would be scorned upon as an estimate of the true accuracy $E$ of the classifier. However, one might hope that the null distribution or permutation distribution would be similarly optimistically biased (instead of being centered around a half), thus nullifying the optimistic bias of $\widehat E^R$. From simulation studies, we observed that the accuracy test based on the resubstitution error performs slightly better than the test based on sample splitting in low-dimensional scenarios but overall performs similarly (e.g.~Figure \ref{Figure: Study2}). It will be interesting to theoretically justify the asymptotic behavior of resubstitution accuracy (and also leave-one-out and ensemble accuracy) and see whether the resulting test is also minimax rate optimal.


\paragraph{Non-linear classification}
Another natural setting is that of nonlinear classification. An examination of the test statistics used (Hotelling and its variants) shows that they are closely related to the statistics based on the kernel Maximum Mean Discrepancy \citep{mmd12} and the kernel FDA \citep{kfda}, when specifically instantiated with the linear kernel. Similarly, for classification, a kernelized LDA \citep{mika1999fisher}  specializes to Fisher's LDA when the linear kernel is employed.

Given the parallels observed, and given that a kernel classifier or two-sample test is effectively a linear method in a higher dimensional space, one might naturally conjecture that the spirit of the results of this paper can be extended to such kernelized nonlinear settings as well. As mentioned before, very recent progress has been made by \cite{hediger2019use} for random forests (but not in the high dimensional setting). 

The use of neural network type classifiers for classifier-based testing on structured data is certainly an interesting direction, though precise theoretical characterizations, such as the ones provided in this paper, seem unlikely given our current understanding.

\section{Technical proofs}
\subsection{Supporting lemmas} \label{sec:preliminaries}

Before we present the detailed proofs of all our results, we collect some supporting lemmas. The first lemma provides the mean and variance of a quadratic form of Gaussian random vectors. 

\begin{lemma}[Chapter 5.2 in \cite{rencher2008linear}] \label{lemma: quadratic form}
	Suppose that $Z$ has a multivariate Gaussian distribution with mean $\mu$ and covariance $\Sigma$. Then, we have
	\begin{align*}
	& \E[Z^\top \Lambda Z] = \text{\emph{tr}}[\Lambda \Sigma] + \mu^\top \Lambda \mu  \ \text{and} \\[.5em]
	& \mathbb{V}[Z^\top \Lambda Z]  = 2\text{\emph{tr}}[\Lambda \Sigma\Lambda \Sigma] + 4\mu^\top \Lambda \Sigma \Lambda \mu.
	\end{align*}
\end{lemma}

\noindent Next we present the Berry-Esseen theorem for non-identically distributed summands, which will be used to prove Proposition~\ref{proposition: conditional CLT}.
\begin{lemma}[Berry-Esseen theorem, \cite{berry1941accuracy}] \label{lemma: berry-esseen}
	Let $X_1,X_2,\ldots,$ be independent random variables with $\E[X_i] =0$, $\E[X_i^2] = \sigma_i^2 >0$ and $\E[|X_i|^3] = \rho_i <\infty$. Define
	$S_n = \frac{\sum_{i=1}^n X_i}{\sqrt{\sum_{i=1}^n \sigma_i^2}}$
	and let $F_n(\cdot)$ be its CDF. Then there exists a constant $c > 0$ such that
	\begin{align*}
	\sup_{t \in \mathbb{R}} | F_n(t) - \Phi(t) | \leq  \frac{c}{\left( \sum_{i=1}^n \sigma_i^2 \right)^{1/2}} \max_{1\leq i \leq n} \frac{\rho_i}{\sigma_i^2}.
	\end{align*}
\end{lemma}


\noindent The following lemma bounds the trace of a product of two matrices in terms of their eigenvalues.

\begin{lemma}[Fan's inequality, page 10 of \cite{borwein2010convex}] \label{lemma: fan's inequality}
	For any symmetric matrices $A,B \in \mathbb{R}^{d \times d}$, we have 
	$	\emph{\tr} (AB) \leq \sum_{i=1}^d \lambda_i(A) \lambda_i(B).$
\end{lemma}

\vskip 0.8em

\noindent Before stating the next two lemmas, let us recall some notation from the main text. First $\mathcal{E}_{0,A}$ and $\mathcal{E}_{1,A}$ are the errors of the generalized LDA conditional on the input data. These can be written as $\mathcal{E}_{0,A} = \Phi ( {V_{0,A}}/{\sqrt{U_A}} )$ and $\mathcal{E}_{1,A} = \Phi ( {V_{1,A}}/{\sqrt{U_A}} )$ where $V_{0,A} = \widehat{\delta}^\top A (\mu_0 - \widehat{\mu}_{\text{pool}})$, $V_{1,A} = \widehat{\delta}^\top A (\widehat{\mu}_{\text{pool}} -\mu_1) $ and $U_A = \widehat{\delta}^\top A \Sigma A \widehat{\delta}$.
Further recall $\Psi_{A,n,d} = -\delta^\top A \delta / 2$, $\Lambda_{A,n,d} = \delta^\top A\Sigma A \delta + \left( 1/{n_{0,\text{tr}}} + 1/{n_{1,\text{tr}}} \right) \text{tr} \{  (A\Sigma)^2 \}$ and $\Xi_{A,n,d}  = ( n_{0,\text{tr}}^{-1} - n_{1,\text{tr}}^{-1})\tr(A\Sigma)/2$.

\vskip .8em

\noindent The following lemma presents approximations of $\mathcal{E}_{0,A}$, $\mathcal{E}_{1,A}$ and $\mathcal{E}_{0,A} + \mathcal{E}_{1,A}$, which plays a key role in proving Theorem~\ref{thm: asymptotic distribution}.
\begin{lemma} \label{lemma: E approximations}
	Under the assumptions \textbf{(A1)}--\textbf{(A6)}, $\mathcal{E}_{0,A}$, $\mathcal{E}_{1,A}$ and $\mathcal{E}_{0,A}+\mathcal{E}_{1,A}$ are
	\begin{align*}
	&\mathcal{E}_{0,A} ~=~ \Phi \Bigg(  \frac{\Psi_{A,n,d} + \Xi_{A,n,d}}{\sqrt{ \Lambda_{A,n,d}}} \Bigg) + O_P\left(n^{-1/2} \right), \\[.5em]
	&\mathcal{E}_{1,A} ~=~ \Phi \Bigg(  \frac{\Psi_{A,n,d} - \Xi_{A,n,d}}{\sqrt{ \Lambda_{A,n,d}}} \Bigg) + O_P\left(n^{-1/2} \right) \ \text{and} \\[.5em]
	&\mathcal{E}_{0,A} + \mathcal{E}_{1,A} ~= ~ \Phi \Bigg(  \frac{\Psi_{A,n,d} + \Xi_{A,n,d}}{\sqrt{ \Lambda_{A,n,d}}} \Bigg)  + \Phi \Bigg(  \frac{\Psi_{A,n,d} - \Xi_{A,n,d}}{\sqrt{ \Lambda_{A,n,d}}} \Bigg)  + o_P\left(n^{-1/2} \right).
	\end{align*}
	Furthermore, when $n_{0,\text{\emph{tr}}} = n_{1,\text{\emph{tr}}}$, 
	\begin{align*}
	\mathcal{E}_{0,A} + \mathcal{E}_{1,A} ~=~ 2  \Phi \Bigg(  \frac{\Psi_{A,n,d}}{\sqrt{ \Lambda_{A,n,d}}} \Bigg)  + O_P(n^{-3/4}).
	\end{align*}
\end{lemma}
\noindent One thing to notice from the above lemma is that the sum of $\mathcal{E}_{0,A}$ and $\mathcal{E}_{1,A}$ converges faster than the individual components because the higher order error terms cancel out in the sum. This critical phenomenon allows us to replace $\mathcal{E}_{0,A}/2 + \mathcal{E}_{1,A}/2$ in $W_A$ with a non-random counterpart. The proof of Lemma~\ref{lemma: E approximations} can be found in Section~\ref{sec: proof of lemma: E approximations}.

\vskip 0.8em

\noindent The following lemma is similar to Lemma~\ref{lemma: E approximations} but by replacing a non-random matrix $A$ with random diagonal matrix $\widehat{D}^{-1} = \text{diag}(\widehat{\Sigma})^{-1}$. This lemma will be used to prove Theorem~\ref{thm: naive Bayes} where we present the power of the naive Bayes classifier test. 

\begin{lemma} \label{lemma: Dinv approximations}
	Assume that $n_{0}=n_{1}$, $n_{0,\text{\emph{tr}}}=n_{1,\text{\emph{tr}}}$ and $n_{\text{\emph{tr}}} = n_{\text{\emph{te}}}$. Then under the assumptions \textbf{(A1)}, \textbf{(A2)} and \textbf{(A5)}, $\mathcal{E}_{0,\widehat{D}^{-1}}$, $\mathcal{E}_{1,\widehat{D}^{-1}}$ and $\mathcal{E}_{0,\widehat{D}^{-1}}+\mathcal{E}_{1,\widehat{D}^{-1}}$ are
	\begin{align*}
	&\mathcal{E}_{0,\widehat{D}^{-1}} = \Phi \Bigg(  \frac{\Psi_{D^{-1},n,d}}{\sqrt{ \Lambda_{D^{-1},n,d}  }} \Bigg)  + O_P\left(n^{-1/2} \right), \\[.5em]
	&\mathcal{E}_{1,\widehat{D}^{-1}} = \Phi \Bigg(  \frac{\Psi_{D^{-1},n,d}}{\sqrt{ \Lambda_{D^{-1},n,d}  }} \Bigg)  + O_P\left(n^{-1/2} \right) \ \text{and} \\[.5em]
	&\mathcal{E}_{0,\widehat{D}^{-1}} + \mathcal{E}_{1,\widehat{D}^{-1}} = 2\Phi \Bigg(  \frac{\Psi_{D^{-1},n,d}}{\sqrt{ \Lambda_{D^{-1},n,d}  }} \Bigg)  + O_P\left(n^{-3/4} \right). 
	\end{align*}
\end{lemma}
\noindent As in Lemma~\ref{lemma: E approximations}, due to the cancellation of higher order error terms, we observe that the sum of $\mathcal{E}_{0,\widehat{D}^{-1}}$ and $\mathcal{E}_{1,\widehat{D}^{-1}}$ converges faster than either individual component. The proof of Lemma~\ref{lemma: Dinv approximations} can be found in Section~\ref{sec: Lemma Dinv}.

\vskip 0.8em

\noindent We now have all the results in place to prove the main results in the paper. 

\subsection{Proof of Proposition~\ref{Proposition: minimax lower bound} (minimax lower bound)} \label{sec: proof of proposition: minimax lower bound}
We begin by recalling the result by \cite{luschgy1982minimax} in \eqref{Eq: Lower Bound}, which implies that to derive a bound on the minimax power, one only needs to analyze the power of the oracle Hotelling's procedure $\varphi^*_H$ with known $\Sigma$.
Next, note that $\frac{n_0n_1}{n_0 + n_1}(\widehat{\mu}_0 - \widehat{\mu}_1)^\top \Sigma^{-1} (\widehat{\mu}_0 - \widehat{\mu}_1)$ has a noncentral chi-square distribution with $d$ degrees of freedom and noncentrality parameter $\frac{n_0n_1}{n_0+n_1} (\mu_0 - \mu_1)^\top \Sigma^{-1} (\mu_0 - \mu_1)$. Using the monotonicity of the distribution function of a noncentral chi-square random variable in its non-centrality parameter, it can thus be seen that 
\begin{align} \label{Eq: minimax lower bound}
\sup_{\varphi_\alpha \in \mathcal{T}_\alpha} \inf_{p_0,p_1 \in \mathcal{P}_1(\rho)} \mathbb{E}_{p_0,p_1}[ \varphi_\alpha ]  &=
\inf_{p_0,p_1 \in \mathcal{P}_1(\rho)} \mathbb{E}_{p_0,p_1}[ \varphi_{H}^\ast] = \mathbb{P} \left( \sum_{i=1}^d Z_i^2 \geq c_{\alpha,d} \right),
\end{align} 
where $Z_i \overset{i.i.d.}{\sim} N(\rho_n,1)$ and $\rho_n \defn \sqrt{\frac{n_0n_1}{n_0+n_1}}\rho$. 
Note that the right-hand side of (\ref{Eq: minimax lower bound}) rearranges to
\begin{align*}
\mathbb{P} \left(\frac{\sum_{i=1}^d Z_i^2  - d - \rho_n^2}{\sqrt{2(d+2\rho_n^2)}}  \geq \frac{c_{\alpha,d} - d}{\sqrt{2d}} \frac{\sqrt{2d}}{\sqrt{2(d+2\rho_n^2)}} - \frac{\rho_n^2}{\sqrt{2(d+2\rho_n^2)}} \right).
\end{align*}
By Lyapunov's central limit theorem, we know that
\begin{align} \label{Eq: normal approximation}
\frac{\sum_{i=1}^d Z_i^2  - d - \rho_n^2}{\sqrt{2(d+2\rho_n^2)}} \convD N(0,1) \quad \text{and} \quad \frac{c_{\alpha,d} - d}{\sqrt{2d}} \rightarrow z_{\alpha},
\end{align}
using which the statement of Proposition~\ref{Proposition: minimax lower bound} immediately follows.

\vspace{1em}

\subsection{Proof of Theorem~\ref{Theorem: minimax power of Hotelling's test with unknown Sigma} (optimality of Hotelling's $T^2$ test)} \label{sec: proof of theorem: minimax power of Hotelling test with unknown Sigma}
We first describe a couple of preliminaries and then prove the main theorem. 

\vskip .5em

\paragraph{Preliminaries}  Under the Gaussian setting, it is well-known \citep{anderson58} that
\begin{align*}
\frac{n_0 + n_1 -1-d}{d(n_0 + n_1 -2)}\frac{n_0n_1}{n_0 + n_1}T_H \sim F(d,n-1-d; \rho_n^2),
\end{align*}
where $F(d,n-1-d; \rho_n^2)$ has the non-central $F$-distribution with noncentrality parameter $\rho_n^2=\frac{n_0n_1}{n_0+n_1} (\mu_0 - \mu_1)^\top \Sigma^{-1} (\mu_0 - \mu_1)$ and $d$ and $n-1-d$ degrees of freedom. Let $\chi_d^2 (\rho_n^2)$ be a noncentral chi-square random variable with noncentrality parameter $\rho_n^2$ and $d$ degrees of freedom and write $\chi_{n-1-d}^2(0) = \chi_{n-1-d}^2$ for simplicity. Using the monotonicity of the distribution function of a noncentral $F$ random variable in its non-centrality parameter, it can be seen that
\begin{align*}
\inf_{p_0,p_1 \in \mathcal{P}_1(\rho)} \mathbb{E}_{p_0,p_1}[ \varphi_{H}] = \mathbb{P} \{F(d,n-1-d; \rho_n^2) \geq q_{\alpha,n,d} \}.
\end{align*} 
Hence it is enough to study the asymptotic behavior of the right-hand side of the above equality. Note that the noncentral $F$-distribution can be written in terms of the ratio of two independent chi-square random variables as
\begin{align*}
F(d,n-1-d; \rho_n^2) ~ \overset{d}{=} ~ \frac{\chi_d^2(\rho_n^2)/d}{\chi_{n-1-d}^2/(n-1-d)}.
\end{align*}
For notational convenience, let us write $\mathcal{V}_{n,d} \defn \chi_{n-1-d}^2/(n-1-d)$. Then, by the weak law of large number, it is clear to see that $\mathcal{V}_{n,d} \convP 1$ as $n,d \rightarrow \infty$ with $d/n \rightarrow 0$. 

\vskip 1em

\paragraph{Main proof} Our main strategy to prove the given claim is to split the cases into two: 1) $\rho_n^2/ n \rightarrow 0$ and 2) $\liminf_{n,d \rightarrow \infty} \rho_n^2/n > 0$. In the first case, we shall show that $\chi_d^2 (\rho_n^2)$ and $F(d,n-1-d; \rho_n^2)$ have the same asymptotic distribution after proper studentization. In the second case, we will verify that the power of both tests converge to one.

\vskip 1em

\noindent \textbf{$\bullet$ Case 1.} 
To begin, we assume $\rho_n^2/ n \rightarrow 0$ and prove that
\begin{align} \label{Eq: Aim 1}
\frac{\chi_d^2(\rho_n^2) / \mathcal{V}_{n,d} - d - \rho_n^2}{\sqrt{2d + 4\rho_n^2}} = \frac{\chi_d^2(\rho_n^2) - d - \rho_n^2}{\sqrt{2d + 4\rho_n^2}} + o_P(1) ~ \convD ~ N(0,1).
\end{align}
If (\ref{Eq: Aim 1}) holds, then the result follows since
\begin{align*}
\frac{\frac{n_0 + n_1 -1-d}{n_0 + n_1 - 2}\frac{n_0n_1}{n}T_H - d - \rho_n^2}{\sqrt{2d + 4\rho_n^2}} \convD  N(0,1) \quad \text{and} \quad \frac{dq_{\alpha,n,d} - d}{\sqrt{2d}} \rightarrow z_\alpha.
\end{align*}
To show (\ref{Eq: Aim 1}), note that a simple algebraic manipulation yields 
\begin{align} \label{Eq: F-statistic decomposition}
\frac{\chi_d^2(\rho_n^2) / \mathcal{V}_{n,d} - d - \rho_n^2}{\sqrt{2d + 4\rho_n^2}} 
& ~=~ \frac{1}{\mathcal{V}_{n,d}} \left[ \frac{\chi_d^2(\rho_n^2) - d - \rho_n^2}{\sqrt{2d + 4\rho_n^2}} \right] + \frac{d+\rho_n^2}{\sqrt{2d + 4\rho_n^2}} \left( \frac{1}{\mathcal{V}_{n,d}} - 1 \right).
\end{align}
Using the moments of an inverse chi-square distribution, 
\begin{align*}
& \mathbb{E} \left[ \frac{1}{\mathcal{V}_{n,d}} \right] = \frac{n-1-d}{n-3-d}, \text{ ~ and ~ } 
\mathbb{V} \left[ \frac{1}{\mathcal{V}_{n,d}} \right] = \frac{2(n-1-d)^2}{(n-3-d)^2(n-5-d)},
\end{align*}
one can conclude that
\begin{align*}
\frac{d+\rho_n^2}{\sqrt{2d + 4\rho_n^2}} \left( \frac{1}{\mathcal{V}_{n,d}} - 1 \right) \convP 0. 
\end{align*}
Then the result follows by Slutsky's theorem combined with $\mathcal{V}_{n,d} \convP 1$ and (\ref{Eq: normal approximation}).

\vskip 1em

\noindent \textbf{$\bullet$ Case 2.} In the second case where $\liminf_{n,d \rightarrow \infty} \rho_n^2/n > 0$, there is no guarantee of (\ref{Eq: Aim 1}). Nevertheless, we can show that the power of both tests converge to one when $\liminf_{n,d \rightarrow \infty} \rho_n^2/n > 0$. Since the first term in (\ref{Eq: Power Expression}) is bounded and 
\begin{align*}
\frac{\rho_n^2}{\sqrt{2d + 4\rho_n^2}} \rightarrow \infty,
\end{align*}
we can conclude that the power of $\varphi_{H}^\ast$ converges to one when $\liminf_{n,d \rightarrow \infty} \rho_n^2/n > 0$.

Now we compute the limiting power of the test based on $T_H$. By putting $$r_{n,d} = \frac{n_0 + n_1 -1-d}{n_0 + n_1 - 2}\frac{n_0n_1}{n_0 + n_1},$$
one can note that 
\begin{align*}
& \mathbb{P} \left( \frac{r_{n,d}T_H - d}{\sqrt{2d}}  \geq \frac{q_{\alpha,n,d}-d}{\sqrt{2d}} \right) \\[.5em] 
= ~ & \mathbb{P} \left( \frac{r_{n,d} T_H - \mathbb{E} [r_{n,d} T_H ]}{\sqrt{\mathbb{V}[r_{n,d} T_H ]}} > \frac{q_{\alpha,n,d}-d}{\sqrt{2d}}\sqrt{\frac{2d}{\mathbb{V}[r_{n,d} T_H ]}} + \frac{d - \mathbb{E}[r_{n,d} T_H ]}{\sqrt{\mathbb{V}[r_{n,d} T_H ]}} \right).
\end{align*}
Using the mean and variance formula for a noncentral $F$-distribution, we have
\begin{align*}
\mathbb{E}[r_{n,d} T_H ] ~=~ & \frac{(n-1-d)(d+\rho_n^2)}{n-3-d} = (d+\rho_n^2)\{1 + o(1)\}, \text{ and } \\[.5em]
\mathbb{V}[r_{n,d} T_H ] ~=~ & 2 (n-1-d)^2  \frac{(d+\rho_n^2)^2 + (d+2\rho_n^2)(n-3-d)}{(n-3-d)^2(n-5-d)} \\[.5em]
\asymp ~ & \frac{(d+\rho_n^2)^2}{n} + \frac{d+\rho_n^2}{n^2}.
\end{align*}
From this, we may infer that
\begin{align*}
& \frac{r_{n,d} T_H - \mathbb{E} [r_{n,d} T_H ]}{\sqrt{\mathbb{V}[r_{n,d} T_H ]}} = O_P(1), \ \ \frac{q_{\alpha,n,d}-d}{\sqrt{2d}} \sqrt{\frac{2d}{\mathbb{V}[r_{n,d} T_H ]}} = O(1) \ \ \text{and} \\[.5em] 
& \frac{d - \mathbb{E}[r_{n,d} T_H ]}{\sqrt{\mathbb{V}[r_{n,d} T_H ]}}  \rightarrow - \infty.
\end{align*}
This immediately implies that
\begin{align*}
\liminf_{n,d \rightarrow \infty}  \mathbb{P} \left( \frac{r_{n,d}T_H - d}{\sqrt{2d}}  \geq  \frac{q_{\alpha,n,d}-d}{\sqrt{2d}} \right)  = 1,
\end{align*}
thus completing the proof of Theorem~\ref{Theorem: minimax power of Hotelling's test with unknown Sigma}.

\vskip 2em

\subsection{Proof of Proposition~\ref{proposition: conditional CLT} (asymptotic normality of $W_A$)} \label{sec: proof of proposition: conditional CLT}
As described in the main text, the sample-splitting error of the generalized LDA classifier is an average of independent (but not all identically distributed) random variables when conditioning on the training set. Hence we apply the Berry-Esseen theorem in Lemma~\ref{lemma: berry-esseen} to first establish the conditional central limit theorem for $W_A$. Then we use the bounded convergence theorem to prove the unconditional counterpart. 

\vskip .5em

\paragraph{Conditional Part} Conditional on the training set $\mathcal{T}_{\text{tr}} \defn  \mathcal{X}_1^{n_{0,\text{{tr}}}} \cup \mathcal{Y}_1^{n_{1,\text{{tr}}}}$, $\widehat{E}_A^S$ is the sum of independent random variables. Specifically, 
\begin{align*}
n_{\text{te}} \widehat{E}_A^S &= \sum_{i=1}^{n_{0,\text{te}}} Q_{0,i} + \sum_{i=1}^{n_{1,\text{te}}} Q_{1,i},\\
\text{ where } \quad Q_{0,i} & = \frac{n_{\text{te}}}{2n_{0,\text{te}}} \I \Big[ \LDA_{A,n_{0,\text{tr}},n_{1,\text{tr}}}(X_{n_{0,\text{tr}} + i}) = 1 \Big], \  \\[.5em]
\text{ and } \quad Q_{1,i} & = \frac{n_{\text{te}}}{2n_{1,\text{te}}} \I \Big[ \LDA_{A,n_{0,\text{tr}},n_{1,\text{tr}}}(Y_{n_{1,\text{tr}} + i}) = 0 \Big].
\end{align*}
Notice that for $k=0,1$, we have
\begin{align*}
& \E [ |Q_{k,i} - \E[Q_{k,i}|\mathcal{T}_{\text{tr}}]|^3 | \mathcal{T}_{\text{tr}}] ~ \leq ~ \frac{n_{\text{te}}^3}{8n_{k,\text{te}}^{3}}, \ \text{and} \\[.5em]
& \E [ (Q_{k,i} - \E[Q_{k,i}|\mathcal{T}_{\text{tr}}] )^2 | \mathcal{T}_{\text{tr}}] ~=~ \frac{n_{\text{te}}^2}{4n_{k,\text{te}}^{2}} \mathcal{E}_{1,A}(1-\mathcal{E}_{1,A}).
\end{align*}
We may then apply Lemma~\ref{lemma: berry-esseen} to yield 
\begin{align} \label{eq: berry-esseen}
& \sup_{t \in \mathbb{R}} | \Pr(W_A \leq t | \mathcal{T}_{\text{tr}}) - \Phi(t) |  \leq c a_{n,1} a_{n,2} a_{n,3},\\
\text{ where }\quad 
a_{n,1} &~=~ \bigg\{\frac{n_{\text{te}}^2}{4n_{0,\text{te}}} \mathcal{E}_{0,A} (1 - \mathcal{E}_{0,A}) + \frac{n_{\text{te}}^2}{4n_{1,\text{te}}} \mathcal{E}_{1,A} (1 - \mathcal{E}_{1,A}) \bigg\}^{-1/2}, \nonumber \\[.5em]
a_{n,2} &~=~ \frac{n_{\text{te}}^3}{8n_{0,\text{te}}^3} + \frac{n_{\text{te}}^3}{8n_{1,\text{te}}^3}, \  \nonumber \\[.5em]
\text{ and } \quad  a_{n,3} &~=~ \frac{4n_{\text{0,te}}^2}{ n_{\text{te}}^2 \mathcal{E}_{0,A} (1 - \mathcal{E}_{0,A})}+  \frac{4n_{\text{1,te}}^2}{ n_{\text{te}}^2 \mathcal{E}_{1,A} (1 - \mathcal{E}_{1,A})}. \nonumber 
\end{align}
Under the eigenvalue conditions for $A$ and $\Sigma$ in \textbf{(A5)} and \textbf{(A6)}, one can find constants $c,c'>0$ such that $c \leq \tr\{(A\Sigma)^2\} /d \leq c'$ and $c \leq \tr(A\Sigma) /d \leq c'$ due to $d \lambda_{\text{min}}^2(A) \lambda_{\text{min}}^2(A)  \leq \tr\{(A\Sigma)^2\} \leq d \lambda_{\text{max}}^2(A) \lambda_{\text{max}}^2(A)$ and $d \lambda_{\text{min}}(A) \lambda_{\text{min}}(A)  \leq \tr(A\Sigma) \leq d \lambda_{\text{max}}(A) \lambda_{\text{max}}(A)$. Then under \textbf{(A1)}--\textbf{(A4)}, there exists another constant $c'' > 0$ such that 
$- c'' \leq (\Psi_{A,n,d} + \Xi_{A,n,d}) /  \sqrt{\Lambda_{A,n,d}} \leq c''$ and $- c'' \leq (\Psi_{A,n,d} - \Xi_{A,n,d}) /  \sqrt{\Lambda_{A,n,d}} \leq c''$ for large $n$. Therefore both $\Phi\{(\Psi_{A,n,d} +\Xi_{A,n,d})/\sqrt{\Lambda_{A,n,d}}\}$ and $\Phi \{ (\Psi_{A,n,d} -\Xi_{A,n,d})/\sqrt{\Lambda_{A,n,d}}\}$ are strictly bounded below by zero and above by one for large $n$. Based on this observation together with Lemma~\ref{lemma: E approximations}, it can be seen that $a_{n,1} = O_P(n^{-1/2})$, $a_{n,2} = O(1)$ and $a_{n,3} = O_P(1)$. Thus the right-hand side of (\ref{eq: berry-esseen}) is $O_P(n^{-1/2})$, which completes the proof of the conditional part.

\vskip 1em 

\paragraph{Unconditional Part} For each $t \in \mathbb{R}$, the previous result gives $\Pr(W_A \leq t |  \mathcal{X}_1^{n_{0,\text{{tr}}}}, \mathcal{Y}_1^{n_{1,{\text{tr}}}} ) - \Phi(t)=o_P(1)$. We then apply the bounded convergence theorem to have $\Pr(W_A \leq t ) - \Phi(t)=o(1)$. Since $\Phi(\cdot)$ is continuous, Polya's theorem yields the final result \citep[e.g. Lemma 2.11 of][]{van2000asymptotic}. This completes the proof of Proposition~\ref{proposition: conditional CLT}.

\vskip 2em

\subsection{Proof of Theorem~\ref{thm: asymptotic distribution} and \ref{thm: naive Bayes}} \label{sec: proof of thm: asymptotic distribution and thm: naive Bayes}

\paragraph{Proof of Theorem~\ref{thm: asymptotic distribution} (Asymptotic normality of $W_A^\dagger$)}
Based on Lemma~\ref{lemma: E approximations} and the facts that $\Psi_{A,n,d} = O(n^{-1/2})$, $\Xi_{A,n,d} = O(1)$ and $\liminf_{n,d \rightarrow \infty} \Lambda_{A,n,d} >0$ (see the proof of Proposition~\ref{proposition: conditional CLT} for details), we have
\begin{align} \nonumber
\mathcal{E}_{0,A}(1 - \mathcal{E}_{0,A}) & = \Phi \Bigg(  \frac{\Psi_{A,n,d} + \Xi_{A,n,d}}{\sqrt{ \Lambda_{A,n,d}}} \Bigg) \Bigg\{ 1-  \Phi \Bigg(  \frac{\Psi_{A,n,d} + \Xi_{A,n,d}}{\sqrt{ \Lambda_{A,n,d}}} \Bigg) \Bigg\} + O_P(n^{-1/2}) \\[.5em]
& = \Phi \Bigg(  \frac{\Xi_{A,n,d}}{\sqrt{ \Lambda_{A,n,d}}} \Bigg) \Bigg\{ 1 -  \Phi \Bigg(  \frac{\Xi_{A,n,d}}{\sqrt{ \Lambda_{A,n,d}}} \Bigg) \Bigg\} + O_P(n^{-1/2}), \label{Eq: E_0(1-E_0)}
\end{align}
where the second line uses the first-order Taylor expansion:
\begin{align*} 
\Phi \Bigg(  \frac{\Psi_{A,n,d} + \Xi_{A,n,d}}{\sqrt{ \Lambda_{A,n,d}}} \Bigg) ~=~&  \Phi \Bigg(  \frac{\Xi_{A,n,d}}{\sqrt{ \Lambda_{A,n,d}}} \Bigg) + O \left( \frac{\Psi_{A,n,d} }{\sqrt{ \Lambda_{A,n,d}}}\right) \\[.5em]
=~&  \Phi \Bigg(  \frac{\Xi_{A,n,d}}{\sqrt{ \Lambda_{A,n,d}}} \Bigg) + O (n^{-1/2} ).
\end{align*}
Similarly, one can obtain
\begin{align} \label{Eq: E_1(1-E_1)}
\mathcal{E}_{1,A}(1 - \mathcal{E}_{1,A}) = \Phi \Bigg(  \frac{\Xi_{A,n,d}}{\sqrt{ \Lambda_{A,n,d}}} \Bigg) \Bigg\{ 1 -  \Phi \Bigg(  \frac{\Xi_{A,n,d}}{\sqrt{ \Lambda_{A,n,d}}} \Bigg) \Bigg\} + O_P(n^{-1/2}).
\end{align}
Then by substituting (\ref{Eq: E_0(1-E_0)}) and (\ref{Eq: E_1(1-E_1)}) into the definition of $W_A$,
\begin{align} \nonumber
W_{A} & ~=~ \frac{\widehat{E}_A^S - \mathcal{E}_{0,A}/2 - \mathcal{E}_{1,A}/2}{\sqrt{\mathcal{E}_{0,A}(1-\mathcal{E}_{0,A})/(4n_{0,\text{{te}}})  + \mathcal{E}_{1,A}(1-\mathcal{E}_{1,A})/(4n_{1,\text{{te}}})}} \\[.5em] \nonumber
& = ~2 \sqrt{\frac{n_{0,\text{te}}n_{1,\text{te}}}{n_{0,\text{te}} + n_{1,\text{te}}}} \times \\
& ~~~~\frac{\widehat{E}_A^S - \mathcal{E}_{0,A}/2 - \mathcal{E}_{1,A}/2}{\sqrt{ \Phi( {\Xi_{A,n,d}}/{\sqrt{ \Lambda_{A,n,d}}} ) \{ 1 -  \Phi ({\Xi_{A,n,d}}/{\sqrt{ \Lambda_{A,n,d}}})\} + O_P(n^{-1/2}) }}. \label{Eq: W_A taylor}
\end{align}
By the Taylor expansion,
\begin{align*}
& \frac{1}{\sqrt{ \Phi( {\Xi_{A,n,d}}/{\sqrt{ \Lambda_{A,n,d}}} ) \{ 1 -  \Phi ({\Xi_{A,n,d}}/{\sqrt{ \Lambda_{A,n,d}}})\} + O_P(n^{-1/2})}}  \\[.5em]
= ~ &  \frac{1}{\sqrt{ \Phi( {\Xi_{A,n,d}}/{\sqrt{ \Lambda_{A,n,d}}} ) \{ 1 -  \Phi ({\Xi_{A,n,d}}/{\sqrt{ \Lambda_{A,n,d}}})\}}}+ O_P(n^{-1/2}).
\end{align*}
Now by plugging this into (\ref{Eq: W_A taylor}) and using the fact that $\widehat{E}_A^S - \mathcal{E}_{0,A}/2 - \mathcal{E}_{1,A}/2 = O_P(n^{-1/2})$, one can obtain that
\begin{align*}
W_A ~=~ & 2 \sqrt{\frac{n_{0,\text{te}}n_{1,\text{te}}}{n_{0,\text{te}} + n_{1,\text{te}}}} \times \\[.5em]
& \frac{\widehat{E}_A^S - \mathcal{E}_{0,A}/2 - \mathcal{E}_{1,A}/2}{\sqrt{ \Phi( {\Xi_{A,n,d}}/{\sqrt{ \Lambda_{A,n,d}}} ) \{ 1 -  \Phi ({\Xi_{A,n,d}}/{\sqrt{ \Lambda_{A,n,d}}}) \}}}  + O_P(n^{-1/2}).
\end{align*}
Lemma~\ref{lemma: E approximations} further allows us to replace $\mathcal{E}_{0,A}/2 + \mathcal{E}_{1,A}/2$ with its non-random counterpart as
\begin{align} \nonumber
W_A  = 2 & \sqrt{\frac{n_{0,\text{te}}n_{1,\text{te}}}{n_{0,\text{te}} + n_{1,\text{te}}}} \frac{1}{\sqrt{ \Phi( {\Xi_{A,n,d}}/{\sqrt{ \Lambda_{A,n,d}}} ) \{ 1 -  \Phi ({\Xi_{A,n,d}}/{\sqrt{ \Lambda_{A,n,d}}}) \}}}\times \\[.5em]
& \Bigg\{\widehat{E}_A^S - \frac{1}{2}\Phi \left( \frac{\Psi_{A,n,d} + \Xi_{A,n,d}}{ \sqrt{ \Lambda_{A,n,d}}} \right) + \frac{1}{2}\Phi  \left( \frac{\Psi_{A,n,d} - \Xi_{A,n,d}}{ \sqrt{ \Lambda_{A,n,d}}} \right) \Bigg\} + o_P(1). \label{Eq: W_A approximation}
\end{align}
Additionally, using Taylor expansion of $\Phi(x)$ around $x = \Xi_{A,n,d} / \sqrt{\Lambda_{A,n,d}}$ or $x = -\Xi_{A,n,d} / \sqrt{\Lambda_{A,n,d}}$, it is seen that 
\begin{align*}
& \Phi \left( \frac{\Psi_{A,n,d} + \Xi_{A,n,d}}{ \sqrt{ \Lambda_{A,n,d}}} \right) = \Phi \left( \frac{ \Xi_{A,n,d}}{ \sqrt{ \Lambda_{A,n,d}}} \right) + \phi \left( \frac{ \Xi_{A,n,d}}{ \sqrt{ \Lambda_{A,n,d}}} \right) \frac{\Psi_{A,n,d}}{\sqrt{ \Lambda_{A,n,d}}} + o(n^{-1/2}), \\[.5em]
& \Phi \left( \frac{\Psi_{A,n,d} - \Xi_{A,n,d}}{ \sqrt{ \Lambda_{A,n,d}}} \right) = \Phi \left( \frac{ - \Xi_{A,n,d}}{ \sqrt{ \Lambda_{A,n,d}}} \right) + \phi \left( \frac{ -\Xi_{A,n,d}}{ \sqrt{ \Lambda_{A,n,d}}} \right) \frac{\Psi_{A,n,d}}{\sqrt{ \Lambda_{A,n,d}}} + o(n^{-1/2}).
\end{align*}
This, together with $\Phi(x) + \Phi(-x) =1$ and $\phi(x) = \phi(-x)$, gives
\begin{equation}
\begin{aligned} \label{Eq: normal CDF approximation}
& \Phi \left( \frac{\Psi_{A,n,d} + \Xi_{A,n,d}}{ \sqrt{ \Lambda_{A,n,d}}} \right) + \Phi \left( \frac{\Psi_{A,n,d} - \Xi_{A,n,d}}{ \sqrt{ \Lambda_{A,n,d}}} \right) \\[.5em] 
=~ & 1 + 2  \phi \left( \frac{ \Xi_{A,n,d}}{ \sqrt{ \Lambda_{A,n,d}}} \right) \frac{\Psi_{A,n,d}}{\sqrt{ \Lambda_{A,n,d}}} + o(n^{-1/2}).
\end{aligned}
\end{equation}
Now combining (\ref{Eq: W_A approximation}) with (\ref{Eq: normal CDF approximation}), our final approximation is given by $W_A = W_A^\dagger + o_P(1)$. This proves the first part of Theorem~\ref{thm: asymptotic distribution}. 

For the second part, since $\Phi(x) + \Phi(-x) = 1$ for all $x \in \mathbb{R}$ and $\Psi_{A,n,d} = -\delta^\top A \delta / 2 = o(1)$, we have that 
\begin{align*}
\Phi \Bigg(  \frac{\Psi_{A,n,d} + \Xi_{A,n,d}}{\sqrt{ \Lambda_{A,n,d}}} \Bigg)  + \Phi \Bigg(  \frac{\Psi_{A,n,d} - \Xi_{A,n,d}}{\sqrt{ \Lambda_{A,n,d}}} \Bigg) = 1 + o(1).
\end{align*}
Thus the result follows by Lemma~\ref{lemma: E approximations}, which completes the proof of Theorem~\ref{thm: asymptotic distribution}.

\vskip 2em 

\paragraph{Proof of Theorem~\ref{thm: naive Bayes} (Power of the naive Bayes classifier test)}
Based on Lemma~\ref{lemma: Dinv approximations}, one can establish as in Theorem~\ref{thm: asymptotic distribution} that 
\begin{align*}
\sqrt{2n} \bigg( \widehat{E}_{\widehat{D}^{-1}}^S - \frac{1}{2}  - \frac{\Psi_{D^{-1},n,d}}{\sqrt{ 2\pi \Lambda_{D^{-1},n,d} }} \bigg) \convD N(0,1),
\end{align*}
where we used $n_0=n_1$, $n_{0,\text{tr}} = n_{1,\text{tr}}$ and $n_{\text{tr}} = n_{\text{te}}$. It is then straightforward to derive the power as in Section~\ref{sec: power of LDA}. Hence the result follows.

\vskip 2em

\subsection{Proof of Lemma~\ref{lemma: E approximations}} \label{sec: proof of lemma: E approximations}
\noindent The proof of Lemma~\ref{lemma: E approximations} consists of three parts. In Part 1, we provide approximations of $V_{0,A}$, $V_{1,A}$ and $V_{0,A} + V_{1,A}$, which are defined around (\ref{eq: definition of E}) (also recalled in Section~\ref{sec:preliminaries}). In Part 2, we focus on $U_A$ and present its approximation. In Part 3, by building on the results from the first two parts, we prove the main statements of Lemma~\ref{lemma: E approximations}.

\vskip 1em

\noindent \textbf{$\bullet$ Part 1.}

\vskip .5em

\noindent Using Fan's inequality in Lemma~\ref{lemma: fan's inequality} under \textbf{(A5)} and \textbf{(A6)}, observe that 
\begin{align} \label{eq: pre1}
\tr\big\{ (A\Sigma)^2 \big\} = \tr (A \Sigma A \Sigma) \leq d  \lambda^2_{\text{max}}(A)   \lambda^2_{\text{max}}(\Sigma)   \lesssim d.
\end{align}
Then under \textbf{(A1)}, $\tr\big\{ (A\Sigma)^2 \big\} = O(n)$. Next based on the sub-multiplicative property of the operator norm and \textbf{(A2)},
\begin{align} \label{eq: pre2}
0 \leq \delta^\top A \Sigma A \delta \leq \lambda_{\text{max}}(A \Sigma A )\delta^\top \delta \leq   \lambda^2_{\text{max}}(A) \lambda_{\text{max}}(\Sigma) \delta^\top \delta = O(n^{-1/2}).
\end{align}
Similarly, one can show that 
\begin{align} \label{eq: pre3}
\delta^\top A\Sigma A \Sigma A \Sigma A \delta = O(n^{-1/2}).
\end{align}
Using the ingredients above, we shall prove
\begin{equation}
\begin{aligned} \label{eq: approximation 1}
& V_{0,A} = - \frac{1}{2} \delta^\top A \delta + \frac{1}{2} \left( \frac{1}{n_{0,\text{tr}}} - \frac{1}{n_{1,\text{tr}}} \right) \tr(A\Sigma) + O_P(n^{-1/2}), \\[.5em]
& V_{1,A} = - \frac{1}{2} \delta^\top A \delta + \frac{1}{2} \left( \frac{1}{n_{1,\text{tr}}} - \frac{1}{n_{0,\text{tr}}} \right) \tr(A\Sigma) + O_P(n^{-1/2}), \quad \text{and} \\[.5em]
& V_{0,A} + V_{1,A} = - \delta^\top A \delta + O_P(n^{-3/4}),
\end{aligned}
\end{equation}
where $V_{0,A}$ and $V_{1,A}$ are defined around (\ref{eq: definition of E}). To this end, we need to calculate the mean and variance of $V_{0,A}$ and $V_{1,A}$. 
The calculation of the mean is rather straightforward as 
\begin{align*}
& \E[V_{0,A}]  =  - \frac{1}{2} \delta^\top A \delta  + \frac{1}{2} \left( \frac{1}{n_{0,\text{tr}}} - \frac{1}{n_{1,\text{tr}}} \right) \tr(A\Sigma),  \\[.5em]
& \E[V_{1,A}] =  - \frac{1}{2} \delta^\top A \delta  + \frac{1}{2} \left( \frac{1}{n_{1,\text{tr}}} - \frac{1}{n_{0,\text{tr}}} \right) \tr(A\Sigma).
\end{align*}
Turning to the variances, we will show that $\V[V_{0,A}] = \V[\widehat{\delta}^\top A (\mu_0 - \widehat{\mu}_{\text{pool}})]$ is $O(n^{-1})$. First note that $(\widehat{\delta}, \mu_0 - \widehat{\mu}_{\text{pool}})^\top$ has a multivariate normal distribution as 
\begin{align*}
\begin{pmatrix}
\widehat{\delta}\\
\mu_0 - \widehat{\mu}_{\text{pool}}
\end{pmatrix} 
\sim 
N \left(
\begin{pmatrix}
\mu_1 - \mu_0 \\[.5em]
\frac{1}{2} \mu_0 - \frac{1}{2} \mu_1
\end{pmatrix}, ~
\begin{pmatrix}
\Sigma_{11} & \Sigma_{12} \\
\Sigma_{21} & \Sigma_{22} 
\end{pmatrix} 
\right),
\end{align*}
where
\begin{align*}
\begin{pmatrix}
\Sigma_{11} & \Sigma_{12} \\[.5em]
\Sigma_{21} & \Sigma_{22} 
\end{pmatrix} 
= 
\begin{pmatrix}
(n_{0,\text{tr}}^{-1} + n_{1,\text{tr}}^{-1}) \Sigma & \frac{1}{2}(n_{1,\text{tr}}^{-1} - n_{0,\text{tr}}^{-1}) \Sigma \\[.5em]
\frac{1}{2}(n_{1,\text{tr}}^{-1} - n_{0,\text{tr}}^{-1}) \Sigma & \frac{1}{4}(n_{0,\text{tr}}^{-1} + n_{1,\text{tr}}^{-1}) \Sigma
\end{pmatrix}.
\end{align*}
We also note that the conditional distribution of $\mu_0 - \widehat{\mu}_{\text{pool}}$ given $\widehat{\delta}$ follows 
\begin{align} \label{Eq: conditional distribution}
\mu_0 - \widehat{\mu}_{\text{pool}} | \widehat{\delta} \sim N\left(\mu^\ast, \Sigma^\ast \right),
\end{align}
where $\mu^\ast = - \delta /2  + \Sigma_{21} \Sigma^{-1}_{11}(\widehat{\delta} - \delta)$ and $\Sigma^\ast = \Sigma_{22} - \Sigma_{21} \Sigma_{11}^{-1} \Sigma_{12}$. Next, by the law of total variance,
\begin{align*}
\V[\widehat{\delta}^\top A (\mu_0 - \widehat{\mu}_{\text{pool}})] = \underbrace{\E[\V[\widehat{\delta}^\top A (\mu_0 - \widehat{\mu}_{\text{pool}}) | \widehat{\delta}]]}_{(I)} + \underbrace{\V[\E[\widehat{\delta}^\top A (\mu_0 - \widehat{\mu}_{\text{pool}}) | \widehat{\delta}]]}_{(II)}.
\end{align*}
Using (\ref{Eq: conditional distribution}), $(I)$ and $(II)$ are simplified as
\begin{align*}
(I) = \E[\widehat{\delta}^\top A \Sigma^\ast A \widehat{\delta} ] \quad \text{and} \quad (II) = \V[\widehat{\delta}^\top A \{ -\delta/2 + \Sigma_{21} \Sigma_{11}^{-1} (\widehat{\delta} - \delta )\}].
\end{align*}
By recalling the definitions of $\Sigma^\ast$, $\Sigma_{21}$ and $\Sigma_{11}$, 
\begin{align} \nonumber
(I) ~ & \lesssim ~ \left( \frac{1}{n_{0,\text{tr}}} + \frac{1}{n_{1,\text{tr}}} \right) \E[\widehat{\delta}^\top A \Sigma A \widehat{\delta}] \\[.5em] \nonumber
& ~~+  \left( \frac{1}{n_{1,\text{tr}}} - \frac{1}{n_{0,\text{tr}}} \right)^2 \left( \frac{1}{n_{0,\text{tr}}} + \frac{1}{n_{1,\text{tr}}} \right)^{-1} \E[\widehat{\delta}^\top A \Sigma A \widehat{\delta}]  \\[.5em]
~ & \lesssim ~ \frac{1}{n} \E[\widehat{\delta}^\top A \Sigma A \widehat{\delta}], \label{Eq: term (I)}
\end{align}
where the second line uses the assumptions \textbf{(A3)} and \textbf{(A4)}. Here the symbol $a_n \lesssim b_n$ means that there exists a constant $c>0$ such that $a_n \leq c b_n$ for large $n$. In addition, it can be checked that 
\begin{align} \label{Eq: term (II)}
(II) ~ & \lesssim ~ \V[\widehat{\delta}^\top A \delta] + \V[\widehat{\delta}^\top A \widehat{\delta}]. 
\end{align} 
Now, based on Lemma~\ref{lemma: quadratic form}, one can verify that 
\begin{align*}
& \E[\widehat{\delta}^\top A \Sigma A \widehat{\delta}] = \left( \frac{1}{n_{0,\text{tr}}} + \frac{1}{n_{1,\text{tr}}} \right) \tr( A \Sigma A \Sigma) + \delta^\top  A \Sigma A  \delta, \\[.5em]
& \V[\widehat{\delta}^\top A \delta] = \left( \frac{1}{n_{0,\text{tr}}} + \frac{1}{n_{1,\text{tr}}} \right) \delta^\top A \Sigma A \delta, \\[.5em]
& \V[\widehat{\delta}^\top A \widehat{\delta}] = 2 \left( \frac{1}{n_{0,\text{tr}}} + \frac{1}{n_{1,\text{tr}}} \right)^2 \tr(A\Sigma A \Sigma) + 4 \left( \frac{1}{n_{0,\text{tr}}} + \frac{1}{n_{1,\text{tr}}} \right) \delta^\top A \Sigma A \delta. 
\end{align*}
By substituting the above expressions into (\ref{Eq: term (I)}) and (\ref{Eq: term (II)}) together with the preliminaries in (\ref{eq: pre1}) and (\ref{eq: pre2}), we have that $\V[V_{0,A}] = O(n^{-1})$ as desired. The same lines of argument also show that $\V[V_{1,A}] = O(n^{-1})$ and therefore the first two lines in (\ref{eq: approximation 1}) follow. Additionally, by noting that $V_{0,A} + V_{1,A}  = \widehat{\delta}^\top A(\mu_0 - \mu_1)$, we have 
\begin{align*}
&\E [V_{0,A} + V_{1,A} ] =  - \delta^\top A \delta, \\[.5em]
&\V [V_{0,A} + V_{1,A} ] = \left( \frac{1}{n_{0,\text{{tr}}}} + \frac{1}{n_{1,\text{{tr}}}} \right) \delta^\top A \Sigma A \delta.
\end{align*}
The above means and variances, together with (\ref{eq: pre1}) and (\ref{eq: pre2}), yield the claim (\ref{eq: approximation 1}).

\vskip 1em

\noindent \textbf{$\bullet$ Part 2.}

\vskip .5em

\noindent Applying Lemma~\ref{lemma: quadratic form} yields
\begin{equation}
\begin{aligned} \label{eq: pre4}
& \E[U_A] = \delta^\top A\Sigma A \delta + \left( \frac{1}{n_{0,\text{{tr}}}} + \frac{1}{n_{1,\text{{tr}}}} \right){\tr} \big\{  (A\Sigma)^2 \big\}   \ \ \text{and} \\[.5em]
& \V[U_A] = 2\left(  \frac{1}{n_{0,\text{{tr}}}} + \frac{1}{n_{1,\text{{tr}}}}\right)^2 {\tr} \big\{  (A\Sigma)^4 \big\} + 4  \left( \frac{1}{n_{0,\text{{tr}}}} + \frac{1}{n_{1,\text{{tr}}}} \right) \delta^\top A\Sigma A \Sigma A \Sigma A \delta.
\end{aligned}
\end{equation}
As in (\ref{eq: pre1}), Fan's inequality shows $\tr\big\{ (A\Sigma)^4 \big\} = O(n)$. This fact, together with (\ref{eq: pre3}) and (\ref{eq: pre4}), gives
\begin{align} \label{eq: approximation 2}
U_A = \delta^\top A\Sigma A \delta + \left( \frac{1}{n_{0,\text{{tr}}}} + \frac{1}{n_{1,\text{{tr}}}} \right){\tr} \big\{  (A\Sigma)^2 \big\} + O_P(n^{-1/2}),
\end{align}
which completes the second part. 

\clearpage 

\noindent \textbf{$\bullet$ Part 3.}

\vskip .5em

\noindent Consider a bivariate function $f(v,u) = \Phi(v/\sqrt{u})$. Recall the definition of $\Psi_{A,n,d}$, $\Lambda_{A,n,d}$ and $\Xi_{A,n,d}$ in (\ref{eq: psi/lambda}) (also recalled in Section~\ref{sec:preliminaries}). Then by the Taylor expansion of $f(v,u)$ around $(\Psi_{A,n,d} + \Xi_{A,n,d}, \Lambda_{A,n,d})$ together with (\ref{eq: approximation 1}) and (\ref{eq: approximation 2}), we have
\begin{equation}
\begin{aligned} \label{Eq: Approx 1}
& \mathcal{E}_{0,A} = f(V_{0,A},U_A) \\[.5em]
& = \Phi \Bigg(  \frac{\Psi_{A,n,d} + \Xi_{A,n,d}}{\sqrt{ \Lambda_{A,n,d}  }} \Bigg) + \phi\Bigg(  \frac{\Psi_{A,n,d} + \Xi_{A,n,d}}{ \sqrt{\Lambda_{A,n,d} }} \Bigg) \frac{1}{\sqrt{ \Lambda_{A,n,d} }} (V_{0,A} - \Psi_{A,n,d} - \Xi_{A,n,d}) \\[.5em]
& - \phi\Bigg(  \frac{\Psi_{A,n,d} + \Xi_{A,n,d} }{ \sqrt{\Lambda_{A,n,d} }} \Bigg) \frac{\Psi_{A,n,d} + \Xi_{A,n,d}}{(\Lambda_{A,n,d} )^{3/2}} (U_A - \Lambda_{A,n,d}) + O_P\left( n^{-1} \right),
\end{aligned}
\end{equation}
where we recall that $\phi(\cdot)$ is the density function of $N(0,1)$. Similarly,
\begin{equation}
\begin{aligned} \label{Eq: Approx 2}
\mathcal{E}_{1,A} & = f(V_{1,A},U_A)  \\[.5em]
& = \Phi \Bigg(  \frac{\Psi_{A,n,d} - \Xi_{A,n,d}}{\sqrt{ \Lambda_{A,n,d}  }} \Bigg)  + \phi\Bigg(  \frac{\Psi_{A,n,d} - \Xi_{A,n,d}}{ \sqrt{\Lambda_{A,n,d} }} \Bigg) \frac{1}{\sqrt{ \Lambda_{A,n,d} }} (V_{1,A} - \Psi_{A,n,d} + \Xi_{A,n,d}) \\[.5em]
& - \phi\Bigg(  \frac{\Psi_{A,n,d} - \Xi_{A,n,d}}{ \sqrt{\Lambda_{A,n,d} }} \Bigg) \frac{\Psi_{A,n,d} - \Xi_{A,n,d}}{(\Lambda_{A,n,d} )^{3/2}} (U_A - \Lambda_{A,n,d}) + O_P\left( n^{-1} \right).
\end{aligned}
\end{equation}
Since the normal density function $\phi(\cdot)$ is bounded and $\liminf_{n,d \rightarrow \infty} \Lambda_{A,n,d}$ is a (strictly) positive constant under the given conditions (see the proof of Proposition~\ref{proposition: conditional CLT}), the first two claims in Lemma~\ref{lemma: E approximations} follow, i.e.
\begin{align*}
&\mathcal{E}_{0,A} ~=~ \Phi \Bigg(  \frac{\Psi_{A,n,d} + \Xi_{A,n,d}}{\sqrt{ \Lambda_{A,n,d}}} \Bigg) + O_P\left(n^{-1/2} \right), \\[.5em]
&\mathcal{E}_{1,A} ~=~ \Phi \Bigg(  \frac{\Psi_{A,n,d} - \Xi_{A,n,d}}{\sqrt{ \Lambda_{A,n,d}}} \Bigg) + O_P\left(n^{-1/2} \right).
\end{align*}
Combining (\ref{Eq: Approx 1}) and (\ref{Eq: Approx 2}) yields that
\begin{equation}
\begin{aligned} \label{Eq: goal of part 3}
\mathcal{E}_{0,A} + \mathcal{E}_{1,A} ~  =  ~ &    \Phi \Bigg(  \frac{\Psi_{A,n,d} + \Xi_{A,n,d}}{\sqrt{ \Lambda_{A,n,d}  }} \Bigg)
+   \Phi \Bigg(  \frac{\Psi_{A,n,d} - \Xi_{A,n,d}}{\sqrt{ \Lambda_{A,n,d}}} \Bigg)  \\[.5em] 
+ ~ &   (I)' - (II)' + O_P(n^{-1}),
\end{aligned}
\end{equation}
where
\begin{align*}
(I)' ~ = & ~ \phi\Bigg(  \frac{\Psi_{A,n,d} + \Xi_{A,n,d}}{ \sqrt{\Lambda_{A,n,d} }} \Bigg) \frac{1}{\sqrt{ \Lambda_{A,n,d} }} (V_{0,A} - \Psi_{A,n,d} - \Xi_{A,n,d}) \\[.5em]
& ~ + \phi\Bigg(  \frac{\Psi_{A,n,d} - \Xi_{A,n,d}}{ \sqrt{\Lambda_{A,n,d} }} \Bigg) \frac{1}{\sqrt{ \Lambda_{A,n,d} }} (V_{1,A} - \Psi_{A,n,d} + \Xi_{A,n,d}) \quad \text{and} \\[.5em]
(II)' ~ = & ~ \phi\Bigg(  \frac{\Psi_{A,n,d} + \Xi_{A,n,d} }{ \sqrt{\Lambda_{A,n,d} }} \Bigg) \frac{\Psi_{A,n,d} + \Xi_{A,n,d}}{(\Lambda_{A,n,d} )^{3/2}} (U_A - \Lambda_{A,n,d}) \\[.5em]
& ~ + \phi\Bigg(  \frac{\Psi_{A,n,d} - \Xi_{A,n,d}}{ \sqrt{\Lambda_{A,n,d} }} \Bigg) \frac{\Psi_{A,n,d} - \Xi_{A,n,d}}{(\Lambda_{A,n,d} )^{3/2}} (U_A - \Lambda_{A,n,d}).
\end{align*}
Focusing on $(I)'$, we use the fact that $\phi(x) = \phi(-x)$ to obtain
\begin{align*}
(I)' ~ & = ~ \phi\Bigg(  \frac{\Xi_{A,n,d}}{ \sqrt{\Lambda_{A,n,d} }} \Bigg) \frac{1}{\sqrt{ \Lambda_{A,n,d} }} (V_{0,A} + V_{1,A} - 2\Psi_{A,n,d}) \\[.5em]
& + \left[ \phi\Bigg(  \frac{\Psi_{A,n,d} + \Xi_{A,n,d}}{ \sqrt{\Lambda_{A,n,d} }} \Bigg) - \phi\Bigg( \frac{\Xi_{A,n,d}}{ \sqrt{\Lambda_{A,n,d} }} \Bigg) \right] \frac{1}{\sqrt{ \Lambda_{A,n,d} }} (V_{0,A} - \Psi_{A,n,d} - \Xi_{A,n,d}) \\[.5em]
& + \left[ \phi\Bigg(  \frac{\Psi_{A,n,d} - \Xi_{A,n,d}}{ \sqrt{\Lambda_{A,n,d} }} \Bigg) - \phi\Bigg(  \frac{-\Xi_{A,n,d}}{ \sqrt{\Lambda_{A,n,d} }} \Bigg) \right] \frac{1}{\sqrt{ \Lambda_{A,n,d} }} (V_{1,A} - \Psi_{A,n,d} + \Xi_{A,n,d}). 
\end{align*}
By using the asymptotic relationships in (\ref{eq: approximation 1}) and $\Psi_{A,n,d} = o(1)$, 
\begin{align*}
(I)' ~ & = ~ O(1) \cdot O_P(n^{-3/4}) + o(1) \cdot O_P(n^{-1/2}) + o(1) \cdot O_P(n^{-1/2}) \\[.5em]
~ & = ~ o_P(n^{-1/2}).
\end{align*}
Similarly, one can establish by using (\ref{eq: approximation 2}) and $\Psi_{A,n,d} = o(1)$ that
\begin{align*}
(II)' ~ & = ~ \Bigg[  \phi\Bigg(  \frac{\Psi_{A,n,d} + \Xi_{A,n,d} }{ \sqrt{\Lambda_{A,n,d} }} \Bigg) \frac{\Psi_{A,n,d} + \Xi_{A,n,d}}{(\Lambda_{A,n,d} )^{3/2}} \\[.5em] 
& ~~~~~~~~~~~~~~~~~+ \phi\Bigg(  \frac{\Psi_{A,n,d} - \Xi_{A,n,d}}{ \sqrt{\Lambda_{A,n,d} }} \Bigg) \frac{\Psi_{A,n,d} - \Xi_{A,n,d}}{(\Lambda_{A,n,d} )^{3/2}} \Bigg](U_A - \Lambda_{A,n,d}) \\[.5em] 
& = ~ o(1) \cdot O_P(n^{-1/2}) = o_P(n^{-1/2}).
\end{align*}
Now by substituting these results to (\ref{Eq: goal of part 3}), 
\begin{align*}
\mathcal{E}_{A,0} + \mathcal{E}_{A,1} ~ = ~ \Phi \Bigg(  \frac{\Psi_{A,n,d} + \Xi_{A,n,d}}{\sqrt{ \Lambda_{A,n,d}  }} \Bigg) +  \Phi \Bigg(  \frac{\Psi_{A,n,d} - \Xi_{A,n,d}}{\sqrt{ \Lambda_{A,n,d}}} \Bigg)  + o_P(n^{-1/2}),
\end{align*}
as desired. If $n_{0,\text{tr}} = n_{1,\text{tr}}$, then the approximations of $(I)'$ and $(II)'$ become much more straightforward with $\Xi_{A,n,d} = 0$. Indeed, one can infer that $(I)' = O_P(n^{-3/4})$ and $(II)' = O_P(n^{-3/4})$, which yields
\begin{align*}
\mathcal{E}_{A,0} + \mathcal{E}_{A,1} ~ = ~ 2\Phi \Bigg(  \frac{\Psi_{A,n,d}}{\sqrt{ \Lambda_{A,n,d}}} \Bigg) + O_P(n^{-3/4}).
\end{align*}
This completes the proof of Lemma~\ref{lemma: E approximations}. 

\vskip 2em

\subsection{Some moments of (scaled) inverse chi-square random variables} \label{sec: Some moments of (scaled) inverse chi-square random variables}
In this section, we provide two lemmas (Lemma~\ref{Lemma: invChi-square moments} and Lemma~\ref{Lemma: sisj expectation}) where we present some moments of (scaled) inverse chi-square random variables. These results will be used to prove Lemma~\ref{lemma: Dinv approximations}. Throughout this section, we assume that $n_{0,\text{tr}} = n_{1,\text{tr}}$. Let us denote the diagonal elements of $\widehat{D}$ by $s_1^2,\ldots,s_d^2$ where 
\begin{align*}
s_k^2 = \frac{1}{2(n_{0,\text{tr}}-1)} \sum_{i=1}^{n_{0,\text{tr}}} (X_{ik} - \overline{X}_k)^2 + \frac{1}{2(n_{1,\text{tr}}-1)} \sum_{i=1}^{n_{1,\text{tr}}} (Y_{ik} - \overline{Y}_k)^2,
\end{align*}
for $k=1,\ldots,d$. Here $\overline{X}_k$ and $\overline{Y}_k$ are the sample means based on the training set, i.e. $\overline{X}_k = n_{0,\text{tr}}^{-1} \sum_{i=1}^{n_{0,\text{tr}}} X_{ik}$ and $\overline{Y}_k = n_{1,\text{tr}}^{-1} \sum_{i=1}^{n_{1,\text{tr}}} Y_{ik}$. Then by putting $\sigma_k^2 = [\Sigma]_{k,k}$, we have
\begin{align} \label{Eq: definition of 1/sk^2}
\frac{1}{s_k^2} \sim \frac{n_{\text{tr}}-2}{\sigma_k^2} \text{inv-}\chi_{n_{\text{tr}}-2}^2, \text{ and } \frac{(n_{\text{tr}}-2) s_k^2}{\sigma_k^2} \sim \chi_{n_{\text{tr}}-2}^2,
\end{align}
where $\text{inv-}\chi_{n_{\text{tr}}-2}^2$ represents the inverse-chi-squared distribution with $n_{\text{tr}}-2$ degrees of freedom. 

\noindent Let us investigate some moments of $s_k^{-2}$, which will be used to control the inverse of $\widehat{D}$.
\begin{lemma} \label{Lemma: invChi-square moments}
	Suppose that $n_{0,\text{\emph{tr}}} = n_{1,\text{\emph{tr}}}$. Then under \textbf{(A4)}, some of non-central moments of $s_k^{-2}$ are given by
	\begin{align*}
	\E \left[\frac{1}{s_k^2} \right] & ~=~ \frac{n_{\text{\emph{tr}}}-2}{\sigma_k^2 } \frac{1}{n_{\text{\emph{tr}}}-4}= \frac{1}{\sigma_k^2} \big\{ 1 + O(n^{ -1})  \big\}, \\[.5em]
	\E \left[ \frac{1}{s_k^4} \right] & ~=~ \frac{(n_{\text{\emph{tr}}}-2)^2}{\sigma_k^4} \frac{1}{(n_{\text{\emph{tr}}}-4)(n_{\text{\emph{tr}}}-8)} = \frac{1}{\sigma_k^4}\big\{ 1 + O(n^{ -1}) \big\}, \\[.5em]
	\E \left[  \frac{1}{s_k^6} \right] & ~=~ \frac{(n_{\text{\emph{tr}}}-2)^3}{\sigma_k^6}\frac{1}{(n_{\text{\emph{tr}}}-12)(n_{\text{\emph{tr}}}-8)(n_{\text{\emph{tr}}}-2)} \\[.5em]
	& ~=~ \frac{1}{\sigma_k^6} \big\{ 1 + O(n^{ -1}) \big\}, \\[.5em]
	\E \left[  \frac{1}{s_k^8} \right] & ~=~ \frac{(n_{\text{\emph{tr}}}-8)^4}{\sigma_k^8} \frac{1}{(n_{\text{\emph{tr}}}-16)(n_{\text{\emph{tr}}}-12)(n_{\text{\emph{tr}}}-8)(n_{\text{\emph{tr}}}-4)} \\[.5em]
	& ~=~ \frac{1}{\sigma_k^8} \big\{ 1 + O(n^{ -1}) \big\}.
	\end{align*}
	In addition, a couple of the central moments are
	\begin{align*}
	\E \left[ \left(  \frac{1}{s_k^2} - \E\left[ \frac{1}{s_k^2}\right] \right)^2 \right] & = \frac{(n_{\text{\emph{tr}}}-2)^2}{\sigma_k^4} \frac{2}{(n_{\text{\emph{tr}}}-4)^2(n_{\text{\emph{tr}}}-6)} \\[.5em]
	& = \frac{1}{\sigma_k^2} \cdot O(n^{ -1}), \\[.5em]
	\E \left[ \left(  \frac{1}{s_k^2} - \E\left[ \frac{1}{s_k^2}\right] \right)^4 \right] & = \frac{(n_{\text{\emph{tr}}}-2)^4}{\sigma_k^8} \frac{12(n_{\text{\emph{tr}}}-2)^2+72(n_{\text{\emph{tr}}}-2)-480}{(n_{\text{\emph{tr}}}-8)(n_{\text{\emph{tr}}}-10)(n_{\text{\emph{tr}}}-4)^4(n_{\text{\emph{tr}}}-6)^2} \\[.5em]
	& =\frac{1}{\sigma_k^8} \cdot O(n^{ -2}).
	\end{align*}
	\begin{proof}
		Suppose that $X \sim \chi_\nu^2$. Then for $\nu \geq 2k + 1$, 
		\begin{align*}
		\E[X^{-k}] & = \int_{0}^{\infty} x^{-k} \frac{1}{2^{\nu/2}\Gamma(\nu/2)} x^{\nu/2-1} e^{-x/2}dx \\[.5em]
		& =  \frac{1}{2^{2k}} \frac{\Gamma(\nu/2-k)}{\Gamma(\nu/2)} \int_{0}^{\infty} \frac{1}{2^{\frac{\nu-2k}{2}}\Gamma\{(\nu-2k)/2\}} x^{\frac{\nu-2k}{2}-1} e^{-x/2} dx \\[.5em]
		& =  \frac{1}{2^{2k}} \frac{\Gamma(\nu/2-k)}{\Gamma(\nu/2)},
		\end{align*}
		where the last equality uses the fact that a density integrates to one. Using this exact inverse moment expression and the relationship in (\ref{Eq: definition of 1/sk^2}), the results follows by straightforward algebra. 
	\end{proof}
\end{lemma}

\noindent Next we study some product moments of (scaled) inverse chi-square random variables.

\begin{lemma}  \label{Lemma: sisj expectation}
	Suppose that $n_{0,\text{\emph{tr}}} = n_{1,\text{\emph{tr}}}$. Then under \textbf{(A4)}, for any $1 \leq i,j,k,l \leq d$,
	\begin{align} \label{Eq: prod2}
	& \E\left[ \frac{1}{s_i^2 s_j^2} - \frac{1}{\sigma_i^2 \sigma_j^2} \right] = O(n^{-1}), \\[.5em]  \label{Eq: prod3}
	& \E\left[  \frac{1}{s_i^2 s_j^2 s_k^2}  - \frac{1}{\sigma_i^2 \sigma_j^2 \sigma_k^2} \right] = O(n^{-1}) \ \text{and} \\[.5em] \label{Eq: prod4}
	& \E\left[  \frac{1}{s_i^2 s_j^2 s_k^2 s_l^2}  - \frac{1}{\sigma_i^2 \sigma_j^2 \sigma_k^2 \sigma_l^2} \right] = O(n^{-1}).
	\end{align}		
	\begin{proof}
		To prove claim (\ref{Eq: prod2}), write
		\begin{align*}
		\frac{1}{s_i^2 s_j^2} - \frac{1}{\sigma_i^2 \sigma_j^2}  = \left( \frac{1}{s_i^2} -\frac{1}{\sigma_i^2} \right)\left( \frac{1}{s_j^2} -\frac{1}{\sigma_j^2} \right) + \left( \frac{1}{s_i^2} - \frac{1}{\sigma_i^2}\right)\frac{1}{\sigma_j^2} + \left( \frac{1}{s_j^2} - \frac{1}{\sigma_j^2}\right)\frac{1}{\sigma_i^2}.
		\end{align*}
		Then by using Cauchy-Schwarz inequality, we see that 
		\begin{equation}
		\begin{aligned} \label{Eq: product of two}
		\Bigg| \E \left[ \frac{1}{s_i^2 s_j^2} - \frac{1}{\sigma_i^2 \sigma_j^2} \right] \Bigg| ~ & \leq ~  \E \left[ \left( \frac{1}{s_i^2} -\frac{1}{\sigma_i^2}  \right)^2 \right]  + \frac{1}{\sigma_j^2}  \Bigg| \E \left[   \frac{1}{s_i^2} - \frac{1}{\sigma_i^2} \right]   \Bigg| \\[.5em]
		& + \frac{1}{\sigma_i^2}  \Bigg| \E \left[   \frac{1}{s_j^2} - \frac{1}{\sigma_j^2} \right]   \Bigg|.
		\end{aligned}
		\end{equation}
		The three terms on the right-hand side are $O(n^{-1})$ by Lemma~\ref{Lemma: invChi-square moments} and thus (\ref{Eq: prod2}) follows. 
		
		\vskip 0.8em
		
		\noindent Next we prove (\ref{Eq: prod3}); the result in (\ref{Eq: prod4}) follows similarly. Write
		\begin{equation}
		\begin{aligned} \label{Eq: product of three}
		\frac{1}{s_i^2 s_j^2 s_k^2}  - \frac{1}{\sigma_i^2 \sigma_j^2 \sigma_k^2} & ~=~ \left( \frac{1}{s_i^2s_j^2} - \frac{1}{\sigma_i^2\sigma_j^2} \right) \left( \frac{1}{s_k^2} - \frac{1}{\sigma_k^2} \right)  \\[.5em]
		&~~~ + \left( \frac{1}{s_i^2s_j^2} - \frac{1}{\sigma_i^2\sigma_j^2} \right) \frac{1}{\sigma_k^2} + \left( \frac{1}{s_k^2} - \frac{1}{\sigma_k^2} \right) \frac{1}{\sigma_i^2\sigma_j^2}.
		\end{aligned}
		\end{equation}
		Note that the expected values of the last two terms in (\ref{Eq: product of three}) are $O(n^{-1})$ by Lemma~\ref{Lemma: invChi-square moments} and (\ref{Eq: prod2}). Therefore we focus on the first term and show that its expected value is $O(n^{-1})$. The first term can be decomposed as
		\begin{align} \nonumber
		& \left( \frac{1}{s_i^2s_j^2} - \frac{1}{\sigma_i^2\sigma_j^2} \right) \left( \frac{1}{s_k^2} - \frac{1}{\sigma_k^2} \right) \\[.5em] \label{Eq: three terms decomposition}
		= ~ &\Bigg[ \left( \frac{1}{s_i^2} -\frac{1}{\sigma_i^2} \right)\left( \frac{1}{s_j^2} -\frac{1}{\sigma_j^2} \right) + \left( \frac{1}{s_i^2} - \frac{1}{\sigma_i^2}\right)\frac{1}{\sigma_j^2} + \left( \frac{1}{s_j^2} - \frac{1}{\sigma_j^2}\right)\frac{1}{\sigma_i^2} \Bigg] \left( \frac{1}{s_k^2} - \frac{1}{\sigma_k^2} \right).
		\end{align}
		We only need to handle the following term in (\ref{Eq: three terms decomposition})
		\begin{align} \label{Eq: core of prod3}
		\left( \frac{1}{s_i^2} -\frac{1}{\sigma_i^2} \right)\left( \frac{1}{s_j^2} -\frac{1}{\sigma_j^2} \right) \left( \frac{1}{s_k^2} - \frac{1}{\sigma_k^2} \right),
		\end{align}
		since the expected values of the other terms are $O(n^{-1}$), which follows as in (\ref{Eq: product of two}) using Cauchy-Schwarz inequality. But the expectation of (\ref{Eq: core of prod3}) is $O(n^{-1})$ again by Cauchy-Schwarz inequality and Lemma~\ref{Lemma: invChi-square moments}. Thus the expectation of (\ref{Eq: three terms decomposition}) is $O(n^{-1})$. Finally, after substituting this result into the expectation of (\ref{Eq: product of three}), we may obtain the result in (\ref{Eq: prod3}). Hence Lemma~\ref{Lemma: sisj expectation} follows. 
	\end{proof}
\end{lemma}

\vskip 2em

\subsection{Proof of Lemma~\ref{lemma: Dinv approximations}} \label{sec: Lemma Dinv}

\noindent Let us denote 
\begin{align*}
V_{0,\widehat{D}^{-1}} &\defn \widehat{\delta}^\top \widehat{D}^{-1} (\mu_0 - \widehat{\mu}_{\text{pool}}),\\
V_{1,\widehat{D}^{-1}} &\defn \widehat{\delta}^\top \widehat{D}^{-1} (\widehat{\mu}_{\text{pool}} -\mu_1), \text{ and } \\
U_{\widehat{D}^{-1}} &\defn \widehat{\delta}^\top \widehat{D}^{-1} \Sigma \widehat{D}^{-1} \widehat{\delta}.
\end{align*}
By assuming \textbf{\textbf{(A1)}}--\textbf{(A5)} with $n_0=n_1$, $n_{0,\text{tr}}=n_{1,\text{tr}}$ and $n_{\text{{tr}}} = n_{\text{{te}}}$, we break the proof up into three parts:

\begin{enumerate}\setlength{\itemindent}{0.3in}
	\item[\textbf{$\bullet$ Part 1.}] $V_{0,\widehat{D}^{-1}} = \Psi_{D^{-1},n,d}  + O_P\left( n^{-1/2} \right) \ \text{and} \  V_{1,\widehat{D}^{-1}} = \Psi_{D^{-1},n,d} + O_P\left( n^{-1/2} \right)$. \\[-.5em]
	\item[\textbf{$\bullet$ Part 2.}] $U_{\widehat{D}^{-1}} =  \Lambda_{D^{-1},n,d} +  O_P \left( n^{-1/2} \right).$  \\[-.5em]
	\item[\textbf{$\bullet$ Part 3.}] $V_{0,\widehat{D}^{-1}}+ V_{1,\widehat{D}^{-1}} = 2\Psi_{D^{-1},n,d} + O_P \left( n^{-3/4} \right).$
\end{enumerate}

\noindent Suppose that the above claims hold. Then the final result of Lemma~\ref{lemma: Dinv approximations} follows similarly as in the proof of Lemma~\ref{lemma: E approximations} via Taylor expansion. We now verify each claim in order. 

\vskip 1em

\clearpage 

\noindent \textbf{$\bullet$ Part 1.}

\vskip .5em

\noindent We only prove that
\begin{align*}
V_{0,\widehat{D}^{-1}} &= -\frac{1}{2}\delta^\top D^{-1} \delta + O_P\left(\frac{1}{\sqrt{n}}\right).
\end{align*}
The argument for $V_{1,\widehat{D}^{-1}}$ follows analogously. Under the Gaussian assumption with mutually independent random samples, one can see that the following vector
\begin{align*}
(\underbrace{\overline{X}_1, \ldots, \overline{X}_d, \overline{Y}_1, \ldots, \overline{Y}_d}_{(A)}, \underbrace{X_{11} - \overline{X}_1,\ldots,X_{n_0 1} - \overline{X}_1, Y_{11} - \overline{Y}_1, \ldots, Y_{n_1 d} - \overline{Y}_d}_{(B)})
\end{align*}
has a multivariate normal distribution. Furthermore, a little algebra shows that the cross-covariance matrix between $(A)$ and $(B)$ is a zero matrix, which implies that $(A)$ and $(B)$ are independent under the Gaussian assumption. Since $\widehat{D}^{-1}$ is a function of $(B)$, it shows that $\widehat{D}^{-1}$ is independent of $\widehat{\delta}$ and $\widehat{\mu}_{\text{pool}}$, which are functions of $(A)$. In addition, when $n_{0,\text{tr}} = n_{1,\text{tr}}$, the covariance between $\widehat{\delta}$ and $\widehat{\mu}_{\text{pool}}$ is a zero matrix as
\begin{align*}
\text{Cov}(\widehat{\mu}_1 - \widehat{\mu}_0, \widehat{\mu}_1/2 + \widehat{\mu}_0/2) = \left( \frac{1}{2n_{1,\text{tr}}} - \frac{1}{2n_{0,\text{tr}}} \right) \Sigma = 0,
\end{align*}
which further implies that $\widehat{D}^{-1}$, $\widehat{\delta}$ and $\widehat{\mu}_{\text{pool}}$ are mutually independent. Based on this observation, the expectation becomes
\begin{align} \nonumber
\E[V_{0,\widehat{D}^{-1}}] & = - \frac{1}{2}\delta^\top \E[\widehat{D}^{-1}]  \delta = -\frac{1}{2}\sum_{i=1}^d \delta_i^2 \E\left[ \frac{1}{s_i^2} \right] = -\frac{1}{2}\delta^\top D^{-1} \delta  + \delta^\top \delta \cdot  O(n^{-1}) \\[-.5em]
& =  -\frac{1}{2}\delta^\top D^{-1} \delta + O \left( \frac{1}{n^{3/2}} \right), \label{Eq: expected value of V}
\end{align}
since $\delta^\top  \delta = O(n^{-1/2})$ under \textbf{\textbf{(A1)}}, \textbf{(A2)} and \textbf{(A5)}. 

\vskip .5em

\noindent Next calculate the second moment using Lemma~\ref{lemma: quadratic form} as
\begin{align*}
& \E [V_{0,\widehat{D}^{-1}}^2] \\[.5em]
~=~ & \E \Bigg[ \tr \Bigg\{ \E \left[ \widehat{\delta} \widehat{\delta}^\top \right] \widehat{D}^{-1} \E \left[ (\mu_0 - \widehat{\mu}_{\text{pool}})  (\mu_0 - \widehat{\mu}_{\text{pool}})^\top \right] \widehat{D}^{-1}  \Bigg\}\Bigg] \\[.5em]
~=~ & \E \Bigg[ \tr \Bigg\{  \left[ \delta \delta^\top + \frac{4}{n_{\text{tr}}} \Sigma \right] \widehat{D}^{-1} \left[ \frac{1}{4}\delta \delta^\top  + \frac{1}{n_{\text{tr}}} \Sigma \right] \widehat{D}^{-1}  \Bigg\}\Bigg] \\[.5em]
~=~ & \underbrace{\frac{1}{4} \E \left[ \left( \delta^\top \widehat{D}^{-1} \delta \right)^2 \right]}_{(I)} + \underbrace{\frac{2}{n_{\text{tr}}} \E \left[ \delta^\top \widehat{D}^{-1} \Sigma \widehat{D}^{-1} \delta \right]}_{(II)} + \underbrace{\frac{4}{n_{\text{tr}}^{2}} \E \left[  \tr  \bigg\{ \left( \Sigma \widehat{D}^{-1} \right)^2 \bigg\} \right]}_{(III)}.
\end{align*}
For $(I)$, we apply Lemma~\ref{Lemma: sisj expectation} to have
\begin{align*}
(I) ~=~ \frac{1}{4} \sum_{i=1}^d \sum_{j=1}^d \delta_i^2 \delta_j^2 \E \left[ \frac{1}{s_i^2 s_j^2}  \right] = \frac{1}{4} \left( \delta^\top D^{-1} \delta \right)^2 + O\left( \frac{1}{n^2} \right).
\end{align*}
For $(II)$, by writing $\sigma_{ij} = [\Sigma]_{ij}$, we infer that
\begin{align*}
(II) & ~=~ \frac{2}{n_{\text{tr}}} \E \left[ \delta^\top \widehat{D}^{-1} \Sigma \widehat{D}^{-1} \delta \right] = \frac{2}{n_{\text{tr}}} \sum_{i=1}^d \sum_{j=1}^d \delta_i \delta_j \sigma_{ij} \E \left[ \frac{1}{s_i^2 s_j^2} \right] \\[.5em]
& ~=~ \frac{2}{n_{\text{tr}}}\delta^\top D^{-1} \Sigma D^{-1} \delta + O\left( \frac{\delta^\top \Sigma \delta}{n^2} \right) = \frac{2}{n_{\text{tr}}}\delta^\top D^{-1} \Sigma D^{-1} \delta  + O \left( \frac{1}{n^{5/2}} \right).
\end{align*}
The last term simplifies as
\begin{align*}
(III) & ~=~ \frac{4}{n_{\text{tr}}^{2}} \E \left[  \tr  \bigg\{ \left( \Sigma \widehat{D}^{-1} \right)^2 \bigg\} \right] = \frac{4}{n_{\text{tr}}^{2}} \sum_{i=1}^d \sum_{j=1}^d \sigma_{ij}^2 \E \left[ \frac{1}{s_i^2 s_j^2} \right] \\[.5em]
& ~=~ \frac{4}{n_{\text{tr}}^{2}} \tr  \big\{ \left( \Sigma D^{-1} \right)^2 \big\} + O \left(  \frac{ \tr (\Sigma^2) }{n^3}\right) = \frac{4}{n_{\text{tr}}^{2}} \tr  \big\{ \left( \Sigma D^{-1} \right)^2 \big\}  + O \left( \frac{1}{n^2} \right).
\end{align*}
Under the given assumptions, one can check that $\delta^\top D^{-1} \Sigma D^{-1} \delta = O(n^{-1/2})$ and $\tr\{(\Sigma D^{-1})^2\} = O(d)$. Thus
\begin{align*}
\E [V_{0,\widehat{D}^{-1}}^2] = (I) + (II) + (III) = \frac{1}{4} \left( \delta^\top D^{-1} \delta \right)^2 + O(n^{-1}),
\end{align*}
which yields together with (\ref{Eq: expected value of V}) that $\V[V_{0,\widehat{D}^{-1}}] = O(n^{-1})$. Hence the result follows.

\vskip 1em

\noindent \textbf{$\bullet$ Part 2.}

\vskip .5em

\noindent First calculate the expectation. Conditioned on $\widehat{D}^{-1}$, apply Lemma~\ref{lemma: quadratic form} to have
\begin{align*}
\E[U_{\widehat{D}^{-1}}] & ~=~ \E  \Big[ \E\Big[ \widehat{\delta}^\top  \widehat{D}^{-1} \Sigma  \widehat{D}^{-1} \widehat{\delta} | \widehat{D} \Big] \Big] \\[.5em]
& ~=~ \underbrace{\E \Big[ \delta^\top  \widehat{D}^{-1} \Sigma  \widehat{D}^{-1} \delta   \Big]}_{(I)} + \underbrace{\frac{4}{n_{\text{tr}}}\E \Big[ \tr \left( \widehat{D}^{-1} \Sigma \widehat{D}^{-1} \Sigma \right) \Big]}_{(II)}.
\end{align*}
For $(I)$, by putting $\sigma_{ij} = [\Sigma]_{ij}$, we apply Lemma~\ref{Lemma: sisj expectation} to have
\begin{align*}
(I) & ~=~ \sum_{i=1}^d \sum_{j=1}^d \delta_i \delta_j \sigma_{ij} \E \Bigg[ \frac{1}{s_i^2 s_j^2}\Bigg] =  \delta^\top  D^{-1} \Sigma D^{-1} \delta + O\left( \frac{1}{n^{3/2}} \right).
\end{align*}
For $(II)$, 
\begin{align*}
(II) & ~=~ \frac{4}{n_{\text{tr}}} \sum_{i=1}^d \sum_{j=1}^d \sigma_{ij}^2 \E \Bigg[ \frac{1}{s_i^2 s_j^2} \Bigg] = \frac{4}{n_{\text{tr}}} \tr \left( D^{-1} \Sigma D^{-1} \Sigma \right) + O \left( \frac{1}{n}  \right).
\end{align*}
Therefore
\begin{align*}
\E[U_{\widehat{D}^{-1}}] = \Lambda_{D^{-1},n,d} + O(n^{-1}).
\end{align*}
Next compute the variance of $U_{\widehat{D}^{-1}}$. 
\begin{align} \label{Eq: variance decomposition}
\V[U_{\widehat{D}^{-1}}] = \E \{ \V[ U_{\widehat{D}^{-1}} | \widehat{D}] \} + \V \{ \E[U_{\widehat{D}^{-1}} | \widehat{D}]\}.
\end{align}
Using Lemma~\ref{lemma: quadratic form} and the fact that $\widehat{\delta}$, $\widehat{D}^{-1}$ and $\widehat{\mu}_{\text{pool}}$ are mutually independent (see Part 1),
\begin{align} \label{Eq: V[U|D]}
\V[ U_{\widehat{D}^{-1}}| \widehat{D}] = \frac{32}{n_{\text{tr}}^{2}} \tr \Big\{ \Big( \widehat{D}^{-1} \Sigma \Big)^4  \Big\} + \frac{16}{n_{\text{tr}}} \delta^\top \widehat{D}^{-1} \Sigma  \widehat{D}^{-1} \Sigma \widehat{D}^{-1} \Sigma  \widehat{D}^{-1} \delta.
\end{align}
For the first term, we use Lemma~\ref{Lemma: sisj expectation} to obtain
\begin{align} \nonumber
\E \Bigg[ \frac{32}{n_{\text{tr}}^{2}} \tr \Big\{ \Big( \widehat{D}^{-1} \Sigma \Big)^4  \Big\} \Bigg] & = \frac{32}{n_{\text{tr}}^{2}} \E \Bigg[ \sum_{i=1}^d\sum_{j=1}^d \left( \sum_{k=1}^d \frac{\sigma_{ik} \sigma_{kj}}{s_i^2 s_k^2}  \right)^2 \Bigg] \\[.5em] \nonumber
& = \frac{32}{n_{\text{tr}}^{2}} \tr \Big\{ \Big( D^{-1} \Sigma \Big)^4  \Big\}  + O\left(\frac{\tr\{\Sigma^4\}}{n^2}\right) \\[.5em] \nonumber
& = \frac{32}{n_{\text{tr}}^{2}} \tr \Big\{ \Big( D^{-1} \Sigma \Big)^4  \Big\}  + O\left( \frac{1}{n} \right) \\[.5em] 
& = O\left( \frac{1}{n} \right).  \label{Eq: first term in V}
\end{align}
Similarly for the second term,
\begin{align} \nonumber
\frac{16}{n_{\text{tr}}} \E \Big[\delta^\top \widehat{D}^{-1} \Sigma  \widehat{D}^{-1} \Sigma \widehat{D}^{-1} \Sigma  \widehat{D}^{-1} \delta \Big] & =  \frac{16}{n_{\text{tr}}} \delta^\top {D}^{-1} \Sigma  {D}^{-1} \Sigma {D}^{-1} \Sigma  {D}^{-1} \delta + O\left( \frac{\delta^\top \Sigma^3 \delta}{n^2} \right) \\[.5em]
& = O\left( \frac{1}{n^{3/2}} \right).  \label{Eq: second term in V}
\end{align}
By substituting (\ref{Eq: first term in V}) and (\ref{Eq: second term in V}) into the the expectation of (\ref{Eq: V[U|D]}), we can conclude that
\begin{align*}
\E \{ \V[U_{\widehat{D}^{-1}}| \widehat{D}] \}  = O(n^{-1}).
\end{align*}
Returning to decomposition (\ref{Eq: variance decomposition}) and next focusing on $\V \{ \E[ U_{\widehat{D}^{-1}} | \widehat{D}]\}$, note that
\begin{align*}
\E [U_{\widehat{D}^{-1}}| \widehat{D}] = \delta^\top  \widehat{D}^{-1} \Sigma  \widehat{D}^{-1} \delta + \frac{4}{n_{\text{tr}}} \tr\big\{ (\widehat{D}^{-1} \Sigma)^2 \big\}.
\end{align*}
Thus
\begin{align} \label{Eq: Decomposition of V[E[U|D]]}
\V \{ \E[ U_{\widehat{D}^{-1}} | \widehat{D}]\} ~ \leq ~&  2 \underbrace{\V\Big[  \delta^\top  \widehat{D}^{-1} \Sigma  \widehat{D}^{-1} \delta \Big]}_{(I)'} + 4 \underbrace{\V \Big[ 2n_{\text{tr}}^{-1} \tr\big\{ (\widehat{D}^{-1} \Sigma)^2 \big\} \Big]}_{(II)'}.
\end{align}
For $(I)'$, we use Lemma~\ref{Lemma: sisj expectation} to obtain
\begin{align*}
\E \big[ \big( \delta^\top  \widehat{D}^{-1} \Sigma  \widehat{D}^{-1} \delta  \big)^2 \big] & ~=~ \sum_{i=1}^d \sum_{j=1}^d \sum_{i^\prime = 1}^d \sum_{j^\prime = 1}^d \delta_i \delta_j \delta_{i^\prime} \delta_{j^\prime} \sigma_{ij} \sigma_{i^\prime j^\prime} \E \Bigg[\frac{1}{s_i^2 s_j^2 s_{i^\prime}^2 s_{j^\prime}^2 }\Bigg] \\[.5em]
& ~=~ \big( \delta^\top  D^{-1} \Sigma  D^{-1} \delta  \big)^2 + O\left( \frac{ \big( \delta^\top \Sigma \delta  \big)^2 }{n} \right) \\[.5em]
& ~=~ \big( \delta^\top  D^{-1} \Sigma  D^{-1} \delta  \big)^2 + O \left( \frac{1}{n^2} \right),
\end{align*}
and
\begin{align*}
\E \big[  \delta^\top  \widehat{D}^{-1} \Sigma  \widehat{D}^{-1} \delta  \big] & ~=~ \sum_{i=1}^d \sum_{j=1}^d \delta_i \delta_j \sigma_{ij} \E\Bigg[ \frac{1}{s_i^2 s_j^2} \Bigg] \\[.5em]
& ~=~ \delta^\top  D^{-1} \Sigma D^{-1} \delta + O\left( \frac{\delta^\top \Sigma \delta}{n} \right) \\[.5em]
& ~=~ \delta^\top  D^{-1} \Sigma D^{-1} \delta  + O \left( \frac{1}{n^{3/2}} \right).
\end{align*}
Therefore, $(I)^\prime =\E \big[ \big( \delta^\top  \widehat{D}^{-1} \Sigma  \widehat{D}^{-1} \delta  \big)^2 \big]  - \big\{ \E \big[  \delta^\top  \widehat{D}^{-1} \Sigma  \widehat{D}^{-1} \delta  \big] \big\}^2  =  O(n^{-2})$. 

\vskip .8em

\noindent Moving onto $(II)^\prime$, we have
\begin{align*}
\E \big[ \big(2n_{\text{tr}}^{-1} \tr\big\{ (\widehat{D}^{-1} \Sigma)^2 \big\} \big)^2 \big] & ~=~ 4n_{\text{tr}}^{-2} \sum_{i=1}^d \sum_{j=1}^d \sum_{i^\prime = 1}^d \sum_{j^\prime = 1}^d \sigma_{ij}^2 \sigma_{i^\prime j^\prime}^2 \E \Bigg[ \frac{1}{s_i^2 s_j^2 s_{i^\prime}^2 s_{j^\prime}^2} \Bigg] \\[.5em]
& ~=~ 4n_{\text{tr}}^{-2} \big[ \tr\big\{ (D^{-1} \Sigma)^2 \big\}  \big]^2 + O \left( \frac{\big\{ \tr ( \Sigma^2) \big\}^2}{n^3} \right) \\[.5em]
& ~=~ 4n_{\text{tr}}^{-2} \big[ \tr\big\{ (D^{-1} \Sigma)^2 \big\}  \big]^2 + O(n^{-1}),
\end{align*}
and
\begin{align*}
\E \big[ 2n_{\text{tr}}^{-1} \tr\big\{ (\widehat{D}^{-1} \Sigma)^2 \big\} \big]  & ~=~ 2n_{\text{tr}}^{-1} \sum_{i=1}^d \sum_{j=1}^d \sigma_{ij}^2 \E \Bigg[ \frac{1}{s_i^2 s_j^2 } \Bigg] \\[.5em]
& ~=~ 2n_{\text{tr}}^{-1}  \tr\big\{ (D^{-1} \Sigma)^2 \big\} + O\left( \frac{\tr\{ \Sigma^2\} }{n^2} \right) \\[.5em]
& ~=~ 2n_{\text{tr}}^{-1}  \tr\big\{ (D^{-1} \Sigma)^2 \big\} + O(n^{-1}).
\end{align*}
Hence $(II)^\prime = O(n^{-1})$. Substituting the bounds $(I)' = O(n^{-2})$ and $(II)^\prime = O(n^{-1})$ into the right-hand side of (\ref{Eq: Decomposition of V[E[U|D]]}), we obtain that
\begin{align*}
\V[U_{\widehat{D}^{-1}}] = O(n^{-1}).
\end{align*}
This verifies the second part.

\vskip 1em

\noindent \textbf{$\bullet$ Part 3.}

\vskip .5em

\noindent Let us start with the expectation. Based on the fact that $\widehat{\delta}$, $\widehat{D}^{-1}$ and $\widehat{\mu}_{\text{pool}}$ are mutually independent (see Part 1),
\begin{align*}
\E[V_{0,\widehat{D}^{-1}}+ V_{1,\widehat{D}^{-1}}] & = \E\left[ \widehat{\delta}^\top \widehat{D}^{-1} (\mu_0 - \widehat{\mu}_{\text{pool}}) + \widehat{\delta}^\top \widehat{D}^{-1} (\widehat{\mu}_{\text{pool}} - \mu_1 )  \right] \\[.5em]
& = - \delta^\top D^{-1} \delta \cdot \big\{1 + O(n^{-1}) \big\}.
\end{align*}
Next calculate the second moment.
\begin{align*}
\E[(V_{0,\widehat{D}^{-1}}+ V_{1,\widehat{D}^{-1}})^2] & ~=~ \E \left[ \tr \left( \widehat{\delta} \widehat{\delta}^\top \widehat{D}^{-1} \delta \delta^\top \widehat{D}^{-1}\right)  \right] \\[.5em]
& ~=~ \E \left[ \tr \Big\{ \left( \delta\delta^\top + 4n_{\text{tr}}^{-1}\Sigma \right) \widehat{D}^{-1} \delta \delta^\top \widehat{D}^{-1}  \Big\} \right] \\[.5em]
& ~=~ \underbrace{\E \left[  \left( \delta^\top \widehat{D}^{-1} \delta \right)^2 \right]}_{(I)''}  ~ + ~ \underbrace{4n_{\text{tr}}^{-1} \E \left[  \delta^\top \widehat{D}^{-1} \Sigma \widehat{D}^{-1} \delta \right]}_{(II)''}.
\end{align*}
For $(I)^{\prime \prime}$, applying Lemma~\ref{Lemma: sisj expectation} yields
\begin{align*}
(I)''  & ~ = ~ \sum_{i=1}^d \sum_{j=1}^d \sum_{i^\prime = 1}^d \sum_{j^\prime = 1}^d \delta_i \delta_j \delta_{i^\prime} \delta_{j^\prime} \E \left[  \frac{1}{s_i^2 s_j^2 s_{i^\prime}^2 s_{j^\prime}^2 } \right] \\[.5em]
& ~=~ \sum_{i=1}^d \sum_{j=1}^d \sum_{i^\prime = 1}^d \sum_{j^\prime = 1}^d \delta_i \delta_j \delta_{i^\prime} \delta_{j^\prime}  \left[  \frac{1}{\sigma_i^2 \sigma_j^2 \sigma_{i^\prime}^2 \sigma_{j^\prime}^2 } + O(n^{-1})  \right] \\[.5em]
& ~=~ \left( \delta^\top D^{-1} \delta \right)^2 + \left( \delta^\top \delta \right)^2 \cdot O(n^{-1}).
\end{align*}
Similarly, for $(II)^{\prime \prime}$, Lemma~\ref{Lemma: sisj expectation} yields
\begin{align*}
(II)'' ~ =  ~ 4n_{\text{tr}}^{-1} \delta^\top D^{-1} \Sigma D^{-1} \delta + \delta^\top \Sigma \delta \cdot O(n^{-2}).
\end{align*}
Since the eigenvalues of $\Sigma$ are uniformly bounded and $\delta^\top \Sigma^{-1} \delta = O(n^{-1/2})$, the variance is bounded by 
\begin{align*}
\V [ V_{0,\widehat{D}^{-1}}+ V_{1,\widehat{D}^{-1}}] & = \left(  \delta^\top D^{-1} \delta \right)^2 \cdot O(n^{-1}) +  \left( \delta^\top \delta \right)^2 \cdot O(n^{-1}) \\[.5em]
& ~~~~ +  4n_{\text{tr}}^{-1} \delta^\top D^{-1} \Sigma D^{-1} \delta + \delta^\top \Sigma \delta \cdot O(n^{-2}) \\[.5em]
& = O \left( \frac{1}{n^{3/2}} \right).
\end{align*}
This verifies the third part.

\vskip 1em

\noindent \textbf{$\bullet$ Concluding the proof.}

\vskip .5em

\noindent Consider a bivariate function $f(v,u) = \Phi(v/\sqrt{u})$. Then by the Taylor expansion of $f(v,u)$ around $(\Psi_{D^{-1},n,d},\Lambda_{D^{-1},n,d})$ together with the results in Part 1 and Part 2, we have
\begin{align*}
& \mathcal{E}_{0,\widehat{D}^{-1}} = f(V_{0,\widehat{D}^{-1}},U_{\widehat{D}^{-1}})  \\[.5em]
= ~ &\Phi \Bigg(  \frac{\Psi_{D^{-1},n,d}}{\sqrt{ \Lambda_{D^{-1},n,d}}} \Bigg)  + \phi\Bigg(  \frac{\Psi_{D^{-1},n,d}}{ \sqrt{\Lambda_{D^{-1},n,d} }} \Bigg) \frac{1}{\sqrt{ \Lambda_{D^{-1},n,d} }} (V_{0,\widehat{D}^{-1} } - \Psi_{D^{-1},n,d}) \\[.5em]
- & \phi\Bigg(  \frac{\Psi_{D^{-1},n,d}}{ \sqrt{\Lambda_{D^{-1},n,d} }} \Bigg) \frac{\Psi_{D^{-1},n,d}}{(\Lambda_{D^{-1},n,d} )^{3/2}} (U_{ \widehat{D}^{-1} } - \Lambda_{D^{-1},n,d}) + O_P\left( n^{-1} \right),
\end{align*}
where $\phi(\cdot)$ is the density function of $N(0,1)$. Similarly,
\begin{align*}
& \mathcal{E}_{1, \widehat{D}^{-1} } = f(V_{1, \widehat{D}^{-1} },U_{\widehat{D}^{-1}} ) \\[.5em] 
=~ &\Phi \Bigg(  \frac{\Psi_{D^{-1},n,d}}{\sqrt{ \Lambda_{D^{-1},n,d}  }} \Bigg)  + \phi\Bigg(  \frac{\Psi_{D^{-1},n,d}}{ \sqrt{\Lambda_{D^{-1},n,d} }} \Bigg) \frac{1}{\sqrt{ \Lambda_{D^{-1},n,d} }} (V_{1,\widehat{D}^{-1}} - \Psi_{D^{-1},n,d}) \\[.5em]
- & \phi\Bigg(  \frac{\Psi_{D^{-1},n,d}}{ \sqrt{\Lambda_{D^{-1},n,d} }} \Bigg) \frac{\Psi_{D^{-1},n,d}}{(\Lambda_{D^{-1},n,d} )^{3/2}} (U_{\widehat{D}^{-1}} - \Lambda_{D^{-1},n,d}) + O_P\left( n^{-1} \right).
\end{align*}
Combining these approximations with the result in Part 3,
\begin{align*}
& \frac{\mathcal{E}_{0,\widehat{D}^{-1}} + \mathcal{E}_{1,\widehat{D}^{-1}}}{2} \\[.5em]
=~ & \Phi \Bigg(  \frac{\Psi_{D^{-1},n,d}}{\sqrt{ \Lambda_{D^{-1},n,d}  }} \Bigg) + \phi\Bigg(  \frac{\Psi_{D^{-1},n,d}}{ \sqrt{\Lambda_{D^{-1},n,d} }} \Bigg) \frac{1}{\sqrt{ \Lambda_{D^{-1},n,d} }} \left(\frac{V_{0,\widehat{D}^{-1}} + V_{1,\widehat{D}^{-1}}}{2} - \Psi_{D^{-1},n,d} \right)  \\[.5em]
- & \phi\Bigg(  \frac{\Psi_{D^{-1},n,d}}{ \sqrt{\Lambda_{D^{-1},n,d} }} \Bigg) \frac{\Psi_{D^{-1},n,d}}{(\Lambda_{D^{-1},n,d} )^{3/2}} (U_{\widehat{D}^{-1}} - \Lambda_{D^{-1},n,d}) + O_P\left( n^{-1} \right) \\[.5em]
= ~ & \Phi \Bigg(  \frac{\Psi_{D^{-1},n,d}}{\sqrt{ \Lambda_{D^{-1},n,d}  }} \Bigg)  + O_P\left(\frac{1}{n^{3/4}} \right) .
\end{align*}
This completes the proof of Lemma~\ref{lemma: Dinv approximations}.

\vskip 2em

\subsection{Proof of Theorem~\ref{thm: power under elliptical distributions}} \label{sec: power under elliptical distributions}

The proof of Theorem~\ref{thm: power under elliptical distributions} basically follows the same lines of arguments as in the proof of Theorem~\ref{thm: power of g-LDA} under the given assumptions. However we note that the proof of Theorem~\ref{thm: power of g-LDA} relies on Lemma~\ref{lemma: quadratic form}, which is tailored to the normality assumption. Hence, in order to complete the proof, we need to verify that the parts that build on Lemma~\ref{lemma: quadratic form} are also valid for elliptical distributions. More specifically there are two parts that depend on Lemma~\ref{lemma: quadratic form}:  (i)~the approximations of $V_{0,A}$, $V_{1,A}$ and $V_{0,A} + V_{1,A}$ given in (\ref{eq: approximation 1}) and (ii)~the approximation of $U_A$ given in (\ref{eq: approximation 2}). In the rest of the proof, we prove that these approximations are still valid for elliptical distributions.

\paragraph{$\bullet$ Moments of elliptical distributions.} Let us start with some useful moment expressions of an elliptical random vector.
\begin{lemma}[Chapter 3.2 of \cite{mathai2012bilinear}] \label{Lemma: moments of elliptical distributions}
	Suppose that $Z = (Z_1,\ldots,Z_d)^\top \in \mathbb{R}^d$ has a multivariate elliptical distribution with parameters $(\mu, S, \xi)$ where $\mu = (\mu_1,\ldots,\mu_{d})^\top$ and $[\Sigma]_{jk} = -2\xi'(0) [S]_{jk} = \sigma_{jk}$ for $j,k = 1,\ldots, d$ such that
	\begin{align*}
	\E \big[e^{it^\top Z} \big] = e^{it^\top \mu} \xi \big( t^\top S t \big) \quad \text{for all $t \in \mathbb{R}^d$.}
	\end{align*}
	Then we have
	\begin{enumerate}
		\item $\E[Z_j] = \mu_{j}$,
		\item $\E[Z_j Z_k] = \mu_{j}\mu_{k} + \sigma_{jk}$,
		\item $\E[Z_jZ_kZ_l] = \mu_{j} \mu_{k} \mu_{l} + \mu_{j} \sigma_{lk} + \mu_{k} \sigma_{jl} + \mu_{l} \sigma_{jk}$.
	\end{enumerate}
	Moreover for a symmetric matrix $A$, we have
	\begin{enumerate}
		\item $\E[Z^\top A Z] = \mu^\top A \mu + \text{\emph{tr}}(A\Sigma)$,
		\item $\V[Z^\top A Z] = 4  \mu^\top A \Sigma A\mu + \zeta_{\text{\emph{kurt}}} \{\text{\emph{tr}}(A\Sigma) \}^2 + 2 (\zeta_{\text{\emph{kurt}}} + 1) \text{\emph{tr}}\{(A\Sigma)^2\}$,
	\end{enumerate}
	where
	\begin{align*} 
	\zeta_{\text{\emph{kurt}}} = \frac{\xi^{''}(0)}{\{\xi'(0)\}^2} - 1 = \frac{\E\big[ \big\{(Z-\mu)^\top \Sigma^{-1} (Z - \mu) \big\}^2\big]}{d(d+2)} - 1.
	\end{align*} 
\end{lemma}
Note that when $Z$ has an multivariate normal distribution, the kurtosis parameter becomes $\zeta_{\text{kurt}} = 0$ and the above result coincides with Lemma~\ref{lemma: quadratic form}.

\paragraph{$\bullet$ Part 1. Approximation (\ref{eq: approximation 1})}Leveraging Lemma~\ref{Lemma: moments of elliptical distributions}, we first prove that the approximations of $V_{0,A}= \widehat{\delta}^\top A (\mu_0 - \widehat{\mu}_{\text{pool}}),V_{1,A}= \widehat{\delta}^\top A (\widehat{\mu}_{\text{pool}} -\mu_1)$ and $V_{0,A} + V_{1,A} = \widehat{\delta}^\top A (\mu_0 -\mu_1)$ in (\ref{eq: approximation 1}) hold true for elliptical distributions under \textbf{(A7)}. By assuming $n_{0,\text{tr}} = n_{1,\text{tr}}$, it is straightforward to see that the expected values of these quantities are 
\begin{align*}
& \E[V_{0,A}] = \E[V_{1,A}] = - \frac{1}{2} \delta^\top A \delta \quad \text{and} \\[.5em]
& \E[V_{0,A} + V_{1,A}] = - \delta^\top A \delta.
\end{align*}
Turning to the variances, we shall prove that $\V[V_{0,A}] = O(n^{-1})$, $\V[V_{1,A}] = O(n^{-1})$ and $\V[V_{0,A} + V_{1,A}] = O(n^{-3/2})$, which in turn yields the claim~(\ref{eq: approximation 1}). 
Focusing on the variance of $V_{0,A}$ and using $n_{0,\text{tr}} = n_{1,\text{tr}}$, we see that
\begin{align*}
\V[V_{0,A}] ~ = ~ & \frac{1}{n_{0,\text{tr}}^4} \V \Bigg[  \sum_{i=1}^{n_{0,\text{tr}}} \Bigg\{ \big( X_i - Y_i  \big)^\top A \bigg( \mu_0 - \frac{1}{2}X_i  - \frac{1}{2} Y_i \bigg) \Bigg\} \\[.5em] 
&~~~~~~~~~ + \sum_{1 \leq i\neq j \leq n_{0,\text{tr}} } \Bigg\{ \big( X_i - Y_i  \big)^\top A \bigg( \mu_0 - \frac{1}{2}X_j  - \frac{1}{2} Y_j \bigg) \Bigg\}\Bigg] \\[.5em]
\leq ~ & \underbrace{\frac{2}{n_{0,\text{tr}}^4} \V \Bigg[  \sum_{i=1}^{n_{0,\text{tr}}} \Bigg\{ \big( X_i - Y_i  \big)^\top A \bigg( \mu_0 - \frac{1}{2}X_i  - \frac{1}{2} Y_i \bigg) \Bigg\}  \Bigg]}_{(I)} \\[.5em]
+ & \underbrace{\frac{2}{n_{0,\text{tr}}^4} \V \Bigg[ \sum_{1 \leq i\neq j \leq n_{0,\text{tr}} } \Bigg\{ \big( X_i - Y_i  \big)^\top A \bigg( \mu_0 - \frac{1}{2}X_j  - \frac{1}{2} Y_j \bigg) \Bigg\}\Bigg]}_{(II)}
\end{align*}
where the last inequality follows by $\V[X+Y] \leq 2\V[X] + 2\V[Y]$. 
For the first term $(I)$, since we assume $\mathcal{X}_0^{n_0}$ and $\mathcal{Y}_0^{n_1}$ are mutually independent, we have
\begin{align*}
(I) ~=~ & \frac{1}{2n_{0,\text{tr}}^3} \V \big\{ (X_1 - Y_1)^\top A (2\mu_0 - X_1 - Y_1) \big\} \\[.5em] 
=~ & \frac{1}{n_{0,\text{tr}}^3} \big[ 2(\zeta_{\text{kurt}} + 1) \text{tr} \big\{(A\Sigma)^2 \big\} + \zeta_{\text{kurt}} \big\{ \text{tr}(A \Sigma) \big\}^2 + 2\delta^\top A \Sigma A \delta  \big],
\end{align*}
where the second equality follows by straightforward calculation using Lemma~\ref{Lemma: moments of elliptical distributions}. Thus under the given conditions, we have established that $(I) = O(n^{-1})$. 
For the second term $(II)$, by expanding the variance of the sum of random variables, we see that
\begin{align*}
& (II) \\[.5em] 
~=~ & O(n^{-2}) \cdot \text{Cov}\big\{ (X_1 - Y_1) ^\top A (2\mu_0 - X_2 - Y_2), \ (X_1 - Y_1) ^\top A (2\mu_0 - X_2 - Y_2) \big\} \\[.5em]
+ ~ & O(n^{-2}) \cdot \text{Cov}\big\{ (X_1 - Y_1) ^\top A (2\mu_0 - X_2 - Y_2), \ (X_2 - Y_2) ^\top A (2\mu_0 - X_1 - Y_1) \big\} \\[.5em]
+ ~ & O(n^{-1}) \cdot \text{Cov}\big\{ (X_1 - Y_1) ^\top A (2\mu_0 - X_2 - Y_2), \ (X_2 - Y_2) ^\top A (2\mu_0 - X_3 - Y_3) \big\} \\[.5em]
+ ~ & O(n^{-1}) \cdot \text{Cov}\big\{ (X_1 - Y_1) ^\top A (2\mu_0 - X_2 - Y_2), \ (X_3 - Y_3) ^\top A (2\mu_0 - X_1 - Y_1) \big\} \\[.5em]
\defn ~ & O(n^{-2}) \cdot (II_1) + O(n^{-2}) \cdot (II_2) + O(n^{-1}) \cdot (II_3) + O(n^{-1}) \cdot (II_4).
\end{align*}
Again, leveraging Lemma~\ref{Lemma: moments of elliptical distributions}, it can be seen that 
\begin{align*}
& (II_1) ~=~ 4 \delta^\top A \Sigma A \delta + 4 \text{tr}\{(A\Sigma)^2\}, \\[.5em]
& (II_2) ~=~ - 4\delta^\top A \Sigma A \delta + 4 \text{tr}\{(A\Sigma)^2\}, \\[.5em]
& (II_3) ~=~ 2 \delta^\top A \Sigma A \delta \quad \text{and} \\[.5em]
& (II_4) ~=~ 2 \delta^\top A \Sigma A \delta. 
\end{align*}
Thus, under the given conditions, we can conclude that $\V[V_{0,A}]= O(n^{-1})$. By symmetry we similarly have $\V[V_{1,A}] = O(n^{-1})$. 
For the last quantity $V_{0,A} + V_{1,A}$,
\begin{align*}
\V[V_{0,A} + V_{1,A}] ~=~ & \V[\widehat{\delta}^\top A (\mu_0 -\mu_1)]  = \frac{1}{n_{0,\text{tr}}^2} \V[(X_1-Y_1)^\top A(\mu_0 - \mu_1)] \\[.5em]
~=~ &  \frac{2}{n_{0,\text{tr}}^2} \delta^\top A \Sigma A \delta = O(n^{-3/2}).
\end{align*}
Combining the pieces together proves the validity of the approximations (\ref{eq: approximation 1}).

\vskip 1em

\paragraph{$\bullet$ Part 2. Approximation (\ref{eq: approximation 2})} Recall that $U_A = \widehat{\delta}^\top A \Sigma A \widehat{\delta}$ and it is relatively straightforward to compute the expectation under $n_{0,\text{tr}} =  n_{1,\text{tr}}$ as
\begin{align*}
\E[U_A] ~=~ \delta^\top A \Sigma A \delta + \frac{2}{n_{0,\text{tr}}^2} \text{tr} \big\{(A\Sigma)^2 \big\}.
\end{align*}
Therefore it is enough to show that the variance of $U_A$ is $O(n^{-1})$, which in turns proves the claim~(\ref{eq: approximation 2}). Similarly as before in part 1, we can upper bound the variance of $U_A$ by
\begin{align*}
\V[U_A] ~ \leq ~ &  \underbrace{\frac{2}{n_{0,\text{tr}}^3}\V \big\{ (X_1 - Y_1)^\top A \Sigma A (X_1 - Y_1) \big\}}_{(I)} \\[.5em] 
+ & \underbrace{\frac{2}{n_{0,\text{tr}}^4} \V \Bigg\{ \sum_{1 \leq i \neq j \leq n_{0,\text{tr}}} (X_i - Y_i)^\top A \Sigma A (X_j - Y_j) \Bigg\}}_{(II)}.
\end{align*} 
For the first term $(I)$, we observe that by the independence between $X_1$ and $Y_1$, the characteristic function of $Z_1 \defn X_1 - Y_1$ is
\begin{align*}
\E \big[e^{it^\top Z_1} \big] = e^{i t^\top \delta } \xi^2(t^\top S t). 
\end{align*}
In other words, $Z_1$ has an elliptical distribution with parameters $(\delta, S, \xi^2)$. Also the corresponding covariance matrix and the kurtosis parameter of $Z_1$ are $2\Sigma$ and $\zeta_{\text{kurt}}/2$, respectively. Then using Lemma~\ref{Lemma: moments of elliptical distributions} yields
\begin{align*}
(I) ~=~ & \frac{4}{n_{0,\text{tr}}^3} \big[ 4 \delta^\top A \Sigma A \Sigma A \Sigma A \delta +  \zeta_{\text{kurt}} \{\text{tr}(A\Sigma A\Sigma
) \}^2 + 2(\zeta_{\text{kurt}}  + 2) \text{tr}\{(A\Sigma)^4 \} \big] \\[.5em] 
= ~ & O(n^{-1}).
\end{align*}
Let $Z_2,Z_3$ be independent copies of $Z_1$. Then for the second term $(II)$, 
\begin{align*}
(II) ~=~ & O(n^{-2}) \cdot \underbrace{\text{Cov} \big\{Z_1^\top A\Sigma A Z_2, \ Z_1^\top A \Sigma A Z_2 \big\}}_{(II_1)} \\[.5em]
+ & O(n^{-1}) \cdot \underbrace{\text{Cov} \big\{Z_1^\top A\Sigma A Z_2, \ Z_1^\top A \Sigma A Z_3 \big\}}_{(II_2)}.
\end{align*}
Building on Lemma~\ref{Lemma: moments of elliptical distributions}, it can be shown that 
\begin{align*}
(II_1) ~=~& 4 \delta^\top A \Sigma A \Sigma A\Sigma A\delta + 4 \text{tr}\{ (A\Sigma)^4 \} \\[.5em]
(II_2) ~=~&  2 \delta^\top A \Sigma A \Sigma A\Sigma A\delta. 
\end{align*}
Therefore the second term also satisfies $(II) = O(n^{-1})$, which verifies the claim~(\ref{eq: approximation 2}). This completes the proof of Theorem~\ref{thm: power under elliptical distributions}.

\vskip 2em

\subsection{Proof of Proposition~\ref{Proposition: Asymptotic test}} \label{sec: proof of Proposition: Asymptotic test}
We let denote the conditional expectations of $\widehat E^S_{0}(\widehat{C})$ and $\widehat E^S_{1}(\widehat{C})$ given the training set by
\begin{align*}
& \mathcal{E}_{0}(\widehat{C}) \defn  \Pr_{Z \sim \dP_0}\Big( \widehat{C}(Z) = 1 ~|~ \mathcal{X}_1^{n_{0,\text{tr}}}, \mathcal{Y}_1^{n_{1,\text{tr}}}\Big) \quad \text{and} \\[.5em]
& \mathcal{E}_{1}(\widehat{C}) \defn \Pr_{Z \sim \dP_1} \Big( \widehat{C}(Z) = 0 ~|~ \mathcal{X}_1^{n_{0,\text{tr}}},  \mathcal{Y}_1^{n_{1,\text{tr}}} \Big).
\end{align*}
For the rest of the proof, we omit the dependence of $\widehat{C}$ on the classification errors to simplify the notation.

Now, since $\widehat E^S_{0}$ and $\widehat E^S_{1}$ are uniformly bounded, the convergence in probability implies that the convergence in moment. Hence we have that $\E\big[\widehat E^S_{0}\big] \rightarrow E_{0}$ and $\E\big[\widehat E^S_{1}\big] \rightarrow E_{1}$, which implies $\mathcal{E}_{0} \overset{p}{\longrightarrow} E_{0}$ and $\mathcal{E}_{1} \overset{p}{\longrightarrow} E_{1}$ using Markov's inequality. Consequently,
\begin{align} \label{Eq: Ratio convergence}
\frac{\widehat E^S_{0} (1-\widehat E^S_{0} ) \big/n_{0,\text{te}}  + \widehat E^S_{1} (1-\widehat E^S_{1}) \big/ n_{1,\text{te}}}{\mathcal{E}_{0} ( 1-\mathcal{E}_{0} ) \big/n_{0,\text{te}}  + \mathcal{E}_{1} (1-\mathcal{E}_{1} ) \big/ n_{1,\text{te}}}  \overset{p}{\longrightarrow} 1.
\end{align}
Suppose that the null hypothesis is true. Then under the given conditions, following the same lines of the proof of Proposition~\ref{proposition: conditional CLT} yields
\begin{align*}
\frac{2\widehat E^S - 1}{\sqrt{\mathcal{E}_{0} ( 1- \mathcal{E}_{0} ) \big/n_{0,\text{te}}  + \mathcal{E}_{1} ( 1- \mathcal{E}_{1} ) \big/ n_{1,\text{te}} } } \convD N(0,1),
\end{align*}
where we use the fact that $\mathcal{E}_{0} + \mathcal{E}_{1} = 1$ under the null hypothesis. It is worth mentioning that Proposition~\ref{proposition: conditional CLT} also requires \textbf{(A1)}, \textbf{(A2)}, \textbf{(A5)} and \textbf{(A6)}. These assumptions are made to show that $\mathcal{E}_{0,A}$ and $\mathcal{E}_{1,A}$ are asymptotically bounded below by 0 and above by 1, which are guaranteed by the assumption~\textbf{(A9)} under the current setting.

Next Slutsky's theorem together with the observation~(\ref{Eq: Ratio convergence}) further shows that
\begin{align*}
\frac{2\widehat E^S - 1}{\sqrt{\widehat E^S_{0} ( 1-\widehat E^S_{0} ) \big/n_{0,\text{te}}  + \widehat E^S_{1} ( 1-\widehat E^S_{1} ) \big/ n_{1,\text{te}}} } \convD N(0,1).
\end{align*}
Therefore $\varphi_{\widehat{C},\text{Asymp}}$ asymptotically controls the type-1 error rate under the given conditions. In terms of power, the assumption~\textbf{(A9)} guarantees that $ 2\widehat E^S - 1 \overset{p}{\longrightarrow} - 2\epsilon < 0$ and 
\begin{align*}
\widehat E^S_{0} ( 1-\widehat E^S_{0} ) \big/n_{0,\text{te}}  + \widehat E^S_{1} (1-\widehat E^S_{1} ) \big/ n_{1,\text{te}} \overset{p}{\longrightarrow} 0.
\end{align*}
Building on this observation, we have under the alternative that
\begin{align*}
& \E_{H_1} \big[\varphi_{\widehat{C},\text{Asymp}}\big] \\[.5em]
= ~ & \mathbb{P}_{H_1} \left[ \frac{2\widehat E^S - 1}{\sqrt{\widehat E^S_{0} (1-\widehat E^S_{0}) \big/n_{0,\text{te}}  + \widehat E^S_{1} ( 1-\widehat E^S_{1}) \big/ n_{1,\text{te}} } } <  -z_\alpha \right] \\[.5em]
= ~& \mathbb{P}_{H_1} \Bigg[ 2\widehat E^S  - 1 < - z_\alpha \sqrt{\widehat E^S_{0} ( 1-\widehat E^S_{0} ) \big/n_{0,\text{te}}  + \widehat E^S_{1}  ( 1-\widehat E^S_{1} ) \big/ n_{1,\text{te}}} \Bigg] \\[.5em] 
\rightarrow ~ & 1,
\end{align*}
which proves consistency of the asymptotic test.

\vskip 2em

\subsection{Proof of Theorem~\ref{Theorem: permutation test}} \label{sec: Proof of Theorem: permutation test}

As mentioned in the main text, both half- and entire-permutation methods yield a valid level $\alpha$ test \citep[see, e.g., Theorem 1 of][]{hemerik2018false}. Hence we focus on proving consistency of the resulting test under the given conditions. To ease notation, we drop the dependence of $\widehat{C}$ on the sample-splitting errors throughout this proof.

Let us consider all possible permutations first, that is $m! \defn n_{\text{te}}!$ for method 1 and $m! \defn n!$ for method 2, and denote the sample-splitting errors (or $1-\text{accuracies}$) by $\widehat E^{S,1},\ldots,\widehat E^{S,m!}$ computed based on each permutation. We then let $\widetilde{E}^{S,1},\ldots,\widetilde{E}^{S,P}$ be $P$ independent samples from $\widehat E^{S,1},\ldots,\widehat E^{S,m!}$ without replacement. Then the permutation test can be equivalently written as
\begin{align} \label{Eq: permutation test}
\varphi_{\widehat{C},\text{Perm}} = \I \left[ \frac{1}{P} \sum_{i=1}^{P} \I \left( \widehat{E}^{S} < \widetilde{E}^{S,i} \right) \geq 1 - \alpha_P  \right],
\end{align}
where $1- \alpha_P \defn \lceil(1-\alpha)(1+P) \rceil / P \rightarrow 1 - \alpha$ as $P \rightarrow \infty$. We note that in order for the test (\ref{Eq: permutation test}) to have power, $1- \alpha_P$ should be less than one (otherwise the test function is always zero), which requires the condition $P > (1-\alpha)/\alpha$.

Let us denote the $\alpha_P$ quantile of $\widetilde{E}^{S,1},\ldots,\widetilde{E}^{S,P}$ by $q_{\alpha_P}$. Using the representation~(\ref{Eq: permutation test}), it can be verified that if the test statistic is less than this quantile, i.e.~$\widehat{E}^{S} < q_{\alpha_P}$, then the permutation test is equal to one, i.e.~$\varphi_{\widehat{C},\text{Perm}} = 1$. This fact implies that if $\I(\widehat{E}^{S} < q_{\alpha_P})$ is a consistent test, then the permutation test is also consistent. Therefore it is enough to work with $\I(\widehat{E}^{S} < q_{\alpha_P})$ and show that it is consistent.

A high-level proof strategy is as follows. By the assumption, $\widehat{E}^S$ converges in probability to a constant strictly less than $1/2 -\epsilon/2$ under the alternative. Therefore the proof is complete if we show that a lower bound for $q_{\alpha_P}$ converges to a constant that is strictly larger than $1/2 -\epsilon/2$. To do so, we let $\mathfrak{n}$ be a random variable uniformly distributed over $\{1,\ldots,P\}$ and write the distribution of $\mathfrak{n}$ by $\mathbb{P}_\mathfrak{n}$ (conditional on everything else) and the expectation with respect to $\mathbb{P}_\mathfrak{n}$ by $\mathbb{E}_\mathfrak{n}$.

For a given $t \in (0, 1/2)$, applying Markov's inequality yields
\begin{align*}
\mathbb{P}_{\mathfrak{n}} \Big( \widetilde{E}^{S,\mathfrak{n}} < t \Big) ~=~& \mathbb{P}_{\mathfrak{n}} \Big( - \widetilde{E}^{S,\mathfrak{n}} + 1/2 >  - t + 1/2 \Big) \\[.5em]
\leq ~ &  \mathbb{P}_{\mathfrak{n}} \Big( \big| \widetilde{E}^{S,\mathfrak{n}} - 1/2 \big| >  - t + 1/2 \Big) \\[.5em]
\leq ~ &  \frac{ \E_{\mathfrak{n}} \big[ \big( \widetilde{E}^{S,\mathfrak{n}} - 1/2 \big)^2 \big]}{(1/2 -t)^2}.
\end{align*}
Now by setting the right-hand side to be $\alpha_P$, we know that the quantile $q_{\alpha_P}$ is lower bounded by 
\begin{align*}
q_{\alpha_P} ~\geq~ \frac{1}{2} - \sqrt{\frac{1}{\alpha_P} \E_{\mathfrak{n}}\big[ \big( \widetilde{E}^{S,\mathfrak{n}} - 1/2 \big)^2 \big]}.
\end{align*}
Here the expected value of the squared difference is
\begin{align} \label{Eq: squared difference}
& \E_{\mathfrak{n}}\big[ \big( \widetilde{E}^{S,\mathfrak{n}} - 1/2 \big)^2\big] = \frac{1}{P} \sum_{i=1}^P \left( \widetilde{E}^{S,i}  -  1/2  \right)^2.
\end{align}
In the rest of the proof, we show that the above quantity converges in probability to zero as $n \rightarrow \infty$ for both method 1 and method 2. Hence the quantile $q_{\alpha_P}$ is lower bounded by $1/2 - \epsilon/2$ in the limit as claimed.

\vskip 1em

\noindent \textbf{$\bullet$ Method 1 (Half-permutation test).}

\vskip .5em		

\noindent To start with method 1, we let $\mathfrak{m}$ be a random variable uniformly distributed over $\{1,\ldots,n_{\text{te}}! \}$ and write the expectation and the variance over $\mathfrak{m}$ (conditional on everything else) by $\mathbb{E}_\mathfrak{m}$ and $\mathbb{V}_{\mathfrak{m}}$, respectively. We note that for each $i \in \{1,\ldots,P\}$, $ \widetilde{E}^{S,i}$ has the same distribution as $\widehat E^{S,\mathfrak{m}}$ and that the expected value of $\widehat E^{S,\mathfrak{m}}$ is calculated as
\begin{align} \label{Eq: expectation under permutation}
\E_{\mathfrak{m}}\big[\widehat E^{S,\mathfrak{m}}\big] = \frac{1}{n_{\text{te}}!} \sum_{i=1}^{n_{\text{te}}!} \widehat E^{S,i} = \frac{1}{2}.
\end{align}		
Therefore the squared difference~(\ref{Eq: squared difference}) is an unbiased estimator of the variance of $\widehat E^{S,\mathfrak{m}}$.

We next upper bound the variance of $\widehat E^{S,\mathfrak{m}}$. To do so, let us write the test set by 
\begin{align*}
\{X_{1 + n_{0,\text{tr}}},\ldots,X_{n_{0}},Y_{1+n_{1,\text{tr}}},\ldots, Y_{n_{1}} \} \defn \{Z_1,\ldots,Z_{n_\text{te}}\}.
\end{align*}
Notice that for each $\mathfrak{m}$, there exists the corresponding permutation of $\{1,\ldots,n_{\text{te}} \}$, denoted by $\omega^{\mathfrak{m}} \defn \{ \omega_1^{\mathfrak{m}},\ldots, \omega_{n_\text{te}}^{\mathfrak{m}} \}$, such that the test statistic $\widehat E^{S,\mathfrak{m}} $ can be written as
\begin{align*}
\widehat E^{S,\mathfrak{m}}  ~=~  \underbrace{\frac1{2n_{0,\text{te}}}\sum_{i=1}^{n_{0,\text{te}}} \I \Big[ \widehat{C}(Z_{\omega_i^{\mathfrak{m}}}) = 1 \Big]}_{(I)} + \underbrace{\frac1{2n_{1,\text{te}}}\sum_{i=1}^{n_{1,\text{te}}} \I \Big[ \widehat{C}(Z_{\omega_{i+n_{0,\text{te}}}^{\mathfrak{m}}}) = 0 \Big]}_{(II)}.
\end{align*}
The variance of the first term $(I)$ is
\begin{align*}
\V_{\mathfrak{m}}[(I)] ~=~ & \frac{1}{4n_{0,\text{te}}^2} \sum_{i=1}^{n_{0,\text{te}}} \V_{\mathfrak{m}}\Big\{   \I \Big[ \widehat{C}(Z_{\omega_i^{\mathfrak{m}}}) = 1 \Big] \Big\}  \\[.5em] 
& + \frac{1}{4n_{0,\text{te}}^2} \sum_{1 \leq i \neq j \leq n_{0,\text{te}}} \text{Cov}_{\mathfrak{m}} \Big\{ \I \Big[ \widehat{C}(Z_{\omega_i^{\mathfrak{m}}}) = 1 \Big], \ \I \Big[ \widehat{C}(Z_{\omega_j^{\mathfrak{m}}}) = 1 \Big]  \Big\},
\end{align*}
where the individual variance and covariance terms are given as
\begin{align*}
\V_{\mathfrak{m}}\Big\{   \I \Big[ \widehat{C}(Z_{\omega_i^{\mathfrak{m}}}) = 1 \Big] \Big\} =~ &  \frac{1}{n_{\text{te}}} \sum_{i=1}^{n_{\text{te}}} \I \Big[ \widehat{C}(Z_{i}) = 1 \Big] \cdot \bigg\{  1- \frac{1}{n_{\text{te}}} \sum_{i=1}^{n_{\text{te}}} \I \Big[ \widehat{C}(Z_{i}) = 1 \Big] \bigg\} \\
\leq ~& 1
\end{align*}
and
\begin{align*}
& \text{Cov}_{\mathfrak{m}} \Big\{ \I \Big[ \widehat{C}(Z_{\omega_i^{\mathfrak{m}}}) = 1 \Big], \ \I \Big[ \widehat{C}(Z_{\omega_j^{\mathfrak{m}}}) = 1 \Big]  \Big\} \\[.5em] 
~=~ & \frac{1}{n_{\text{te}} (n_{\text{te}} - 1)} \sum_{1 \leq i \neq j \leq n_{\text{te}}} \I \Big[ \widehat{C}(Z_{i}) = 1 \Big] \cdot  \I \Big[ \widehat{C}(Z_{j}) = 1 \Big]  -  \bigg\{   \frac{1}{n_{\text{te}}} \sum_{i=1}^{n_{\text{te}}} \I \Big[ \widehat{C}(Z_{i}) = 1 \Big]  \bigg\}^2 \\[.5em]
\leq ~ & 0.
\end{align*}
Hence the variance of $(I)$ is bounded by $\V_{\mathfrak{m}}[(I)]  \leq 1/(4n_{0,\text{te}})$ and similarly one can show that $\V_{\mathfrak{m}}[(II)]  \leq 1/(4n_{1,\text{te}})$. Now applying the basic inequality $\V(X+Y) \leq 2 \V(X) + 2 \V(Y)$ yields
\begin{align} \label{Eq: Variance bound}
\V_{\mathfrak{m}}\big[ \widehat E^{S,\mathfrak{m}} \big]  \leq \frac{1}{2n_{0,\text{te}}} + \frac{1}{2n_{1,\text{te}}}.
\end{align}
This in turn implies that $(\widetilde{E}^{S,i}  -  1/2)^2 \convP 0$ as $n \rightarrow \infty$ for any $i \in \{1,\ldots,P \}$ and thus
\begin{align} \label{Eq: convergence result}
\frac{1}{P} \sum_{i=1}^P \left( \widetilde{E}^{S,i}  -  1/2  \right)^2 \convP 0 \quad \text{as $n\rightarrow \infty$.}
\end{align}
This completes the proof for method 1.

\vskip 1em

\noindent \textbf{$\bullet$ Method 2 (Entire-permutation test).}

\vskip .5em

\noindent  Next we show that the squared difference~(\ref{Eq: squared difference}) converges to zero in probability for method 2. We first note that the half-permutation procedure can be understood as the entire-permutation procedure conditional on the first $n_{\text{tr}}$ permutation labels. From this perspective, $\E_{\mathfrak{m}}$ and $\V_{\mathfrak{m}}$ are the conditional expectation and the conditional variance of the permuted test statistic given the first $n_{\text{tr}}$ permutation labels. More specifically we let $\mathfrak{m}^\ast$ be a random variable uniformly distributed over $\{1,\ldots,n!\}$ and write the distribution of $\mathfrak{m}^\ast$ by $\mathbb{P}_{\mathfrak{m}^\ast}$ (conditional on everything else) and the expectation and the variance with respect to $\mathbb{P}_{\mathfrak{m}^\ast}$ by $\mathbb{E}_{\mathfrak{m}^\ast}$ and $\mathbb{V}_{\mathfrak{m}^\ast}$, respectively. Then for each $\mathfrak{m}^\ast$, there exists the corresponding permutation of $\{1,\ldots,n\}$, denoted by $\omega^{\mathfrak{m}^\ast} \defn \{ \omega_1^{\mathfrak{m}^\ast},\ldots, \omega_{n}^{\mathfrak{m}^\ast}\}$, such that the permuted test statistic can be expressed as a function of $\omega^{\mathfrak{m}^\ast}$ as 
\begin{align*}
\widehat E^{S,\mathfrak{m}^\ast} \defn \widehat E^{S,\mathfrak{m}^\ast} (Z_{\omega_1^{\mathfrak{m}^\ast}},\ldots, Z_{\omega_n^{\mathfrak{m}^\ast}}),
\end{align*}
where $\{Z_1,\ldots,Z_n\}$ are the pooled samples denoted by $\{Z_1,\ldots,Z_n\} \defn \{X_1,\ldots,X_{n_{0,\text{tr}}}, Y_1, \ldots, Y_{n_{1, \text{tr}}}, X_{1 + n_{0,\text{tr}}},\ldots,X_{n_{0}},Y_{1+n_{1,\text{tr}}},\ldots, Y_{n_{1}} \}$. Following the same reasoning in (\ref{Eq: expectation under permutation}), it can be seen that the conditional expectation of $\widehat E^{S,\mathfrak{m}^\ast}$ given the first $n_{\text{tr}}$ components of $\omega^{\mathfrak{m}^\ast}$ is always equal to half, that is
\begin{align*}
\E_{\mathfrak{m}^\ast} \Big[ \widehat E^{S,\mathfrak{m}^\ast} \Big| ~ \omega_1^{\mathfrak{m}^\ast},\ldots, \omega_{n_{\text{tr}}}^{\mathfrak{m}^\ast} \Big] = \frac{1}{2}.
\end{align*}
Hence applying the law of total expectation yields that the unconditional expectation is also equal to half. Next we use the law of total variance and observe that 
\begin{align*}
\mathbb{V}_{\mathfrak{m}^\ast} \Big[ \widehat E^{S,\mathfrak{m}^\ast}  \Big] ~=~ &  \mathbb{V}_{\mathfrak{m}^\ast} \Big[ \E_{\mathfrak{m}^\ast} \Big\{ \widehat E^{S,\mathfrak{m}^\ast} \Big| ~ \omega_1^{\mathfrak{m}^\ast},\ldots, \omega_{n_{\text{tr}}}^{\mathfrak{m}^\ast} \Big\} \Big] \\[.5em]
+ ~ & \mathbb{E}_{\mathfrak{m}^\ast} \Big[ \V_{\mathfrak{m}^\ast} \Big\{ \widehat E^{S,\mathfrak{m}^\ast} \Big| ~ \omega_1^{\mathfrak{m}^\ast},\ldots, \omega_{n_{\text{tr}}}^{\mathfrak{m}^\ast} \Big\} \Big] \\[.5em]
= ~ & \mathbb{E}_{\mathfrak{m}^\ast} \Big[ \V_{\mathfrak{m}^\ast} \Big\{ \widehat E^{S,\mathfrak{m}^\ast} \Big| ~ \omega_1^{\mathfrak{m}^\ast},\ldots, \omega_{n_{\text{tr}}}^{\mathfrak{m}^\ast} \Big\} \Big] ~ \leq ~  \frac{1}{2n_{0,\text{te}}} + \frac{1}{2n_{1,\text{te}}},
\end{align*}
where the last inequality can be similarly proved as in the bound~(\ref{Eq: Variance bound}). Having these two observations at hand, we know that conclusion~(\ref{Eq: convergence result}) is also true for method 2 and thus complete the proof of Theorem~\ref{Theorem: permutation test}.

\section{Simulation results on sample-splitting ratio}  \label{sec: Sample-Splitting Ratio}
In this section we examine the power of classification tests under the Gaussian setting by varying the splitting ratio $\kappa$ for the balanced sample case. As in Section~\ref{sec: experiments} of the main text, we set $n_0 = n_1 = d = 200$ and consider the accuracy tests $\varphi_{\Sigma^{-1}}$ and $\varphi_{\widehat{D}^{-1}}$ based on the Fisher's LDA classifier and the naive Bayes classifier, respectively. Note that the critical values of $\varphi_{\Sigma^{-1}}$ and $\varphi_{\widehat{D}^{-1}}$ are chosen based on a normal approximation. Given $\kappa \in \{0.1,0.2,\ldots,0.9\}$, the number of samples in the training set is decided by $n_{0,\text{tr}} = \lfloor \kappa n_0  \rfloor$ and $n_{1,\text{tr}} = \lfloor \kappa n_1  \rfloor$, which leads to $n_{0,\text{te}} = n_0 - n_{0,\text{tr}}$ and $n_{1,\text{te}} = n_1 - n_{1,\text{tr}}$.

\begin{table}[h!]
	\begin{center}
		\renewcommand{\tabcolsep}{3.8pt}
		\renewcommand\arraystretch{1}
		\small
		\caption{\small Comparisons of the empirical power of classification tests by varying the sample-splitting ratio $\kappa$. The results show that the power is approximately maximized when the splitting ratio is $\kappa=1/2$. See Appendix~\ref{sec: Sample-Splitting Ratio} for details.}  \label{Table: Splitting ratio}
		\begin{tabular}{clccccccccc} \toprule
			& Ratio $\kappa$     & 0.1   & 0.2   & 0.3   & 0.4   & 0.5   & 0.6   & 0.7   & 0.8   & 0.9   \\  \hline
			\multirow{2}{*}{$\delta=0.15$} & LDA & 0.155 & 0.189 & 0.212 & 0.220 & \textbf{0.224} & 0.207 & 0.176 & 0.157 & 0.103 \\
			& Bayes     & 0.150 & 0.185 & 0.218 & 0.221 & \textbf{0.222} & 0.212 & 0.176 & 0.156 & 0.100 \\ \midrule
			\multirow{2}{*}{$\delta=0.25$} & LDA & 0.437 & 0.616 & 0.691 & 0.714 & \textbf{0.715} & 0.686 & 0.598 & 0.499 & 0.301 \\
			& Bayes     & 0.406 & 0.613 & 0.682 & 0.710 & \textbf{0.714} & 0.677 & 0.596 & 0.496 & 0.306 \\ \bottomrule
		\end{tabular}
	\end{center}
\end{table}

The results are presented in Table~\ref{Table: Splitting ratio}. It is apparent from Table~\ref{Table: Splitting ratio} that the power is maximized when the training set and the testing set are well-balanced, i.e.~$\kappa=1/2$. This coincides with our theoretical result discussed in Section~\ref{sec: power of LDA}. However, unlike our asymptotic power expression in (\ref{Eq: Power approximation}) with $\lambda = 1/2$, the empirical power seems asymmetric in $\kappa$. This unexpected result might be attributed to the fact that when $\kappa$ is far from $1/2$, either $n_{\text{tr}}$ or $n_{\text{te}}$ becomes too small to justify a normal approximation. Nevertheless, the powers in these extreme cases are less than the power in the balanced case.


\end{document}